\documentclass[runningheads]{llncs}

\usepackage{graphicx}
\usepackage{comment}
\usepackage{amsmath,amssymb}
\usepackage{color}
\usepackage{url}
\usepackage{hyperref}
\usepackage{graphicx}

\usepackage{tikz}
\usepackage{comment}
\usepackage{amsmath,amssymb} 
\usepackage{color}
\usepackage{booktabs}
\usepackage{multirow}
\usepackage{fixltx2e}
\usepackage{dblfloatfix}
\usepackage{subcaption}
\usepackage[abs]{overpic}
\usepackage{cite}

\newcommand{\modelname}{LocOv }
\newcommand{\ex}{\emph{e.g.}}
\newcommand{\etal}{\emph{et al.}}
\newcommand{\eg}{\emph{e.g.}}

\newif\ifreview
\reviewfalse

\ifreview
	\usepackage{lineno}

	\linenumbers
\fi

\begin{document}


\def\SubNumber{000}

\def\GCPRTrack{Main Track}

\title{Localized Vision-Language Matching for Open-vocabulary Object Detection}

\ifreview
\else

	\author{Mar\'ia A. Bravo \and
	Sudhanshu Mittal \and
	Thomas Brox}
	
	\authorrunning{M. Bravo et al.}
	
	\institute{Department of Computer Science \\ University of Freiburg, Germany \\
	\email{\{bravoma, mittal, brox\}@cs.uni-freiburg.de}}
\fi

\maketitle              

\begin{abstract}
In this work, we propose an open-vocabulary object detection method that, based on image-caption pairs, learns to detect novel object classes along with a given set of known classes. It is a two-stage training approach that first uses a location-guided image-caption matching technique to learn class labels for both novel and known classes in a weakly-supervised manner and second specializes the model for the object detection task using known class annotations. 
We show that a simple language model fits better than a large contextualized language model for detecting novel objects. Moreover, we introduce a consistency-regularization technique to better exploit image-caption pair information. Our method compares favorably to existing open-vocabulary detection approaches while being data-efficient. Source code is available at \url{https://github.com/lmb-freiburg/locov}.

\end{abstract}
%
%
%
\section{Introduction}\label{sec:intro}
Recent advances in deep learning have rapidly advanced the state-of-the-art object detection algorithms. The best mean average precision score on the popular COCO \cite{502} benchmark has improved from 40 mAP to over 60 mAP in less than 4 years.  However, this success required large datasets with annotations at the bounding box level and was a achieved in a closed-world setting, where the number of classes is assumed to be fixed. The closed-world setting restricts the object detector to only discover known annotated objects and annotating all possible objects in the world is infeasible due to high labeling costs. Therefore, research of open-world detectors, which can also discover unmarked objects, has recently come into focus specially using textual information together with images for open-vocabulary detection~\cite{zareian2021open, gu2022openvocabulary, zhong2021regionclip}.

\begin{figure}
\centering
\begin{subfigure}[t]{0.43\linewidth}
\includegraphics[width=\linewidth]{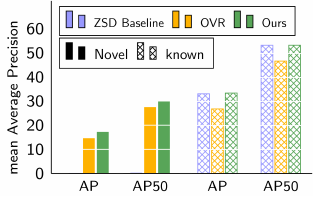}
\caption{SOTA method comparison}\label{fig:teaser_graph}
\end{subfigure}
\begin{subfigure}[t]{.18\linewidth}
\includegraphics[width=\linewidth]{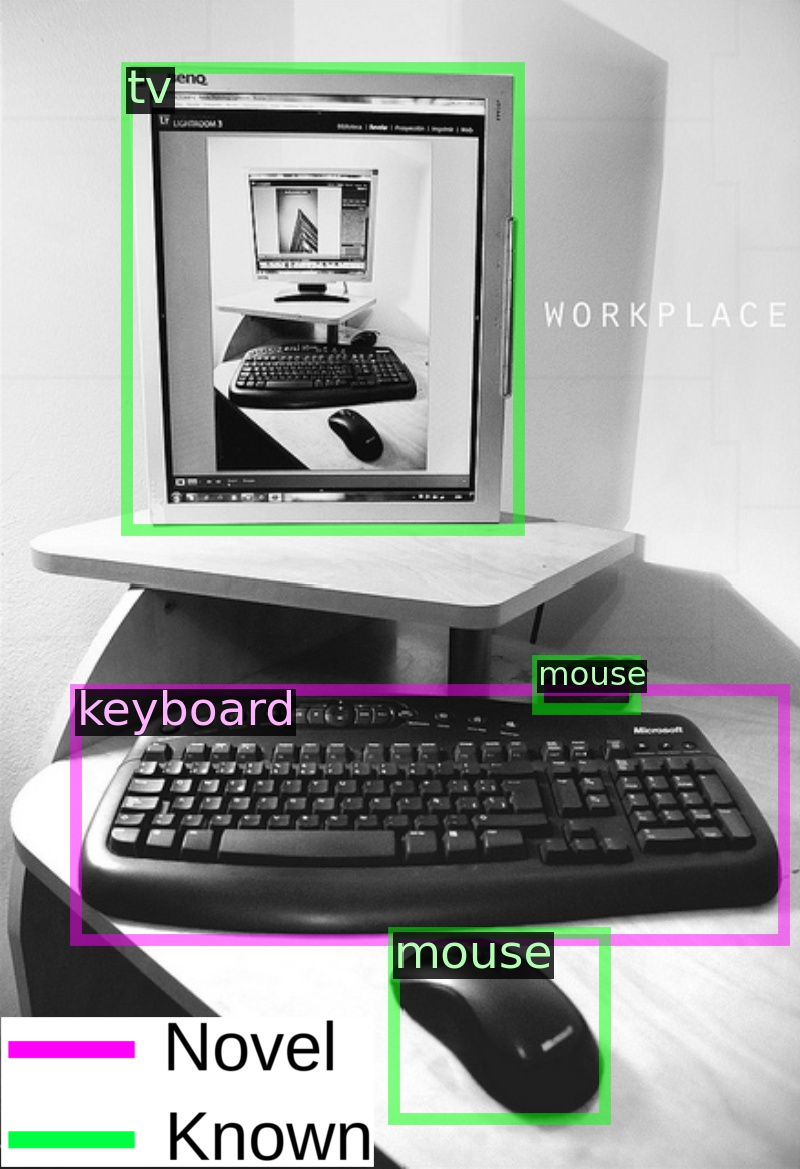}
\caption{GT}\label{fig:teaser_gt}
\end{subfigure}
\begin{subfigure}[t]{.18\linewidth}
\includegraphics[width=\linewidth]{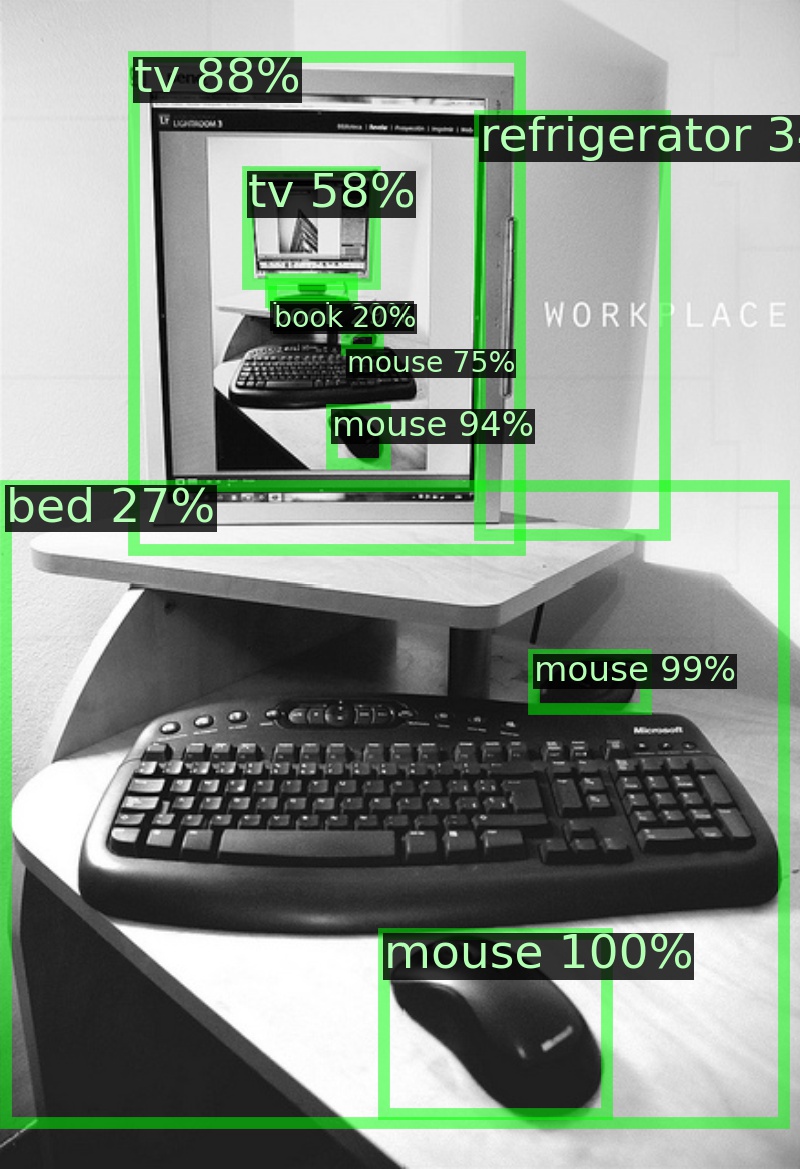}
\caption{ZSD}\label{fig:teaser_imagenet}
\end{subfigure}
\begin{subfigure}[t]{.18\linewidth}
\includegraphics[width=\linewidth]{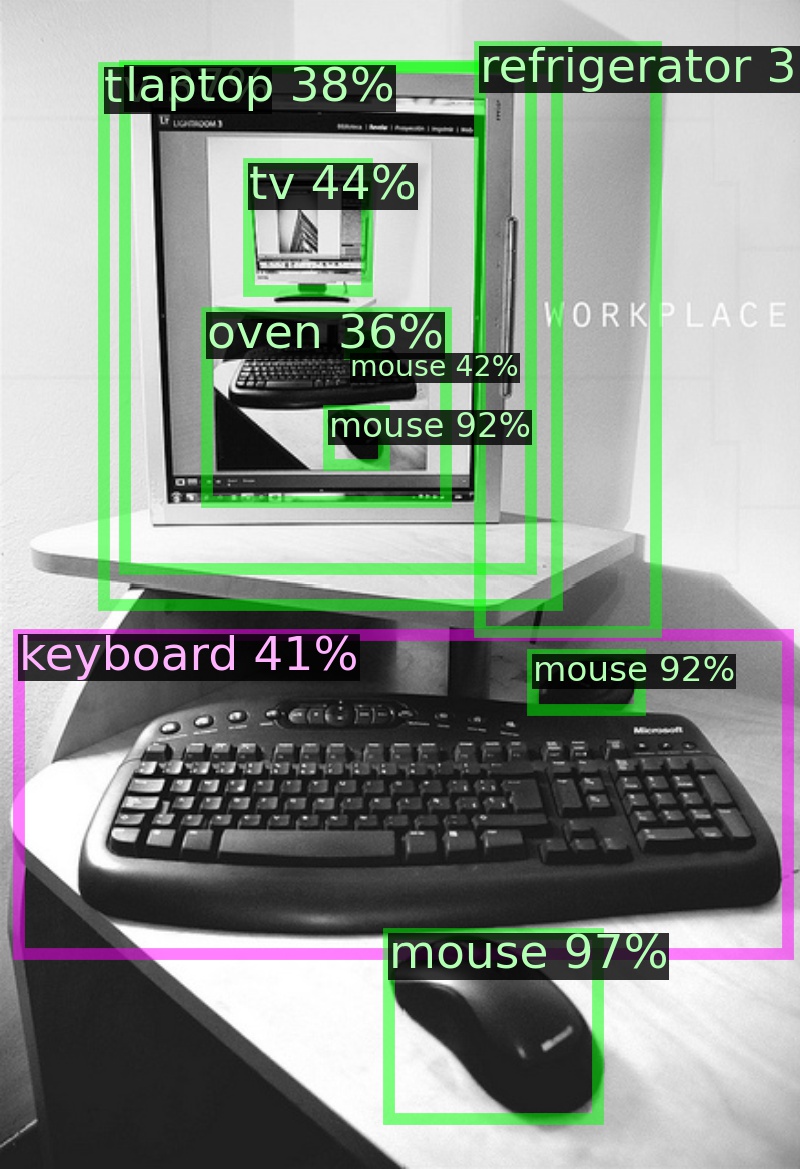}
\caption{\modelname}\label{fig:teaser_ours}
\end{subfigure}
\caption{Open-vocabulary object detection. (a) Compares our method \modelname with the baseline method (OVR) and our zero-shot baseline STT-ZSD (ZSD). \modelname improves on both novel and known classes without dropping the performance on known classes. The zero-shot method, only trained with known classes, obtains low performance ($<0.5$ mAP) on novel classes. (b-d) \modelname is able to detect the novel object `keyboard' along with known objects, shown in figure. } 
\label{fig:teaser}
\end{figure}

To learn a visual concept, humans receive the majority of the supervision in the form of narrations rather than class tags and bounding boxes. Consider the example of Figure \ref{fig:teaser} together with the annotations of mouse and tv only. Even after learning to detect these objects, finding and identifying the keyboard without any other source of information is ambitious. Instead, if we consider the image together with the caption - ``A mouse, keyboard, and a monitor on a desk", it is possible to identify that the other salient object in the image is very likely a keyboard. This process involves successful localization of the objects in the scene, identification of different nouns in the narrated sentence, and matching the two together.
Exploiting the extensive semantic knowledge contained in natural language is a reasonable step towards learning such open-vocabulary models without expensive annotation costs. 


In this work, we aim to learn novel objects using image-caption pairs. 
Along with image-caption pairs, the detector is provided with box annotations for a limited set of classes. We follow the problem setting as introduced by Zareian \etal \cite{zareian2021open}. They refer to this problem as \textit{Open-vocabulary Object Detection}.
There are two major challenges to this problem: First, image-caption pairs themselves are too weak to learn localized object-regions. Analyzing previous works, we find that randomly sampled feature maps provide imprecise visual grounding for foreground objects, therefore they receive insufficient supervisory signals to learn object properties.
Second, the granularity of the information captured by image-region features should align with the level of information captured by the text representation for an effective matching.
For example, it would be ill-suited to match a text representation that captures global image information with image features that capture localized information.


In this work, we propose a method that improves the matching between image and text representations. Our model is a two-stage approach: in the first stage, \textit{Localized Semantic Matching} (LSM), it learns semantics of objects in the image by matching image-regions to the words in the caption; and in the second stage, \textit{Specialized Task Tuning} (STT), it learns specialized visual features for the target object detection task using known object annotations. We called our method \modelname for \textbf{Loc}alized Image-Caption Matching for \textbf{O}pen-\textbf{v}ocabulary.

For the given objects in an image, our goal is to project them to a feature space where they can be matched with their corresponding class in the form of text embeddings. We find that simple text embeddings are better candidates for matching object representations than contextualized embeddings produced by large-scale language models.

Using image-caption pairs as weak supervision for object detection requires the understanding of both modalities in a fine and a coarse way. This can be obtained by processing each modality independently in a uni-modal fashion and then matching, or using cross-modal attention to process them together. To ensure consistent training between the uni-modal and cross-modal methods, we propose a consistency-regularization between the two matching scores. 
To summarize, our contributions are: (1) We introduced localized-regions during the image-caption matching stage to improve visual feature learning of objects. (2) We show that simplified text embeddings match better with identified object features as compared to contextualized text embeddings. (3) We propose a consistency regularization technique to ensure effective cross-modal training.

These three contributions allow \modelname to be not only competitive against state-of-the-art models but also data-efficient by using less than 0.6 million image-caption pairs for training, $\sim$700 times smaller than CLIP-based methods. Additionally, we define an open-vocabulary object detection setup based on the VAW \cite{Pham_2021_CVPR} dataset, which offers challenging learning conditions like few-instances per object and a long-tailed distribution. Based on the above mentioned three contributions, we show that our method achieves state-of-the-art performance on both open-vocabulary object detection benchmarks, COCO and VAW. 

\section{Related Work}\label{sec:related_work}
\textbf{Object detection with limited supervision}
Semi-supervised (SSOD) \cite{NEURIPS2019_d0f4dae8, liu2021unbiased,  DBLP:journals/corr/abs-2005-04757} and weakly-supervised (WSOD) \cite{Bilen_2016_CVPR, 7420739, Kosugi_2019_ICCV} object detection are two widely explored approaches to reduce the annotation cost. WSOD approaches aim to learn object localization using image-level labels only. Major challenges in WSOD approaches include differentiation between object instances \cite{Ren_2020_CVPR} and precisely locating the entire objects. SSOD approaches use a small fully-annotated set and a large set of unlabeled images. Best SSOD \cite{liu2021unbiased, DBLP:journals/corr/abs-2005-04757} methods are based on pseudo-labeling, which usually suffers from foreground-background imbalance and overfitting on the labeled set of images. In this work, we address a problem which shares similar challenges with the WSOD and SSOD approaches, however they are limited to a closed-world setting with a fixed and predefined set of classes. Our method addresses a mixed semi- and weakly-supervised object detection problem where the objective is open-vocabulary object detection.

\textbf{Multi-modal visual and language models.}
Over the past years, multiple works have centered their attention on the intersection of vision and language by exploiting their consistent semantic information contained in matching pairs. The success of using this pairwise information has proved to be useful for pre-training transformer-like models for various vision-language tasks \cite{lu2019vilbert, tan2019lxmert, chen2019uniter, su2019vl, zhou2020unified, zhang2021vinvl, li2020oscar} which process the information jointly using cross-attention. 
Other approaches \cite{klein2015associating, wang2018learning, dong2019dual, ging2020coot, radford2021learning, Miech_2021_CVPR}, centered on the vision and language retrieval task use separate encoders for each modality, in a uni-modal fashion. These models give the flexibility to transfer the knowledge learned by the pairwise information to single modality tasks, which is the case of object detection. In particular Miech \etal \cite{Miech_2021_CVPR} showed that combining a cross-attention model with two uni-modal encoders 
is beneficial for large-scale retrieval tasks. In this paper, we combine the strengths of both types of approaches to train a model using different consistency losses that exploit the information contained in image-caption pairs.


 \begin{figure*}[t!]
    \centering
    \includegraphics[trim={0 0cm 0cm 0cm}, clip, width=1.00\textwidth]{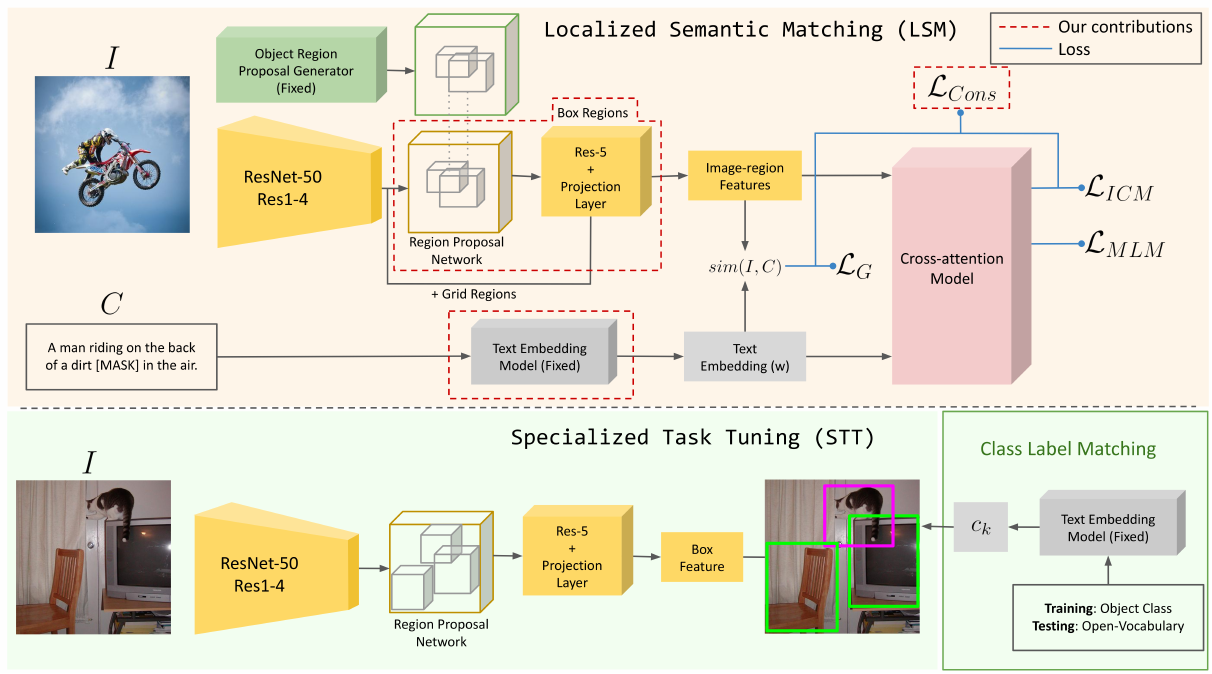}
    \caption{Overview of \modelname. It is a two-stage model: (1) Localized Semantic Matching stage trains a Faster R-CNN-based model to match corresponding image-caption pairs using a grounding loss $\mathcal{L}_G$. We exploit the multi-modal information by using a cross-attention model and an Image-Caption matching loss $\mathcal{L}_{ICM}$, the mask language modeling loss $\mathcal{L}_{MLM}$ and a consistency-regularization loss $\mathcal{L}_{Cons}$. 
    (2) Specialized Task Tuning stage tunes the model using the known class annotations and specializes the model for object detection. See Section \ref{sec:method}.}
    \label{fig:pipeline}
\end{figure*}
\textbf{Language-guided object detection.}
Zero-shot object detection methods learn to align proposed object-region features to the class-text embeddings. Bansal \etal \cite{bansal2018zero} is among the first to propose the zero-shot object detection problem. They identified that the main challenge in ZSD is to separate the background class from the novel objects. Zhu \etal \cite{Zhu_2020_CVPR} trained a generative model to ``hallucinate" (synthesize visual features) unseen classes and used these generated features during training to be able to distinguish novel objects from background.
Rahman \etal \cite{Rahman_Khan_Barnes_2020} proposed a polarity loss to handle foreground-background imbalance and to improve visual-semantic alignment.
However, such methods fail to perform well on the novel classes since the detection model has never seen these novel objects, and semantics learned by matching known object-text embeddings does not extrapolate to novel classes. 

%
To learn the semantics of novel classes, recent methods \cite{zareian2021open, NEURIPS2020_acaa23f7, gu2022openvocabulary, huynh2021open, zhong2021regionclip} have simplified the problem by providing image-caption pairs as a weak supervision signal. Such pairs are cheap to acquire and make the problem tractable. Image-caption pairs allow the model to observe a large set of object categories along with object labels. 
These methods either use this model to align image-regions with captions and generate object-box pseudo labels \cite{huynh2021open, zhong2021regionclip} or as region-image feature extractor to classify the regions \cite{gu2022openvocabulary}. 
Many weakly-supervised \cite{chen-etal-2019-weakly, NEURIPS2020_acaa23f7, Amrani_2020_CVPR_Workshops, Sadhu_2020_CVPR, Zhang_2020_CVPR} approaches have been proposed to perform such object grounding.
Due to the large performance gap between zero-shot/weakly-supervised and fully-supervised approaches for object detection, Zareian \etal \cite{zareian2021open} introduced an open-vocabulary problem formulation. It utilizes extra image-caption pairs to learn to detect both known and novel objects. Their approach matches all parts of the image with the caption, whereas we emphasize object localized regions and a consistency loss to enforce more object-centric matching.


\section{Method}\label{sec:method}

We propose a two-stage approach for the task of open-vocabulary object detection as shown in Figure \ref{fig:pipeline}. The first stage, \textit{Localized Semantic Matching} (LSM), learns to match objects in the image to their corresponding class labels in the caption in a weakly-supervised manner. The second stage, \textit{Specialized Task Tuning} (STT) stage, includes specialized training for the downstream task of object detection. We consider two sets of object classes: known classes $O_K$ and novel classes $O_N$. Bounding box annotations, including class labels, are available for known classes whereas there are no annotations for the novel classes.

The LSM receives image-caption pairs $(I, C)$ as input, where the caption provides the weak supervision to different image-regions. Captions contain rich information which often include words corresponding to object classes from both known and novel sets. Captions are processed using a pre-trained text-embedding model (\ex BERT \cite{DBLP:journals/corr/abs-1810-04805} embedding) to produce word or part-of-word features. Images are processed using an object detection network (Faster R-CNN \cite{NIPS2015_14bfa6bb}) to obtain object region features.
We propose to utilize an object proposal generator OLN \cite{kim2021oln} to provide regions as pseudo-labels to train the Faster R-CNN. 
This helps obtaining object-rich regions which improve image region-caption matching. This way, during the LSM our model learns to match all present objects in the image in a class-agnostic way. See Section \ref{sec:LSM} for details.
The STT stage tunes the Faster R-CNN using known object annotations primarily to distinguish foreground from background and learns corresponding precise location of the foreground objects. See Section \ref{sec:STT} for details.

\subsection{Localized Semantic Matching (LSM)} \label{sec:LSM}
The LSM stage consists of three main components: (1) localized object region-text matching, (2) disentangled text features and (3) consistency-regularization.

\textbf{Localized object region-text matching.}
Given the sets 
$R^I$ = \{ $r : r$ is an image-region feature vector from the image I\} and 
$W^C$ = \{ $w : w$ is a word or part-of-word feature vector from the caption C\},
we calculate the similarity score between an image and a caption in a fine-grained manner, by comparing image-regions with words, since our final objective is to recognize objects in regions. The image is processed using a Faster R-CNN model and a projection layer that maps image-regions into the text-embedding feature space. 
The similarity score is calculated by taking an image composed of  $|R^I|$ region features and a caption composed of $|W^C|$ part-of-word features by:
\begin{equation}\label{eq:sim_score}
    sim(I, C) = \frac{1}{|R^I|}\sum_{i=1}^{|R^I|}\sum_{j=1}^{|W^C|} d_{i,j} (r_i \cdot w_j)
\end{equation}
where $d_{i,j}$ corresponds to:
\begin{equation}
    d(r_i,w_j) = d_{i,j} = \frac{\exp(r_i \cdot w_j)}{\sum_{j'=1}^{|W^C|}\exp(r_i \cdot w_{j'})}.
\end{equation}

Based on the similarity score (Eq.~\ref{eq:sim_score}) , we apply a contrastive learning objective to match the corresponding pairs together by considering all other pairs in the batch as negative pairs. We define this grounding loss as:
\begin{equation} \label{eq:grounding_loss}
    \mathcal{L}_{G_r}(I) = -\log \frac{\exp(sim(I, C))}{\sum_{C' \in \text{Batch}}\exp(sim(I, C'))}
\end{equation}
We apply this loss in a symmetrical way, where each image in the batch is compared to all captions in the batch (Eq.~\ref{eq:grounding_loss}) and each caption is compared to all images in the batch $\mathcal{L}_{G_r}(C)$.
The subscript $r$ denotes the type of image-regions used for the loss calculation. We consider two types of image-regions: box-regions and grid-regions. Box-region features are obtained naturally using the region of interest pooling (RPN) from the Faster R-CNN.
We make use of the pre-trained object proposal generator (OLN) to train the Faster-RCNN network.
OLN is a class-agnostic object proposal generator which estimates all objects in the image with a high average recall rate. We train OLN using the known class annotations and use the predicted boxes to train our detection model, shown in Figure \ref{fig:pipeline}.
Since captions sometimes refer to background context in the image, parallel to the box-region features, we also use grid-region features similar to the OVR  \cite{zareian2021open} approach.
Grid-region features 
are obtained by skipping the RPN in the Faster R-CNN and simply using the output of the backbone network. 
We apply the grounding loss to both type of image-region features. 
Our final grounding loss is given by:
\begin{equation} 
    \mathcal{L}_G = \mathcal{L}_{G_{box}}(C)+\mathcal{L}_{G_{box}}(I)+\mathcal{L}_{G_{grid}}(C)+\mathcal{L}_{G_{grid}}(I)
\end{equation}

\textbf{Disentangled text features.} 
Many previous works \cite{huang2020pixel, chen2019uniter, lu2019vilbert, su2019vl} use contextualized language models to extract text representations of the sentence. Although, this might be suitable for a task that requires a global representation of a phrase or text, this is not ideal for the case for object detection, where each predicted bounding box is expected to contain a single object instance.
We show that using a simple text representation, which keeps the disentangled semantics of words in a caption, gives the flexibility to correctly match object boxes in an image with words in a caption. 
Our method uses only the embedding module \cite{DBLP:journals/corr/abs-1810-04805, pennington2014glove} of a pre-trained language model to encode the caption and perform matching with the proposed image-regions. For embedding model we refer to the learned dictionary of vector representations of text tokens, which correspond to words or part-of-words. For cases where the text representing an object category is divided into multiple tokens, we consider the average representation of the tokens as the global representation of the object category. We show empirically, in Section \ref{sec:ablations}, that using such a lightweight text embedding module has better performance than using a whole large-scale language model.



\textbf{Consistency-regularization}
Miech \etal \cite{Miech_2021_CVPR} showed that processing multi-modal data using cross-attention networks brings improvements in retrieval accuracy over using separate encoders for each modality and projecting over a common embedding space. However, this cross-attention becomes very expensive when the task requires large-scale retrieval. To take the benefit of cross-attention models, we consider a model similar to PixelBERT \cite{huang2020pixel} to process the image-caption pairs. This cross-attention model takes the image-regions $R^I$ together with the text embeddings $W^C$ and matches the corresponding image-caption pairs in a batch. 
The image-caption matching loss ($\mathcal{L}_{ICM}$) of the cross-attention model together with the traditional Masking Language Modeling loss ($\mathcal{L}_{MLM}$) enforces the model to better project the image-region features to the language semantic space.
To better utilize the cross-attention model, we propose a consistency-regularization loss ($\mathcal{L}_{Cons}$) between the final predicted distribution over the image-caption matching scores in the batch, before and after the cross-attention model.
We use the Kullback-Leibler divergence loss to impose this consistency. In summary, we use three consistency terms over different image-caption pairs: 
\begin{align}
    \mathcal{L}_{Cons}= & D_{KL}(p(I_{box}, C)|| q(I_{box}, C)) \notag \\ &
    + D_{KL}(p(I_{grid}, C)|| q(I_{grid}, C)) \notag \\ 
    &  + D_{KL}(p(I_{grid}, C)|| q(I_{box}, C))
\end{align}
where $p(I_*, C)$ and $q(I_*, C)$ correspond to the softmax of the image-caption pairs in a batch before and after the cross-attention model respectively, and the sub-index of the image corresponds to the box- or grid-region features.
Our final loss for the LSM stage corresponds to the sum of the above defined losses:
\begin{equation}
    \mathbf{L}_{LSM} = \mathcal{L}_G + \mathcal{L}_{ICM} + \mathcal{L}_{MLM} + \mathcal{L}_{Cons}
\end{equation}

\subsection{Specialized Task Tuning (STT)} \label{sec:STT}
In this stage, we fine-tune the model using known class annotations to learn to localize the objects precisely.
We initialize the weights from the LSM stage model, and partially freeze part of the backbone and the projection layer to preserve the learned semantics. 
Freezing the projection layer is important to avoid overfitting on the known classes and generalize on novel classes.
To predict the class of an object, we compute the similarity score between the proposed object box-region feature vector ($r_i$) and all the class embedding vectors $c_k$ and apply softmax
\begin{equation}
    p(r_i,c_k) = \frac{\exp(r_i \cdot c_k)}{1+\sum_{c_k' \in O_K} \exp(r_i \cdot c_{k'})}.
\end{equation} 
The scalar $1$ included in the denominator corresponds to the background class, which has a representation vector of all-zeros.
We evaluate the performance across three setups: (Novel) considering only the novel class set $O_N$, (Known) comparing with the known classes only $O_K$ and (Generalized) considering all novel and known classes together.


\section{Experiments}\label{sec:exp}

\subsection{Training Details}
\subsubsection{Datasets.}
The \textbf{Common Objects in Context (COCO) dataset} \cite{lin2014microsoft} is a large-scale object detection benchmark widely used in the community. We use the 2017 train and val split for training and evaluation respectively. We use the known and novel object class splits proposed by Bansal~\etal~\cite{bansal2018zero}. The known set consists of 48 classes while the novel set has 17 classes selected from the total of 80 classes of the original COCO dataset. We remove the images which do not contain the known class instances from the training set. 
For the localized semantic matching phase, we use the captions from \textbf{COCO captions \cite{chen2015microsoft}} dataset which has the same train/test splits as the COCO object detection task. COCO captions dataset contains 118,287 images with 5 captions each. 
Additionally in the supplementary material, we test \modelname using \textbf{Visual Attributes in the Wild (VAW) dataset \cite{Pham_2021_CVPR}} a more challenging dataset containing fine-grained classes with a long-tailed distribution.
\begin{table*}[t]\scriptsize
\centering 
    \caption{Comparing mAP and AP\textsubscript{50} state-of-the-art methods. \modelname outperforms all other methods for Novel objects in the generalized setup while using only 0.6M of image-caption pairs. Training dataset: $^*$ImageNet1k, $^\mathsection$COCO captions, $^\dagger$CLIP400M, $^\ddagger$Conceptual Captions, $^\star$Open Images, and $^c$COCO
    }
    \label{tab:coco_sota_gen}
\begin{tabular}{ |c|c|c c|c c|c c|c c|c c|} 
 \hline 
 \multirow{3}{*}{Method} & Img-Cap &\multicolumn{4}{|c|}{Constrained} & \multicolumn{6}{|c|}{Generalized} \\
 & Data &\multicolumn{2}{|c|}{Novel (17)} & \multicolumn{2}{|c|}{Known (48)} & \multicolumn{2}{|c|}{Novel (17)} & \multicolumn{2}{|c|}{Known (48)} & \multicolumn{2}{|c|}{All (65)}\\ 
 & Size & AP & AP\textsubscript{50} & AP & AP\textsubscript{50} & AP & AP\textsubscript{50} & AP & AP\textsubscript{50} & AP & AP\textsubscript{50}\\ 
 \hline
 Faster R-CNN & \multirow{6}{*}{-} & & - & - & 54.5 & - & - & - & - & - & - \\ %
 \hline
 SB \cite{bansal2018zero} & & - & 0.70 & - & 29.7 & - & 0.31 & - & 29.2 & - & 24.9 \\ %
 LAB \cite{bansal2018zero} & & - & 0.27 & - & 21.1 & - & 0.22 & - & 20.8 & - & 18.0 \\ %
 DSES \cite{bansal2018zero} & & - & 0.54 & - & 27.2 & - & 0.27 & - & 26.7 & - & 22.1 \\ %
 DELO \cite{Zhu_2020_CVPR} & & - & 7.6 & - & 14.0 & - & 3.41 & - & 13.8 & - & 13.0 \\ 
 PL \cite{Rahman_Khan_Barnes_2020} & & - & 10.0 & - & 36.8 & - & 4.12 & - & 35.9 & - & 27.9 \\
 STT-ZSD \tiny{(Ours)} & & 0.21 & 0.31 & 33.2 & 53.4 & 0.03 & 0.05 & \textbf{33.0} & 53.1 & 24.4 & 39.2 \\
 \hline
 OVR$^{*\mathsection c}$ \cite{zareian2021open} & \multirow{2}{*}{0.6M} & 14.6 & 27.5 & 26.9 & 46.8 & - & 22.8 & - & 46.0 & 22.8 & 39.9\\
 \modelname$^{*\mathsection c}$ \tiny{(Ours)}& &\textbf{17.2} & 30.1 & \textbf{33.5} & 53.4 & \textbf{16.6} & \textbf{28.6} & 31.9 & 51.3 & \textbf{28.1} & 45.7\\
 \hline
 XP-Mask$^{\ddagger \mathsection \star c}$ \cite{huynh2021open}& 5.7M & - & 29.9 & - & 46.8 & - & 27.0 & - & 46.3 & - & 41.2 \\ 
 CLIP (cropped reg)$^\dagger$ \cite{gu2022openvocabulary}& 400M & - & - & - & - & - & 26.3 & - & 28.3 & - & 27.8 \\ 
 RegionCLIP$^{\dagger \mathsection c}$ \cite{zhong2021regionclip}& 400.6M & - & \textbf{30.8} & - & \textbf{55.2} & - & 26.8 & - & 54.8 & - & 47.5 \\ 
 ViLD$^{\dagger c}$ \cite{gu2022openvocabulary}& 400M & - & - & - & - & - & 27.6 & - & \textbf{59.5} & - & \textbf{51.3} \\ 
 \hline
    \end{tabular}
\end{table*}

\subsubsection{Evaluation metric.} 
We evaluate our method using mean Average Precision (AP) over IoU scores from 0.5 to 0.95 with a step size of 0.05, and using two fixed thresholds at 0.5 (AP\textsubscript{50}) and 0.75 (AP\textsubscript{75}). We compute these metrics separately for novel and known classes, calculating the softmax within the subsets exclusively; and in a generalized version both sets are evaluated in a combined manner, calculating the probability across all classes.

\subsubsection{Implementation details.}
We base our model on Faster R-CNN C4 \cite{NIPS2015_14bfa6bb} configuration, using ResNet50 \cite{7780459} backbone pre-trained on ImageNet \cite{5206848}, together with a linear layer (projection layer) to obtain the object feature representations. We use Detectron2 framework \cite{wu2019detectron2} for our implementation. For the part-of-word feature representations, we use the embedding module of the pre-trained BERT \cite{DBLP:journals/corr/abs-1810-04805} ``base-uncased" model from the HuggingFace implementation \cite{wolf-etal-2020-transformers}. To get the object proposals for the LSM stage, we train a generic object proposal network, OLN \cite{kim2021oln}. OLN is trained using only the known classes on COCO training set. We use all the proposals generated for the training images which have an objectness score higher than 0.7. For our cross-attention model, we use a transformer-based architecture with 6 hidden layers and 8 attention heads trained from scratch. We train our LSM stage with a base learning rate of 0.001, where the learning rate is divided by 10 at 45k and 60k iterations. We use a batch size of 32 and train on 8 GeForce-RTX-2080-Ti GPUs for 90k iterations.
For the STT stage, we initialize the weights of the Faster R-CNN and projection layer from the LSM stage, freezing the first two blocks of ResNet50 and the projection layer. For object classes that contain more than one part-of-word representation given BERT embedding module, we consider the average of their vector representation. We use a base learning rate of 0.005 with a 10 times drop at 60k iterations and do early stopping to avoid over-fitting. 


\begin{figure}[t!]
\centering
\begin{tabular}{c c c c}
(a) Ground Truth & (b) STT-ZSD & (c) OVR \cite{zareian2021open} & (d) \modelname  \\[6pt]

 \includegraphics[trim={0cm 0cm 0 0cm}, clip, width=30mm]{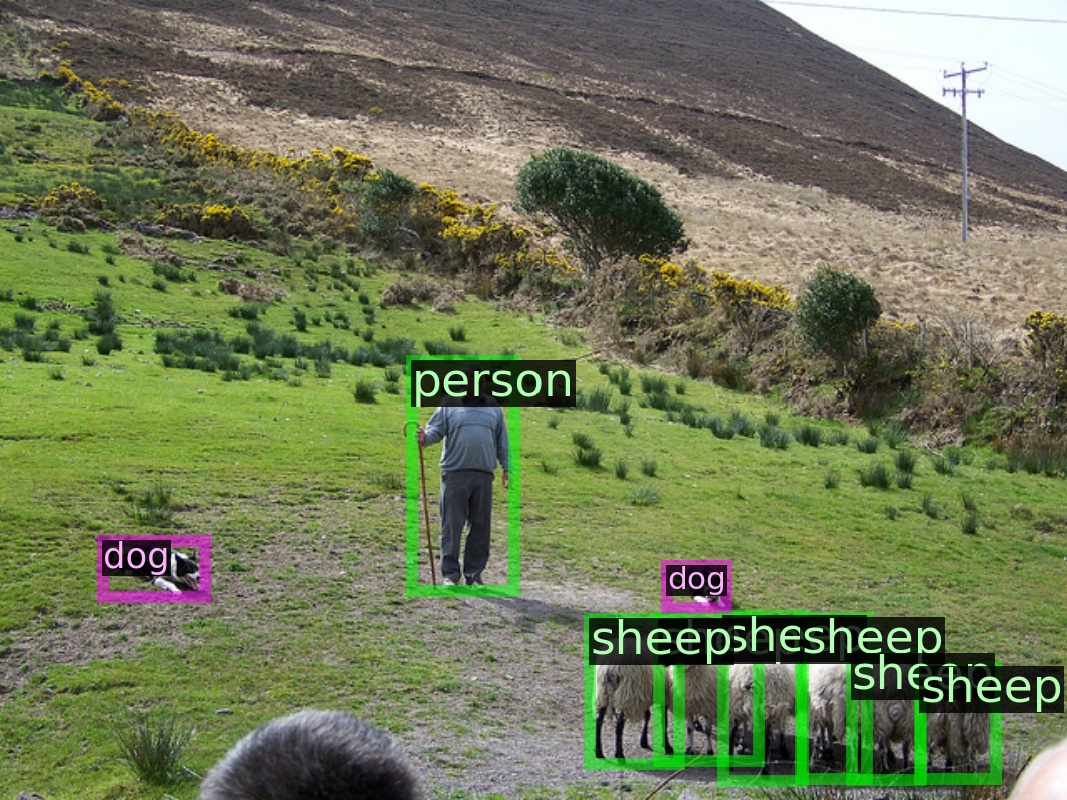} &
 \includegraphics[trim={0 0cm 0 0cm}, clip, width=30mm]{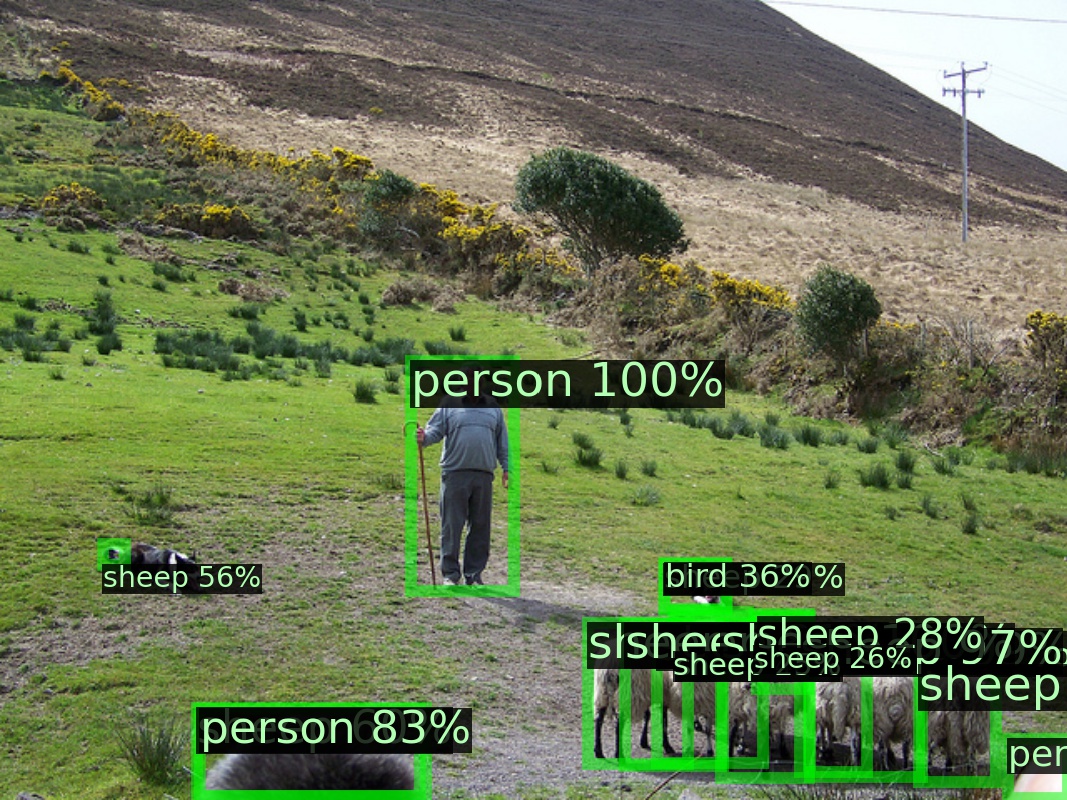} &
 \includegraphics[trim={0 0cm 0 0cm}, clip, width=30mm]{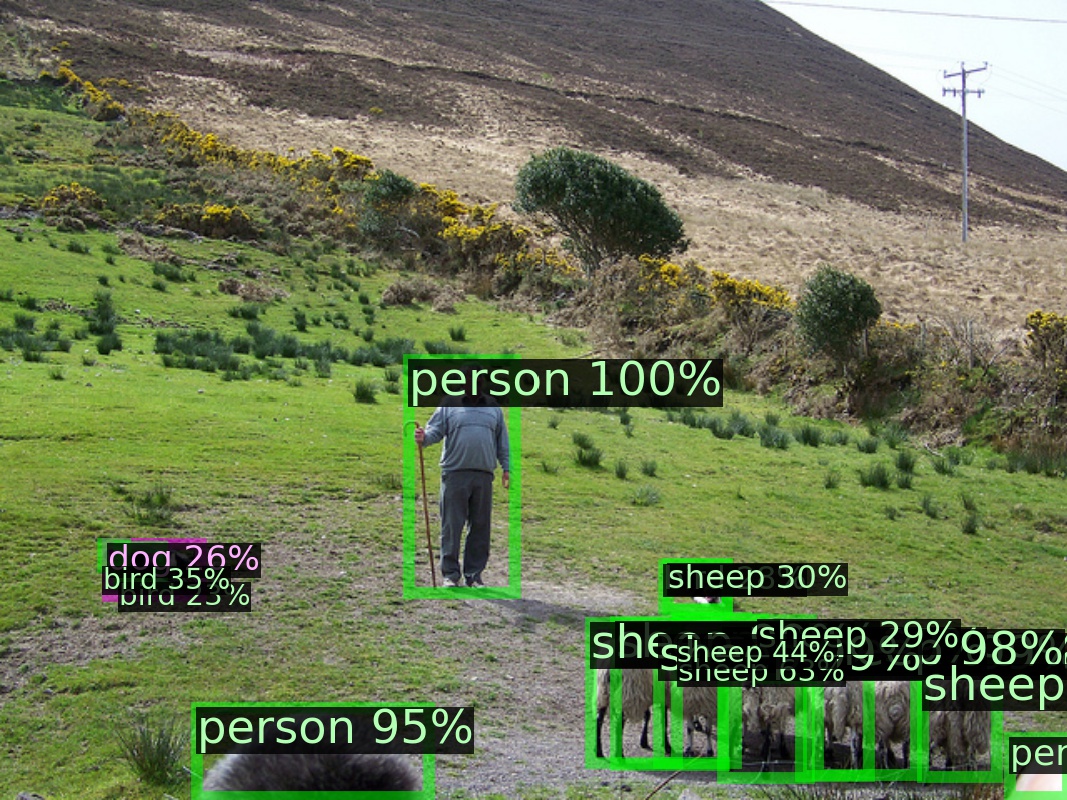} &
 \includegraphics[trim={0 0cm 0 0cm}, clip, width=30mm]{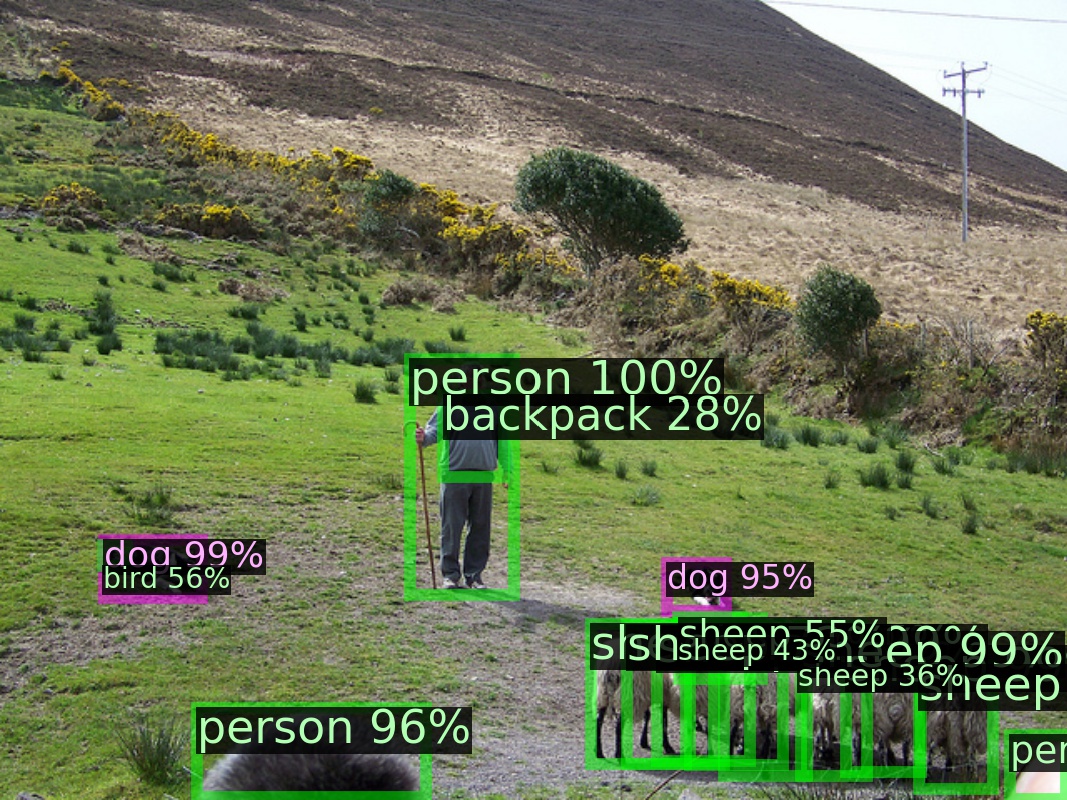}  \\

 \includegraphics[trim={0 0cm 0 0cm}, clip, width=30mm]{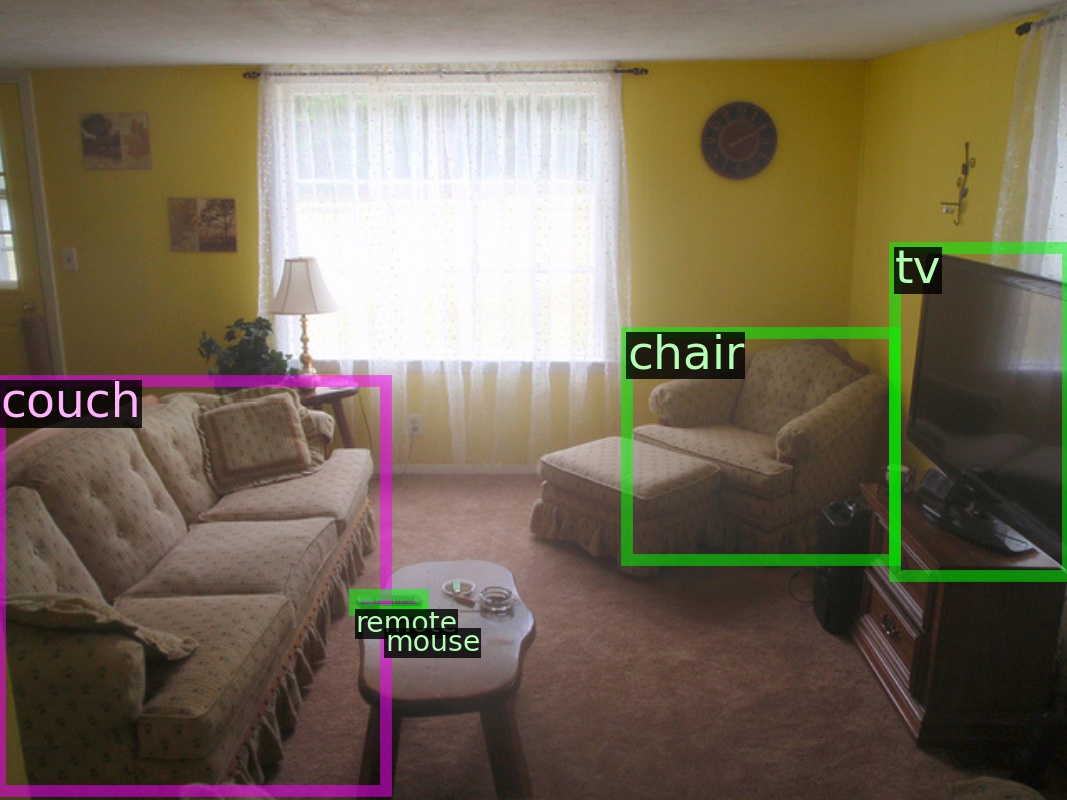} &
 \includegraphics[trim={0 0cm 0 0cm}, clip, width=30mm]{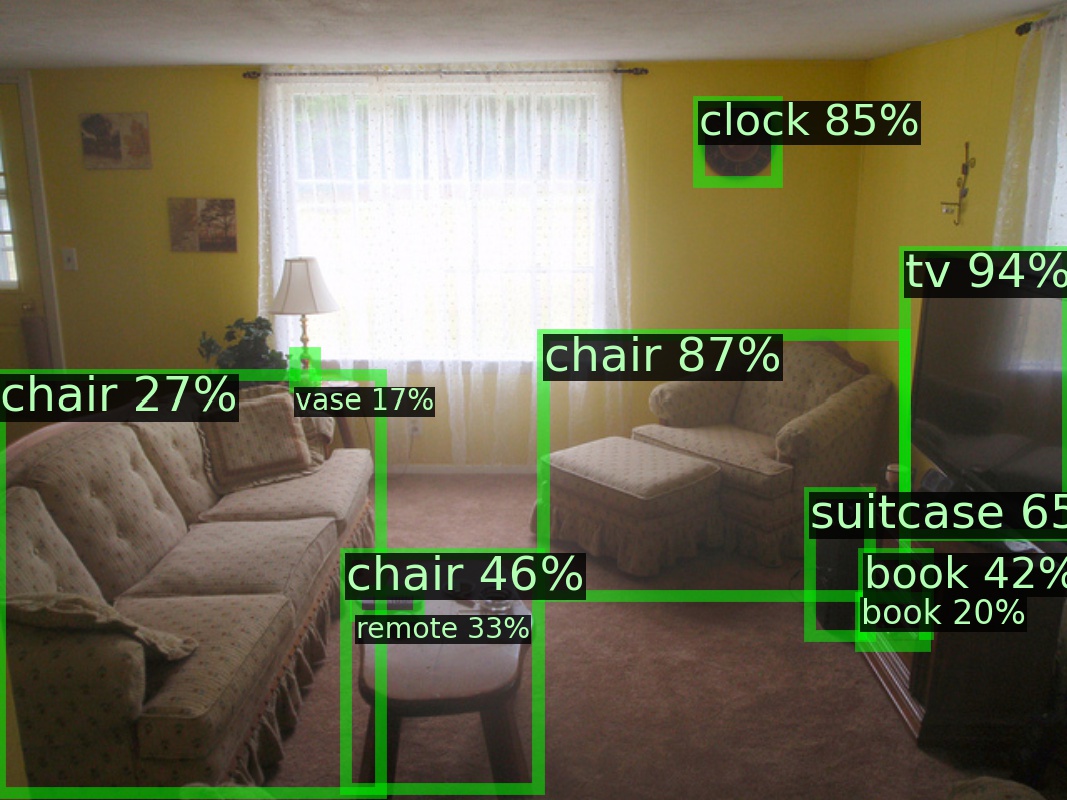} &
 \includegraphics[trim={0 0cm 0 0cm}, clip, width=30mm]{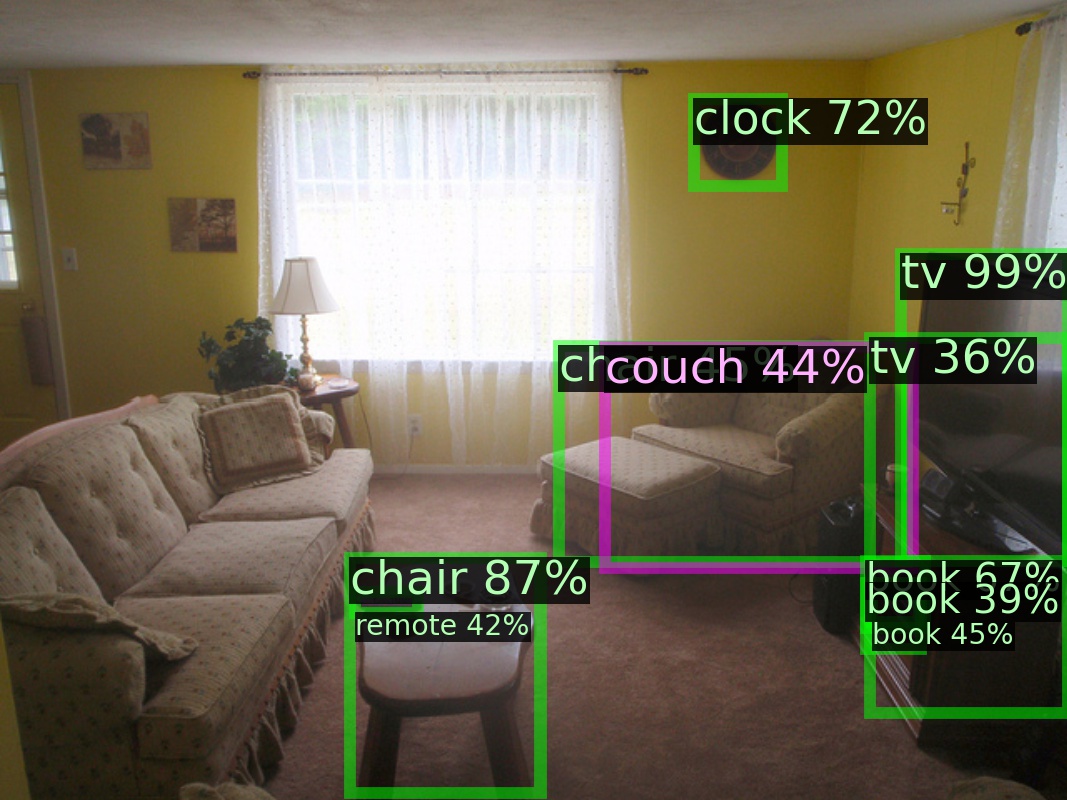} &
 \includegraphics[trim={0 0cm 0 0cm}, clip, width=30mm]{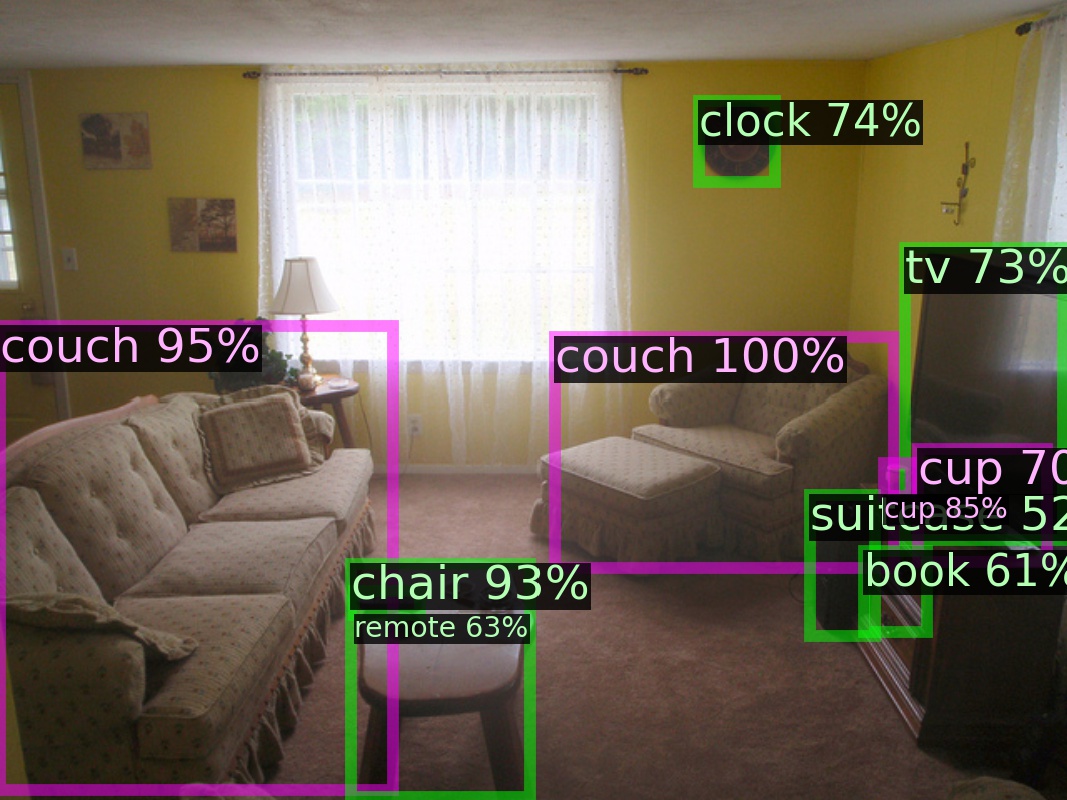}  \\

 \includegraphics[trim={0 0cm 0 0cm}, clip, width=30mm]{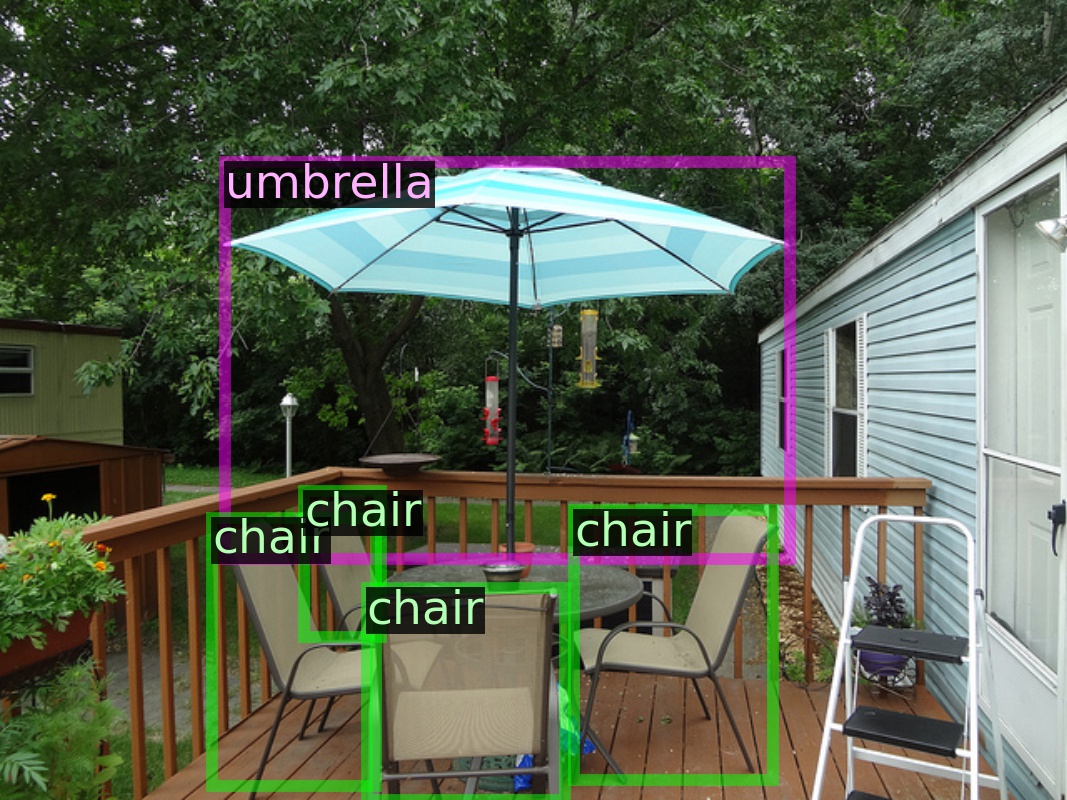} &
 \includegraphics[trim={0 0cm 0 0cm}, clip, width=30mm]{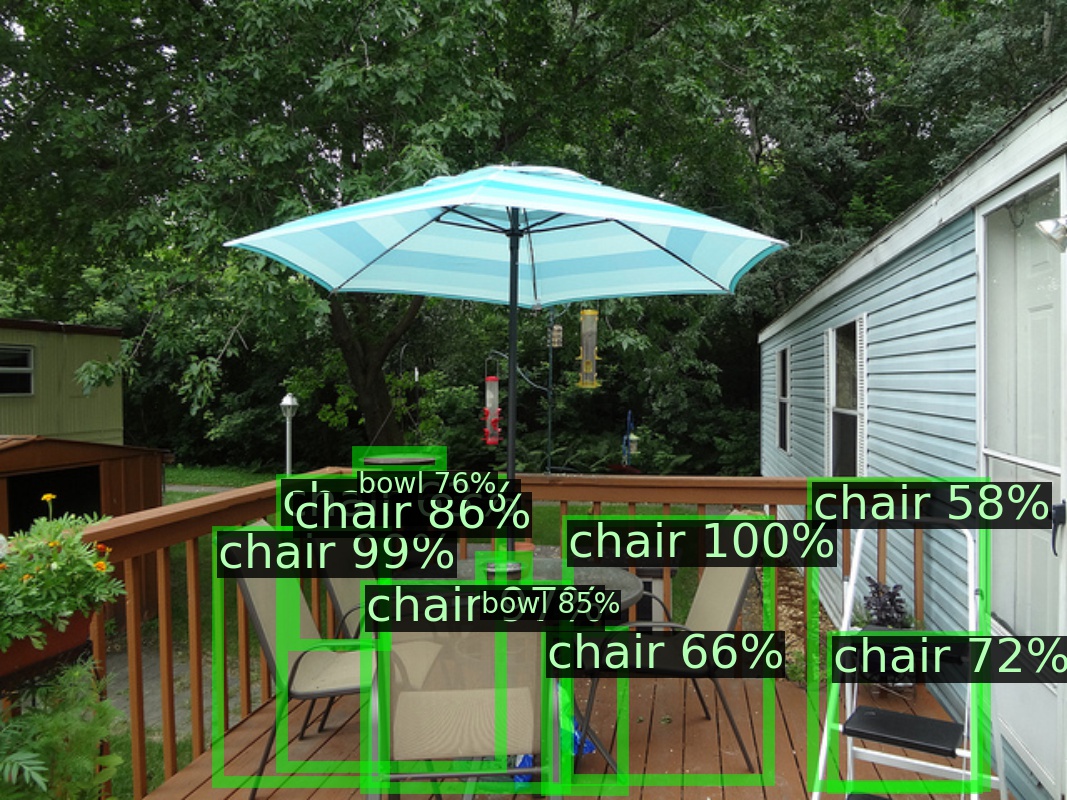} &
 \includegraphics[trim={0 0cm 0 0cm}, clip, width=30mm]{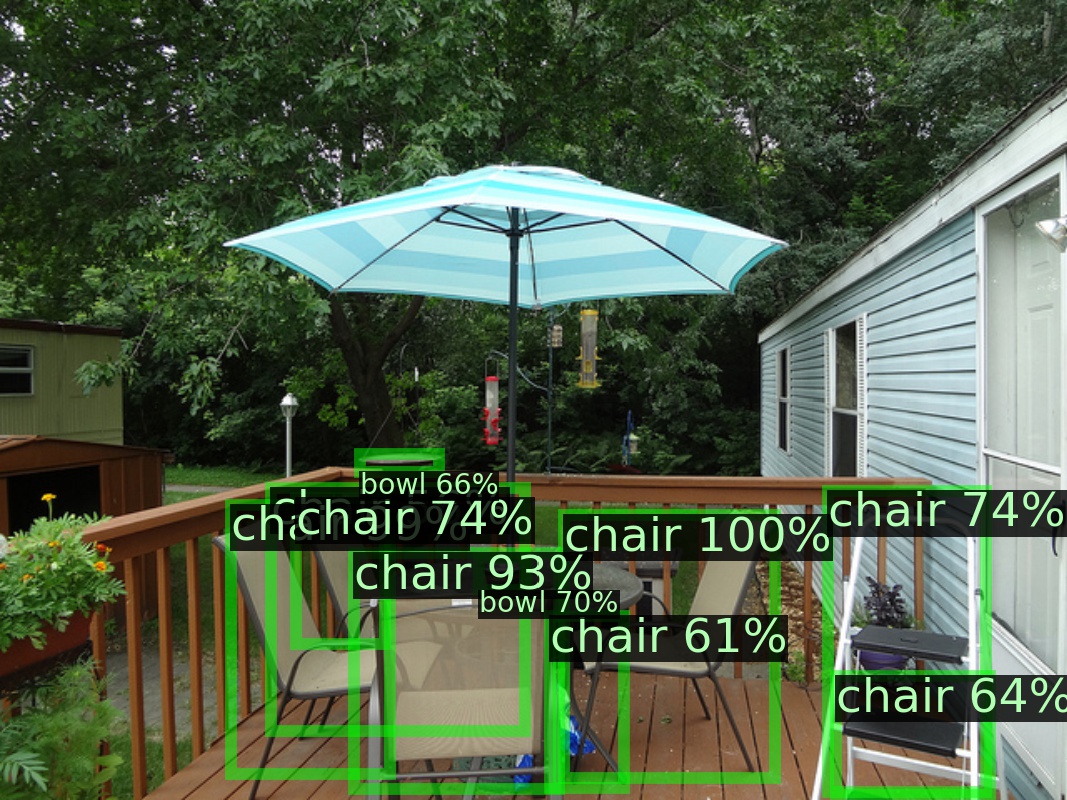} &
 \includegraphics[trim={0 0cm 0 0cm}, clip, width=30mm]{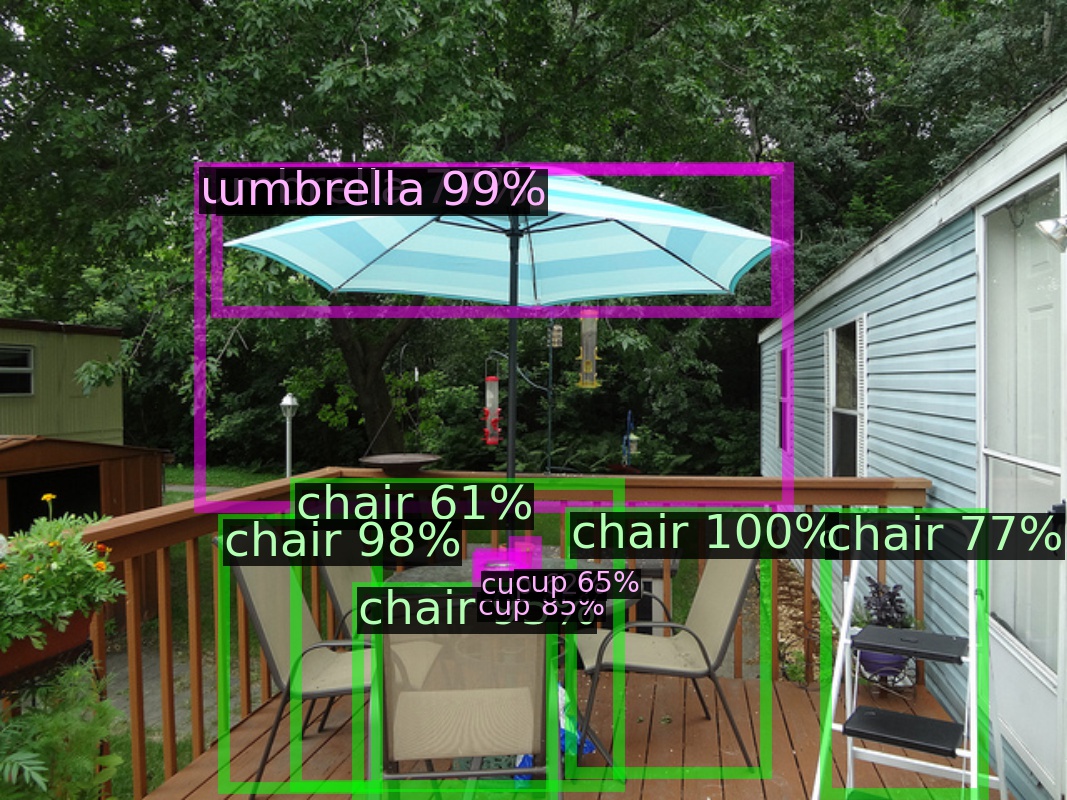}  \\
\end{tabular}
\caption{Qualitative results for open-vocabulary object detection on MSCOCO dataset. Novel classes are shown in \textcolor{magenta}{magenta} while known are in \textcolor{green}{green}. Methods compared are described in Section \ref{sec:baselines}. (Best viewed in color)}
\label{fig:coco_qual_results}
\end{figure}
\subsection{Baselines} \label{sec:baselines}
\textbf{OVR}. The main baseline approach is proposed by Zareian \etal \cite{zareian2021open}. We utilize some components proposed in the work including the two-stage design, grounding loss and usage of a cross-attention model. In this work, we propose new components, which simplify and improve the model performance over OVR. \\
\textbf{STT-ZSD}. Our second baseline uses only the Specialized Task Tuning stage. This resembles a zero-shot object detection setting. 
The model is initialized with ImageNet~\cite{5206848} weights with a trainable projection layer.\\
\textbf{Zero-shot methods}. We compare to some zero-shot object detection approaches which do not include the weak supervision provided by the captions. We compare to three background-aware zero-shot detection methods, introduced by Bansal \etal \cite{bansal2018zero}, which project features of an object bounding box proposal method to word embeddings. The \textbf{SB} method includes a fixed vector for the background class in order to select which bounding boxes to exclude during the object classification, \textbf{LAB} uses multiple latent vectors to represent the different variations of the background class, and \textbf{DSES} includes more classes than the known set as word embedding to train in a more dense semantic space. \textbf{DELO}~\cite{Zhu_2020_CVPR} method uses a generative model and unknown classes to synthesize visual features and uses them while training to increase background confidence. \textbf{PL}~\cite{Rahman_Khan_Barnes_2020} work deals with the imbalance between positive vs. negative instance ratio by proposing a method that maximizes the margin between foreground and background boxes. \\
\textbf{Faster R-CNN}. We also compare with training the classical Faster R-CNN model only using the known classes.\\
\textbf{Open-vocabulary with large data}. We compare our method with recent open-vocabulary models. RegionClip \cite{zhong2021regionclip} uses the CLIP \cite{radford2021learning} pre-trained model to produce image-region pseudo labels and train an object detector. CLIP (cropped reg)~\cite{gu2022openvocabulary} uses the CLIP pre-trained model on 400M image-caption pairs on object proposals obtained by an object detector trained on known classes. XP-Mask \cite{huynh2021open} learns a class-agnostic region proposal and segmentation model from the known classes and then uses this model as a teacher to generate pseudo masks for self-training a student model. Finally, we also compare with VILD~\cite{gu2022openvocabulary} which uses CLIP soft predictions to distil semantic information and train an object detector.

\subsection{Results}
\begin{table}[t]
\footnotesize
\centering
\caption{Different image regions for the LSM stage. $R^I_{grid}$- grid-regions, $R^I_{box}$- proposed box-regions and $R^I_{ann}$- ground truth box-regions of (k) known or (n) novel objects use during the LSM stage}
\label{tab:reg_ablation}
\begin{tabular}{ |c@{\hspace{1mm}}c@{\hspace{1mm}}c|c@{\hspace{1mm}}c@{\hspace{1mm}}c@{\hspace{1mm}}|c@{\hspace{1mm}}c@{\hspace{1mm}}c@{\hspace{1mm}}|c@{\hspace{1mm}}c@{\hspace{1mm}}c@{\hspace{1mm}}| } 
 \hline 
 \multicolumn{3}{|c|}{Regions} & \multicolumn{3}{|c|}{Novel (17)} & \multicolumn{3}{|c|}{Known (48)} & \multicolumn{3}{|c|}{Generalized}\\ 
 $R^I_{grid}$ & $R^I_{box}$ & $R^I_{ann}$ & AP & AP\textsubscript{50} & AP\textsubscript{75} & AP & AP\textsubscript{50} & AP\textsubscript{75} & AP & AP\textsubscript{50} & AP\textsubscript{75}\\
 \hline
 100 & & k$+$n & 18.2 & 31.6 & 18.2 & 32.5 & 52.7 & 34.0 & 27.9 & 46.0 & 28.8\\ 
  & & k$+$n & 16.3 & 28.4 & 15.9 & 32.9 & 53.1 & 34.9 & 27.6 & 45.3 & 28.8\\ 
 \hline
 100 & 100 & & \textbf{17.2} & \textbf{30.1} & \textbf{17.5} & 33.5 & 53.4 & 35.5 & \textbf{28.1} & \textbf{45.7} & \textbf{29.6}\\
 200 & & &15.5 & 27.1 & 15.4 & 32.2 & 52.1 & 33.9 & 27.1 & 44.5 & 28.2\\ 
 & 200 & & 13.7 & 25.7 & 12.9 & \textbf{34.2} & \textbf{53.8} & \textbf{36.5} & 27.5 & 43.8 & 29.1\\  
 \hline
    \end{tabular}
\end{table}

\textbf{COCO dataset.} Table \ref{tab:coco_sota_gen} shows the comparison of our method with several zero-shot and open-vocabulary object detection approaches. \modelname outperforms previous zero-shot detection methods, which show weak performance on detecting novel objects. In comparison to OVR, we improve by 2.53 AP, 3.4 AP\textsubscript{50} for the novel classes and 3.91 AP, 3.92 AP\textsubscript{50} for the known categories. We observe open-vocabulary methods including OVR and our methods have a trade-off between known and novel class performance. Our method finds a better trade-off as compared to the previous work. It reduces the performance gap on known classes as compared to the Faster R-CNN and improves over the novel classes as compared to all previous works. Our method is competitive with recent state-of-the-art methods which use more than $\sim$700 times more image-captions pairs to train, which makes our method data efficient.


Figure \ref{fig:coco_qual_results} shows some qualitative results of our method compared with the STT-ZSD baseline and OVR. Known categories are drawn in green while novel are highlighted in magenta. The columns correspond to the ground truth, STT-ZSD, OVR and our method from left to right.
\modelname is able to find novel objects with a high confidence, such as the dogs in the first example, the couch in the second and the umbrella in the third one. We observe that our method sometimes misclassifies objects with plausible ones, such as the case of the chair in the second example which shares a similar appearance to a couch. 
These examples show a clear improvement of our approach, over the other methods.
In the supplementary material we include some examples of our method showing the limitations and main cause of errors of \modelname.

\subsection{Ablation Experiments} \label{sec:ablations}


\begin{table}[t]\footnotesize
    \centering
    \caption{Ablation study showing the contribution of our proposed consistency-regularization term ($\mathcal{L}_{Cons}$) and usage of BERT text embeddings on COCO validation set. We compared using frozen pretrained weights (fz) of the language model and embedding, fine-tuning (ft) or training from scratch}
    \label{tab:cons_bert_ablation_ext}
\begin{tabular}{ |@{\hspace{1mm}}c@{\hspace{1mm}}c@{\hspace{1mm}}c@{\hspace{1mm}}|c@{\hspace{1mm}}c@{\hspace{1mm}}c@{\hspace{1mm}}|c@{\hspace{1mm}}c@{\hspace{1mm}}c@{\hspace{1mm}}|c@{\hspace{1mm}}c@{\hspace{1mm}}c@{\hspace{1mm}}| } 
 \hline
 \multirow{2}{*}{$\mathcal{L}_{Cons}$} & BERT & BERT & \multicolumn{3}{|c|}{Novel (17)} & \multicolumn{3}{|c|}{Known (48)} & \multicolumn{3}{|c|}{Generalized} \\
  & Model & Emb. & AP & AP\textsubscript{50} & AP\textsubscript{75} & AP & AP\textsubscript{50} & AP\textsubscript{75} & AP & AP\textsubscript{50} & AP\textsubscript{75} \\
 \hline
 \checkmark &  & fz &  \textbf{17.2} & \textbf{30.1} & \textbf{17.5} & \textbf{33.5} & 53.4 & \textbf{35.5} & 28.1 & 45.7 & \textbf{29.6}\\
 \checkmark & fz & fz & 16.7 & 29.7 & 16.7 & 33.4 & \textbf{53.5} & \textbf{35.5} & \textbf{28.2} & \textbf{45.9} & 29.5\\ 
 \checkmark &  & ft & 16.9 & 29.5 & 16.9 & 33.4 & 53.0 & 35.4 & 28.1 & 45.7 & 29.4 \\ 
 \checkmark &  & scratch & 16.0 & 28.3 & 16.2 & 30.4 & 49.6 & 31.8 & 25.8 & 42.9 & 26.6 \\ 
  & & fz & 15.4 & 27.9 & 15.2 & 32.2 & 52.1 & 34.1 & 26.3 & 43.6 & 27.3\\ 
 \hline
    \end{tabular}
\end{table}
\textbf{Localized objects matter.} Table \ref{tab:reg_ablation} presents the impact of using box- vs grid-region features in the LSM stage. 
We compare our method using grid-region features $R^I_{grid}$, proposed box-region features $R^I_{box}$, and using box-region features from the known ($k$) or novel ($n$) class annotations $R^I_{ann}$.
We find that the combination of grid- and box-regions proves to be best, showing a complementary behavior. 
We also considered two oracle experiments (row 1 and 2) using ground-truth box-region features from both known and novel class annotations instead of proposed box-region features.
The best performance is achieved when combined with additional grid regions (row 1). The additional grid-regions help in capturing the background objects beyond the annotated classes while box-regions focus on foreground objects, which improves the image-caption matching.




\textbf{Consistency loss and text embedding selection.} Table \ref{tab:cons_bert_ablation_ext}, shows the contribution of our consistency-regularization term. We get an improvement of 1.76 AP by introducing our consistency loss.
We compare the performance of using a pre-trained text embedding module vs learning it from scratch, fine-tuning it or considering the complete contextualized language model during the LSM stage in Table \ref{tab:cons_bert_ablation_ext}. Using the pre-trained text embedding, results in a better model.
We find out that using only the embeddings module is sufficient and better than using the complete contextualized BERT language model for the task of object detection. We argue that this is because objects are mostly represented by single word vectors, using simple disentangled text embeddings is better suited for generating object class features. In the supplementary material we include an ablation study showing that both stages of training are necessary and complementary for the success of \modelname. 


Table \ref{tab:ablation_contribution} shows the the improvement in performance for each of our contributions. Our baseline method is our implementation of OVR~\cite{zareian2021open}. Both the consistency-regularization together with the inclusion of the box-regions gives the most increment in performance for both novel and known classes. Using only the BERT Embeddings improves the novel class performance although it affects the known classes. Overall we can see that the three contributions are complementary and improve the method for open-vocabulary detection.

\begin{table}[t]\footnotesize
    \centering
    \caption{Ablation study showing the contributions \modelname. $\mathcal{L}_{Cons}$ = consistency-regularization, $R^I_{box}$ = inclusion of box-regions together with grid-regions, BERT Emb. only.}
    \label{tab:ablation_contribution}
\begin{tabular}{ |@{\hspace{1mm}}c@{\hspace{1mm}}c@{\hspace{1mm}}c@{\hspace{1mm}}|c@{\hspace{1mm}}c@{\hspace{1mm}}c@{\hspace{1mm}}|c@{\hspace{1mm}}c@{\hspace{1mm}}c@{\hspace{1mm}}|c@{\hspace{1mm}}c@{\hspace{1mm}}c@{\hspace{1mm}}| } 
 \hline
 \multirow{2}{*}{$\mathcal{L}_{Cons}$} & \multirow{2}{*}{$R^I_{box}$} & BERT & \multicolumn{3}{|c|}{Novel (17)} & \multicolumn{3}{|c|}{Known (48)} & \multicolumn{3}{|c|}{Generalized} \\
  &  & Emb. & AP & AP\textsubscript{50} & AP\textsubscript{75} & AP & AP\textsubscript{50} & AP\textsubscript{75} & AP & AP\textsubscript{50} & AP\textsubscript{75} \\
 \hline
 \checkmark & \checkmark & \checkmark &  \textbf{17.2} & \textbf{30.1} & \textbf{17.5} & \textbf{33.5} & 53.4 & \textbf{35.5} & 28.1 & 45.7 & \textbf{29.6}\\
 $\checkmark$ & $\checkmark$ & $\times$ & 16.7 & 29.7 & 16.7 & 33.4 & \textbf{53.5} & \textbf{35.5} & \textbf{28.2} & \textbf{45.9} & 29.5\\ 
 $\checkmark$ & $\times$ & $\checkmark$  & 15.5 & 27.1 & 15.4 & 32.2 & 52.1 & 33.9 & 27.1 & 44.5 & 28.2\\ 
 $\times$ & $\checkmark$ & $\checkmark$ & 15.4 & 27.9 & 15.2 & 32.2 & 52.1 & 34.1 & 26.3 & 43.6 & 27.3\\ 
 $\times$ & $\times$ & $\times$ & 14.3 & 25.6 & 14.4 & 28.1 & 47.8 & 29.3 & 23.7 & 40.9 & 24.5\\ 
 \hline
    \end{tabular}
\end{table}


\section{Conclusion}\label{sec:conclusion}
In this work, we proposed an image-caption matching method for open-vocabulary object detection. We introduced a localized matching technique to learn improved labels of novel classes as compared to only using grid features. We also showed that the language embedding model is preferable over a complete language model, and proposed a regularization approach to improve cross-modal learning.
In conjunction, these components yield favorable results compared to previous open-vocabulary methods on COCO and VAW benchmarks, particularly considering the much lower amount of necessary data to learn from.

\section*{Acknowledgement}
This work was supported by Deutscher Akademischer Austauschdienst - German Academic Exchange Service (DAAD) Research Grants - Doctoral Programmes in Germany, 2019/20; grant number: 57440921.

%
%
%
%

\clearpage  
\newpage

\appendix
\section*{Appendix}
\begin{table}[b]\small
    \centering
    \begin{tabular}{ |@{\hspace{1mm}}c@{\hspace{1mm}}|c@{\hspace{1mm}}c@{\hspace{1mm}}c@{\hspace{1mm}}|c@{\hspace{1mm}}c@{\hspace{1mm}}c@{\hspace{1mm}}|c@{\hspace{1mm}}c@{\hspace{1mm}}c@{\hspace{1mm}}| } 
     \hline
     \multirow{2}{*}{Method} & 
     \multicolumn{3}{|c|}{Novel (297)} & \multicolumn{3}{|c|}{Known (1020)} & \multicolumn{3}{|c|}{Generalized (2060)} \\
      & AP & AP\textsubscript{50} & AP\textsubscript{75} & AP & AP\textsubscript{50} & AP\textsubscript{75} & AP & AP\textsubscript{50} & AP\textsubscript{75} \\ 
     \hline
     STT-ZSD \scriptsize{(Ours)}& 0.14 & 0.28 & 0.15 & \textbf{1.33} & \textbf{2.56} & \textbf{1.16} & \textbf{0.95} & \textbf{1.84} & \textbf{0.82} \\ 
     OVR \cite{zareian2021open}& 0.59 & 1.27 & 0.45 & 0.92 & 2.08 & 0.72 & 0.70 & 1.57 & 0.54 \\ 
     \modelname \scriptsize{(Ours)} & \textbf{0.67} & \textbf{1.42} & \textbf{0.59} & 1.21 & 2.31 & 1.11 & 0.91 & 1.77 & 0.81 \\ 
     \hline
    \end{tabular}
    \caption{Comparing open-vocabulary object detection results on the VAW test set.}
    \label{tab:vaw_sota}
\end{table}
\section{VAW dataset}\label{sec:vaw_experiments}
 \textbf{Visual Attributes in the Wild (VAW) dataset \cite{Pham_2021_CVPR}} We use the training, validation and test set of images as defined with the proposed dataset \cite{Pham_2021_CVPR}. The dataset contains 58,565 images for training, 3,317 images for validation, and 10,392 images for testing. We define the splits for known and novel classes taking approximately 20\% of the total classes (2260) to be novel, resulting in 1792 known and 468 novel classes. We make sure that all known and novel classes from COCO split are kept in the same subset for VAW splits. After removing images with no known annotations from the training and splitting into known and novel classes, there are 54,632 images for training spanning over 1790 known classes, 818 known / 200 novel classes for the validation set, and 1020 known / 297 novel classes for the test set.
This dataset is much more challenging as compared to COCO since it contains fine-grained classes with a long-tailed distribution. It not only contains more classes as compared to the COCO benchmark, but also poses additional challenges like plural versions defined as different classes, \ex kites vs kite.
In the LSM phase, we use the captions from \textbf{Visual Genome Region Descriptions \cite{krishnavisualgenome}} which contain 108,077 images with a total of 4,297,502 region descriptions. We combine these region descriptions for every image to have a single caption per image.


\textbf{VAW dataset results.} \modelname successfully generalizes to the VAW benchmark. Table \ref{tab:vaw_sota} shows the comparison of our approach to both STT-ZSD and OVR baselines on the test set.
Our method improves consistently over the other two methods for the novel classes, showing that it can scale to more challenging settings with long-tailed distribution and large number of classes. 

\section{Ablation Experiments}\label{sec:sup_ablation}

\begin{table*}[h]
    \centering
    \caption{Comparison of the different stages of the model on the novel object detection. The table also shows different configurations of model update in the STT stage by freezing parts of the backbone network}
\begin{tabular}{ |c c c@{\hspace{3.5mm}} c@{\hspace{3.5mm}} c|c c c|c c c|c c c| } 
 \hline
 \multirow{2}{*}{LSM} & \multirow{2}{*}{STT} & \multicolumn{3}{c|}{Freezing blocks} & \multicolumn{3}{|c|}{Novel (17)} & \multicolumn{3}{|c|}{Known (48)} & \multicolumn{3}{|c|}{Generalized} \\
   & & 1-4 & 1-3 & 1-2 & AP & AP\textsubscript{50} & AP\textsubscript{75} & AP & AP\textsubscript{50} & AP\textsubscript{75} & AP & AP\textsubscript{50} & AP\textsubscript{75} \\
 \hline
 \checkmark & \checkmark & & & \checkmark & \textbf{17.17} & 30.86 & \textbf{16.78} & 30.79 & 50.68 & 32.21 & \textbf{26.14} & \textbf{43.80} & \textbf{27.05}\\ 
 \checkmark & \checkmark & & \checkmark &  & 16.77 & \textbf{30.91} & 16.24 & 30.10 & 49.71 & 31.14 & 25.44 & 43.14 & 25.98\\ 
 \checkmark & \checkmark & \checkmark & &  & 15.96 & 29.09 & 15.59 & 29.14 & 48.50 & 30.63 & 24.82 & 41.99 & 25.73\\ 
 \checkmark & & & & \checkmark & 0.73 & 1.89 & 0.37 & 0.82 & 2.06 & 0.48 & 0.89 & 2.27 & 0.52\\ 
  & \checkmark & & & \checkmark & 0.21 & 0.31 & 0.21 & \textbf{33.23} & \textbf{53.43} & \textbf{35.03} & 24.38 & 39.19 & 25.72\\ 
 \hline
    \end{tabular}
    \label{tab:stt_ablation_ext}
\end{table*}

\paragraph{Two-stage model performs best.} In Table \ref{tab:stt_ablation_ext}, we show the extended performance of our method using single stage, either LSM or STT, and fine-tuning different sets of the backbone weights during the STT stage. The last two rows of Table \ref{tab:stt_ablation_ext} consider our method using only the STT stage (same as our baseline STT-ZSD from Section 4.2 in the main paper) and using only the LSM stage. Individual stage models are not able to detect novel objects well, which shows that both stages are fundamental for the detection of novel objects. We further compare the performance of different model configurations by freezing different number of blocks of the backbone network during the STT stage. Our results show that only freezing the first two blocks and the projection layer leads to the best configuration for the STT. In conclusion we can observe two main results: first, using both stages is crucial to detect novel objects. Second, freezing the backbone weights of the 1st and 2nd ResNet blocks during the STT stage results in the best configuration for both, novel and known, performances.  

\begin{table}[h]
\small
\centering
\caption{Different image regions for the LSM stage. $R^I_{grid}$- grid-regions, $R^I_{box}$- proposed box-regions and $R^I_{ann}$- ground truth box-regions of (k) known or (n) novel objects use during the LSM stage}
\label{sup:tab:reg_ablation}
\begin{tabular}{ |c@{\hspace{1mm}}c@{\hspace{1mm}}c|c@{\hspace{1mm}}c@{\hspace{1mm}}c@{\hspace{1mm}}|c@{\hspace{1mm}}c@{\hspace{1mm}}c@{\hspace{1mm}}|c@{\hspace{1mm}}c@{\hspace{1mm}}c@{\hspace{1mm}}| } 
 \hline 
 \multicolumn{3}{|c|}{Regions} & \multicolumn{3}{|c|}{Novel (17)} & \multicolumn{3}{|c|}{Known (48)} & \multicolumn{3}{|c|}{Generalized}\\ 
 $R^I_{grid}$ & $R^I_{box}$ & $R^I_{ann}$ & AP & AP\textsubscript{50} & AP\textsubscript{75} & AP & AP\textsubscript{50} & AP\textsubscript{75} & AP & AP\textsubscript{50} & AP\textsubscript{75}\\
 \hline
 100 & & k$+$n & 18.2 & 31.6 & 18.2 & 32.5 & 52.7 & 34.0 & 27.9 & 46.0 & 28.8\\ 
  & & k$+$n & 16.3 & 28.4 & 15.9 & 32.9 & 53.1 & 34.9 & 27.6 & 45.3 & 28.8\\ 
 100 & & k & 14.2 & 26.8 & 13.4 & 30.0 & 50.2 & 31.3 & 24.8 & 42.4 & 25.5\\
 \hline
 100 & 100 & & \textbf{17.2} & \textbf{30.1} & \textbf{17.5} & 33.5 & 53.4 & 35.5 & \textbf{28.1} & \textbf{45.7} & \textbf{29.6}\\
 200 & & &15.5 & 27.1 & 15.4 & 32.2 & 52.1 & 33.9 & 27.1 & 44.5 & 28.2\\ 
 100 & & & 14.9 & 25.8 & 15.0 & 31.7 & 51.8 & 33.3 & 26.6 & 43.9 & 27.7\\ 
 & 200 & & 13.7 & 25.7 & 12.9 & \textbf{34.2} & \textbf{53.8} & \textbf{36.5} & 27.5 & 43.8 & 29.1\\  
 & 100 & & 13.4 & 22.8 & 13.4 & 33.9 & 53.7 & 35.8 & 27.0 & 43.3 & 28.5\\ 
 \hline
    \end{tabular}
\end{table}

\paragraph{Localized objects matter.} Table \ref{sup:tab:reg_ablation} presents the impact of using box- vs grid-region features in the LSM stage. 
We compare our method using grid-region features $R^I_{grid}$, proposed box-region features $R^I_{box}$, and using box-region features from the known ($k$) or novel ($n$) class annotations $R^I_{ann}$.
When training the LSM stage, we only consider a fixed amount of image regions to calculate the losses and drop the rest of the regions. To illustrate that the improvement comes from the combination of grid- and box-regions and not simply from more boxes, we trained with an increased number of image regions (100 and 200) for every case explicitly stated in Table \ref{sup:tab:reg_ablation}. Even though increasing the number of regions results in a better performance the combination of both types of regions proves to be best, showing a complementary behavior. 
We also considered two oracle experiments (row 1 and 2) using ground-truth box-region features from both known and novel class annotations instead of proposed box-region features. These two experiments improve performance on novel classes showing that object-centered box regions are crucial and the best performance is achieved when combined with additional grid regions (row 1). The additional grid-regions help in capturing the background objects beyond the annotated classes while box-regions focus on precise foreground objects, which improves the image-caption matching.

\begin{figure}[t]
\centering
\begin{tabular}{c@{\hspace{1mm}}c@{\hspace{1mm}}c@{\hspace{1mm}}c}
(a) GT  & (b) \modelname & (c) GT  & (d) \modelname   \\[6pt]
\includegraphics[width=0.23\linewidth]{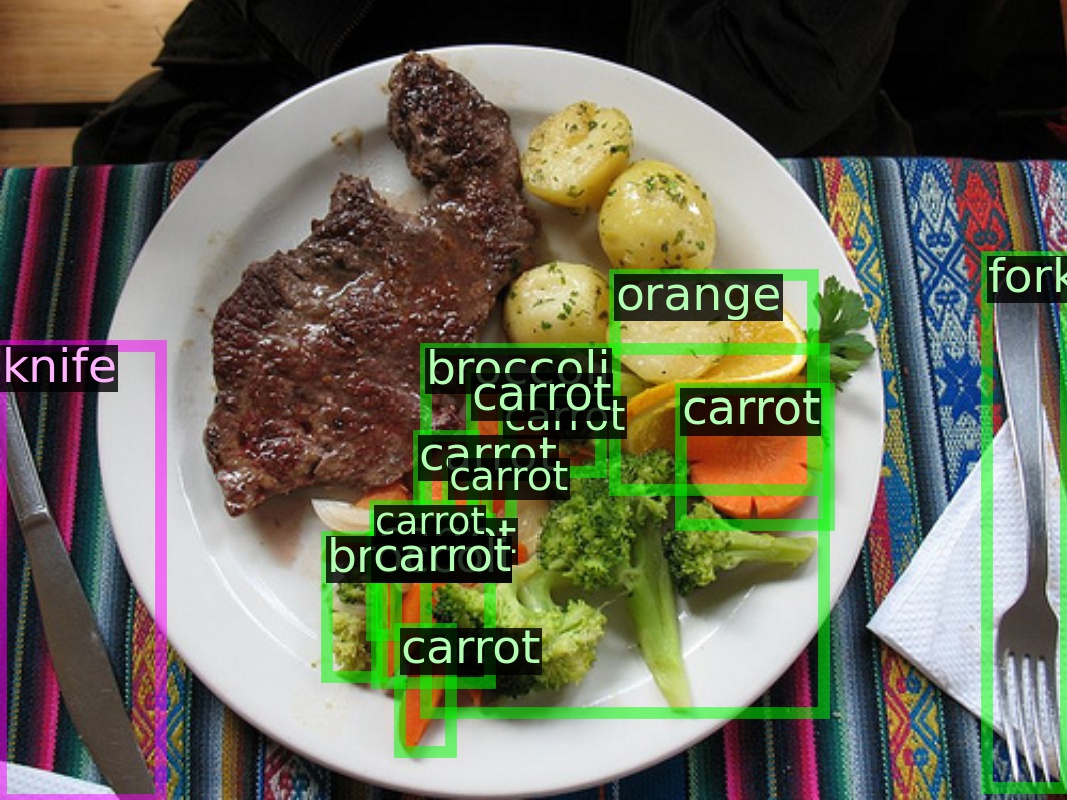} &
\includegraphics[width=0.23\linewidth]{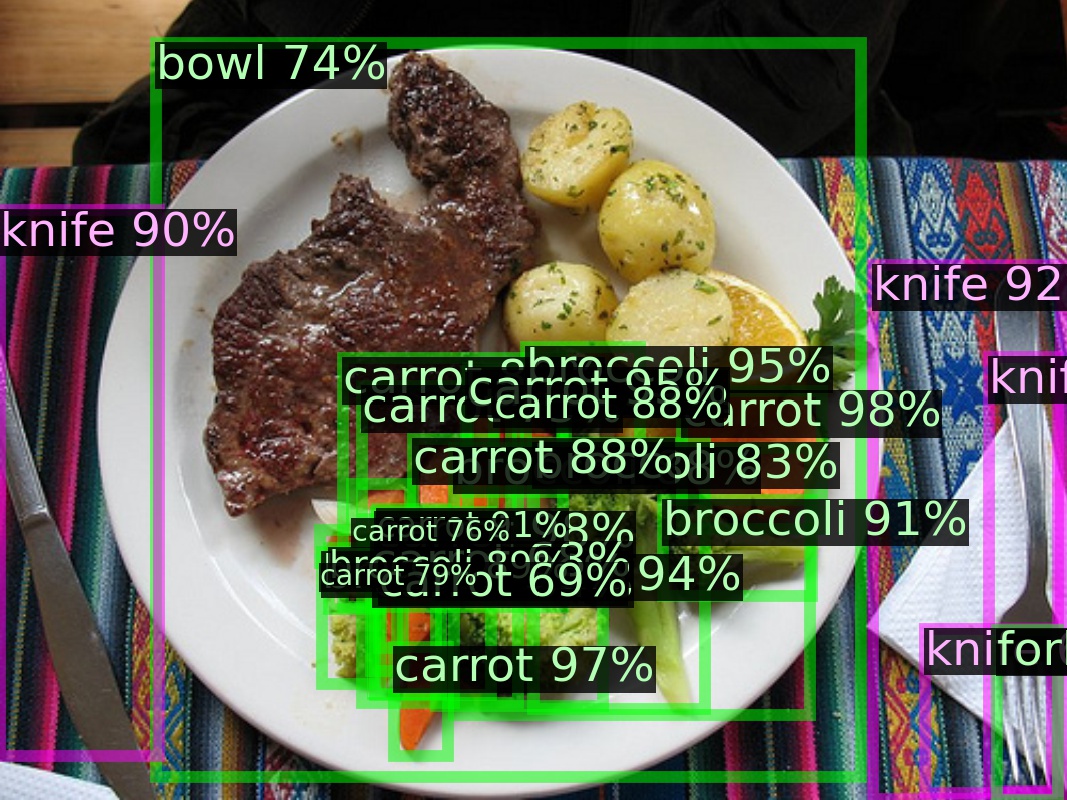} &
\includegraphics[trim={0 0 0 0cm}, clip, width=0.23\linewidth]{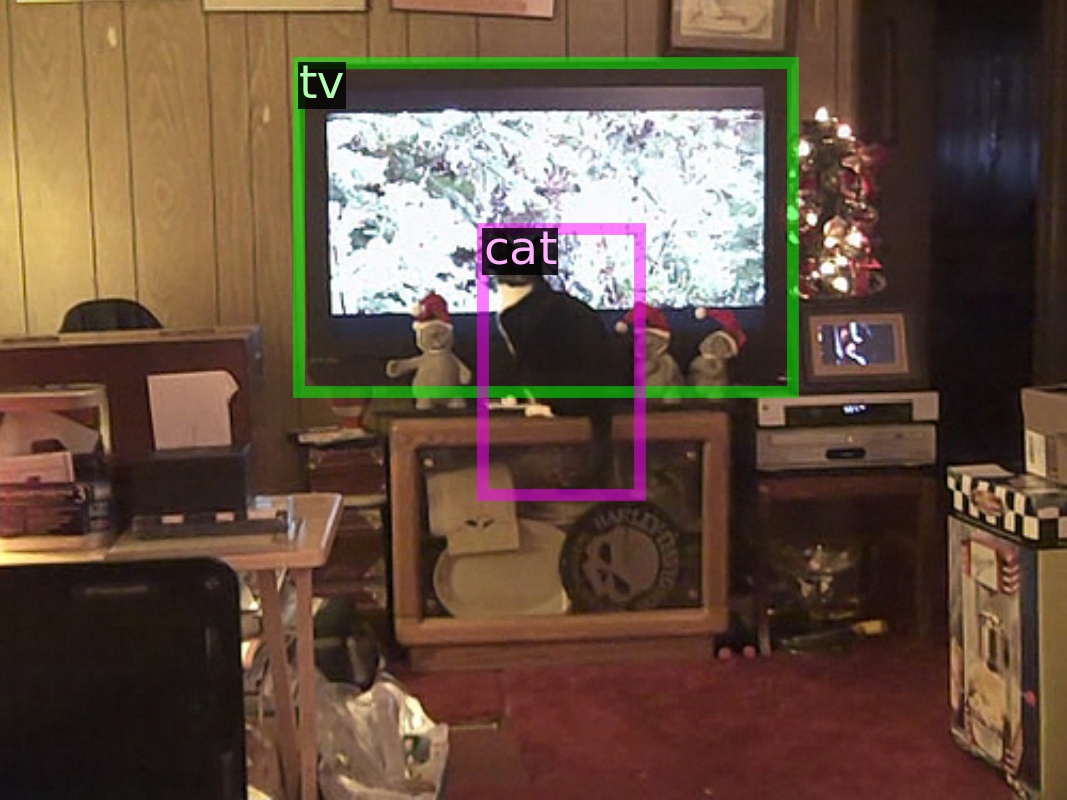} &
\includegraphics[trim={0 0 0 0cm}, clip, width=0.23\linewidth]{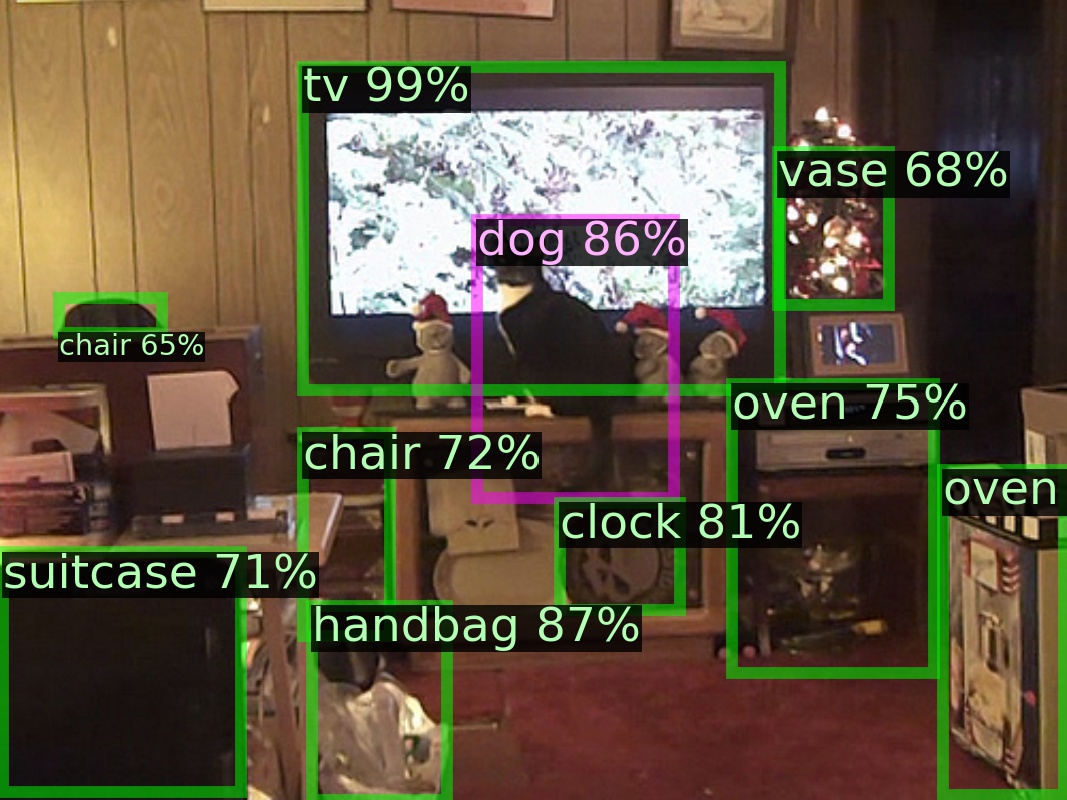}  \\
\end{tabular}
\caption{Failure cases. The method fails to learn fine-grained classification for novel objects. The model confuses between similar classes. For \eg~ the model sometimes predicts `fork' as `knife'(left image) and `cat' as `dog'(right image). }
\label{fig:fail}
\end{figure}
\section{Limitations}\label{sec:limitations}
Visual features of novel object classes are learned during the Localized Semantic Matching stage using image-caption pairs. We notice that such a weak form of supervision is not sufficient to learn fine-grained classification. 
Similar classes such as `dog' and `cat' or `knife' and `fork' are often confused, as shown in Figure \ref{fig:fail}, since they can be used exchangeably in the caption description and they sometimes even co-occur in the image (\eg knife-fork), making the matching process ambiguous.
We also observe a clear drop in performance of known object classes when a similar novel object class is detected. 
A table showing this analysis quantitatively is included in the supplementary.

\section{Per Class Performance}\label{sec:per_class}
\begin{figure}[t]
    \centering
    \includegraphics[trim={0 0cm 0cm 0cm}, clip, width=\textwidth]{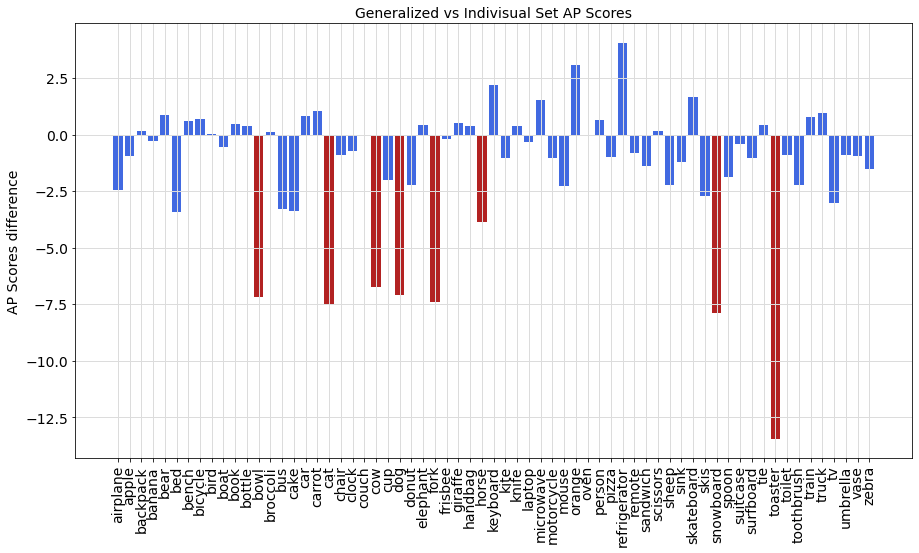}
    \caption{We plot the difference in AP score when considering the generalized setup (all classes together) as compare to considering the individual sets of known and novel separately. Most of the classes present a drop when considering all classes together. Red bars correspond to classes with a drop larger than 3.5 AP.}
    \label{fig:diff_class_scores}
\end{figure}

Figure \ref{fig:diff_class_scores} presents the difference of AP per class when considering the generalized setup, all classes together, minus the AP for the individual setup, only the novel or only the known classes. Most of the scores present a drop when considering the generalized case. Analyzing cases where this drop is larger than 3.5 AP (the red bars in Figure \ref{fig:diff_class_scores}) we can deduce that these classes are mostly confused. Figures \ref{fig:coco_qual_results_extended1} and \ref{fig:coco_qual_results_extended2} show some qualitative examples of our method. We show the ground truth image with annotations and results using our method for comparison. In Figure \ref{fig:coco_qual_results_extended1} we can observe that classes such as bowl and cup are frequently confused, and similar error occurs for classes: fork, knife and spoon. These errors occur due to the fact that these classes look similar or appear together very often. These type of errors are also noticeable between other such classes like cow/sheep/dog and snowboard/skis/skateboard. The class toaster is a special case since it is the class with the least instances present in the dataset (only 9 vs a median of 275), which makes it harder for our method to distinguish this class among the known set and the task becomes harder when considering all 65 classes. 

\section{Qualitative examples}\label{sec:per_class}
\begin{figure*}[t!]
\begin{tabular}{c@{\hspace{1mm}}c@{\hspace{1mm}}c@{\hspace{1mm}}c}
(a) Ground Truth & (b) Our Results & (c) Ground Truth & (d) Our Results  \\[6pt]
 \includegraphics[trim={0 1.5cm 0 0cm}, clip, width=30mm]{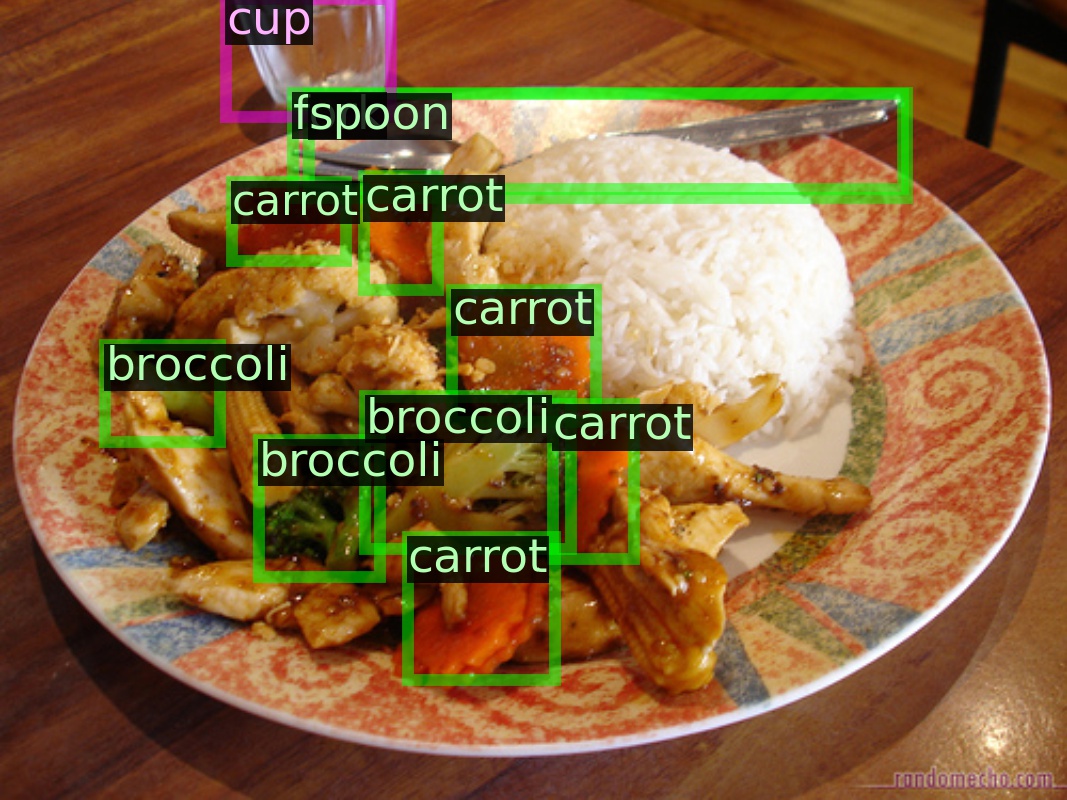} &
 \includegraphics[trim={0 1.5cm 0 0cm}, clip, width=30mm]{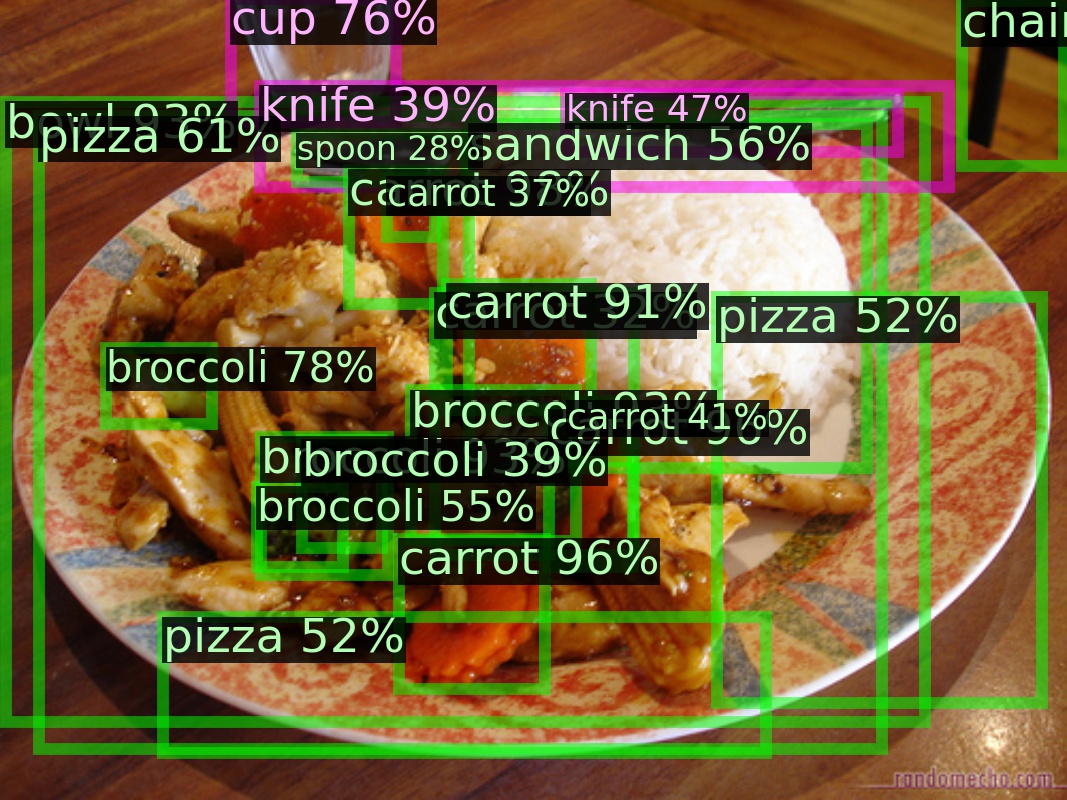} &
 \includegraphics[trim={0 0cm 0 0cm}, clip, width=30mm]{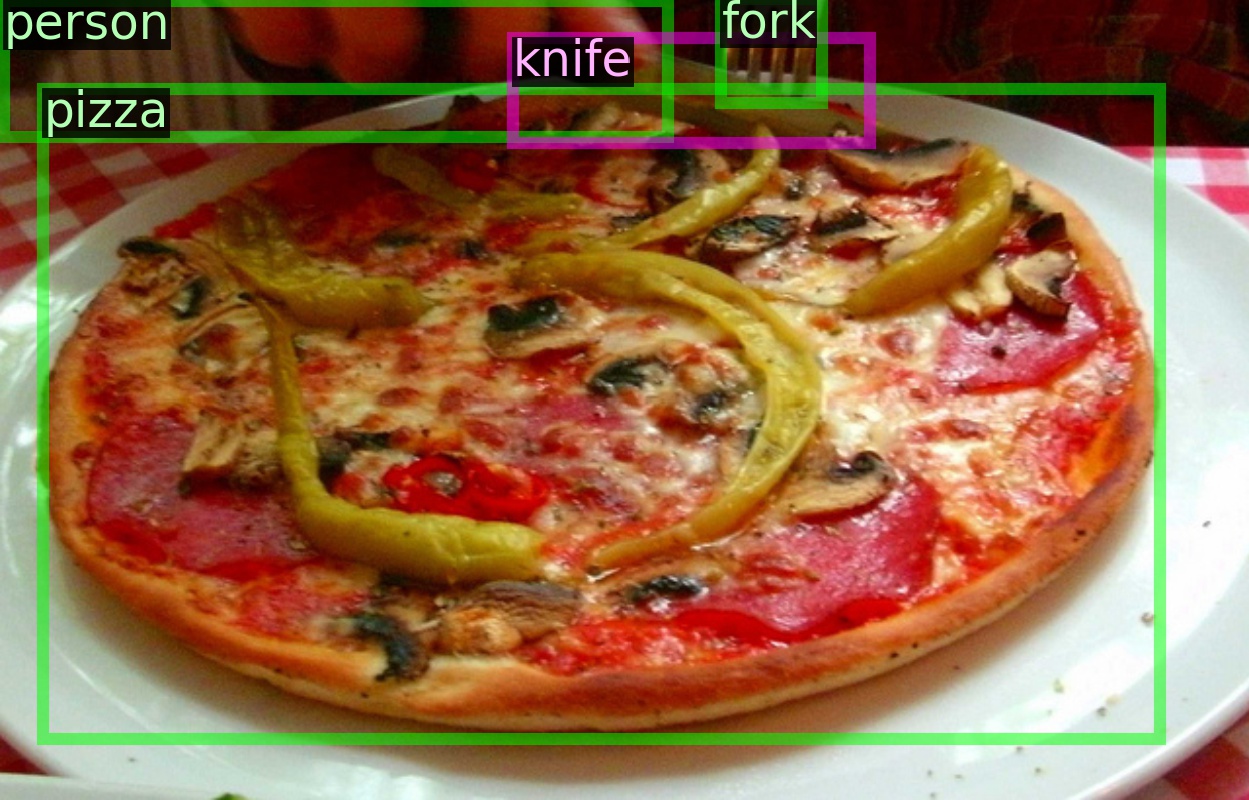} &
 \includegraphics[trim={0 0cm 0 0cm}, clip, width=30mm]{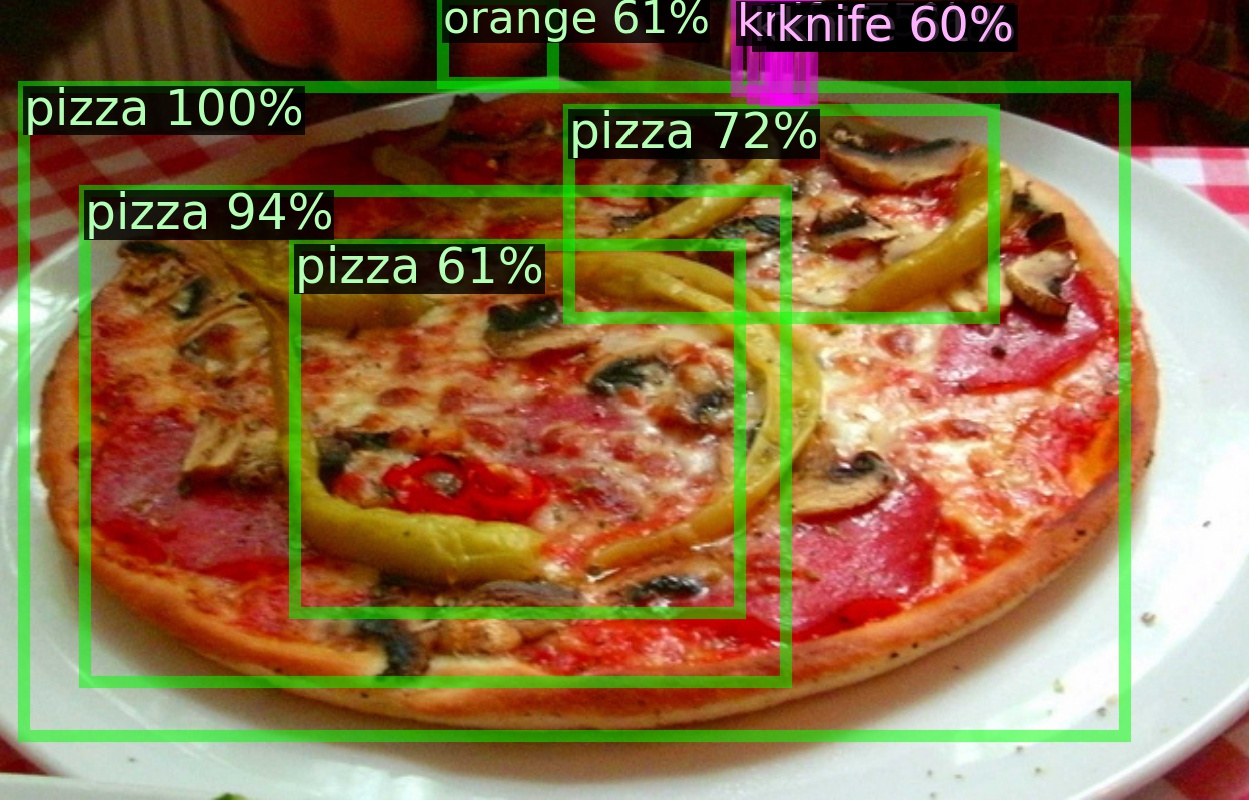} \\
 
 \includegraphics[trim={0 0.5cm 0 1.0cm}, clip, width=30mm]{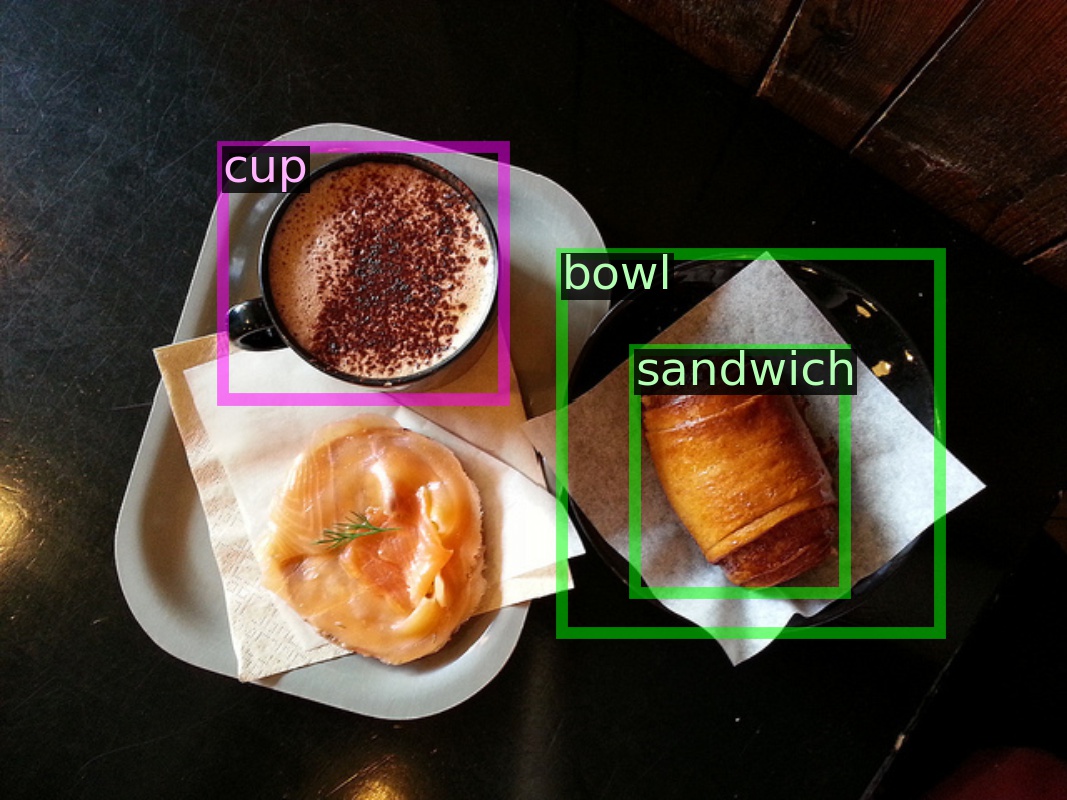} &
 \includegraphics[trim={0 0.5cm 0 1.0cm}, clip, width=30mm]{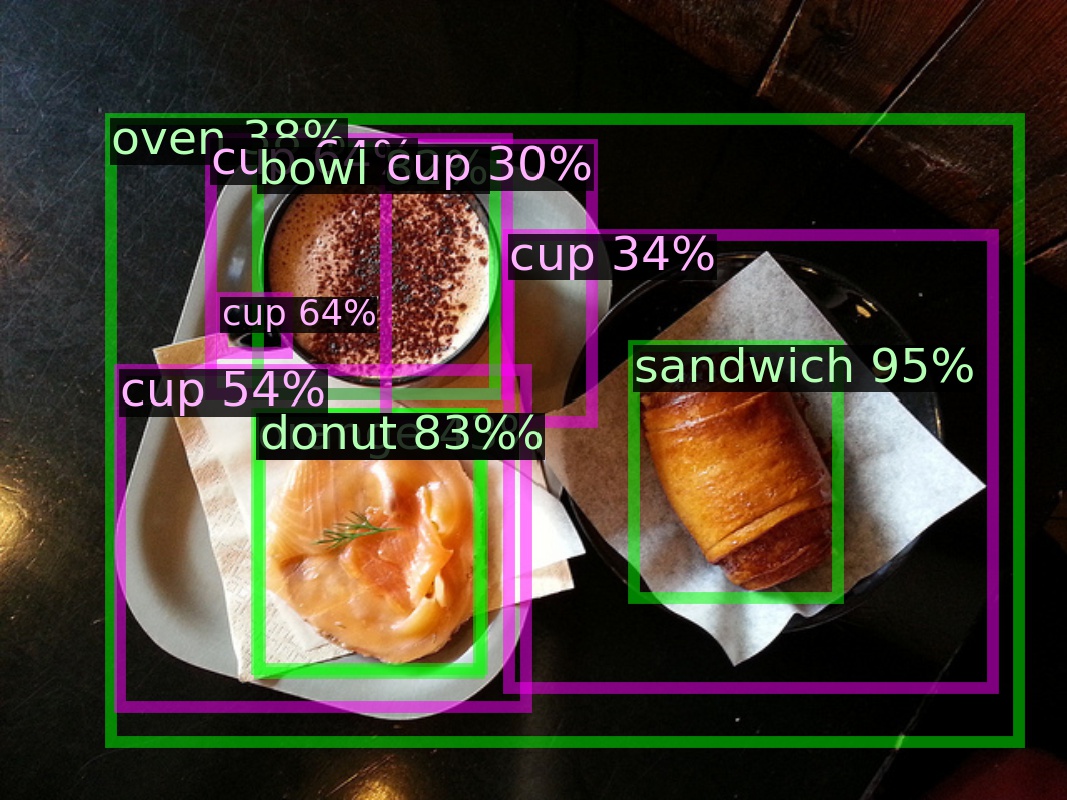} &
 \includegraphics[trim={0 0cm 0 0cm}, clip, width=30mm]{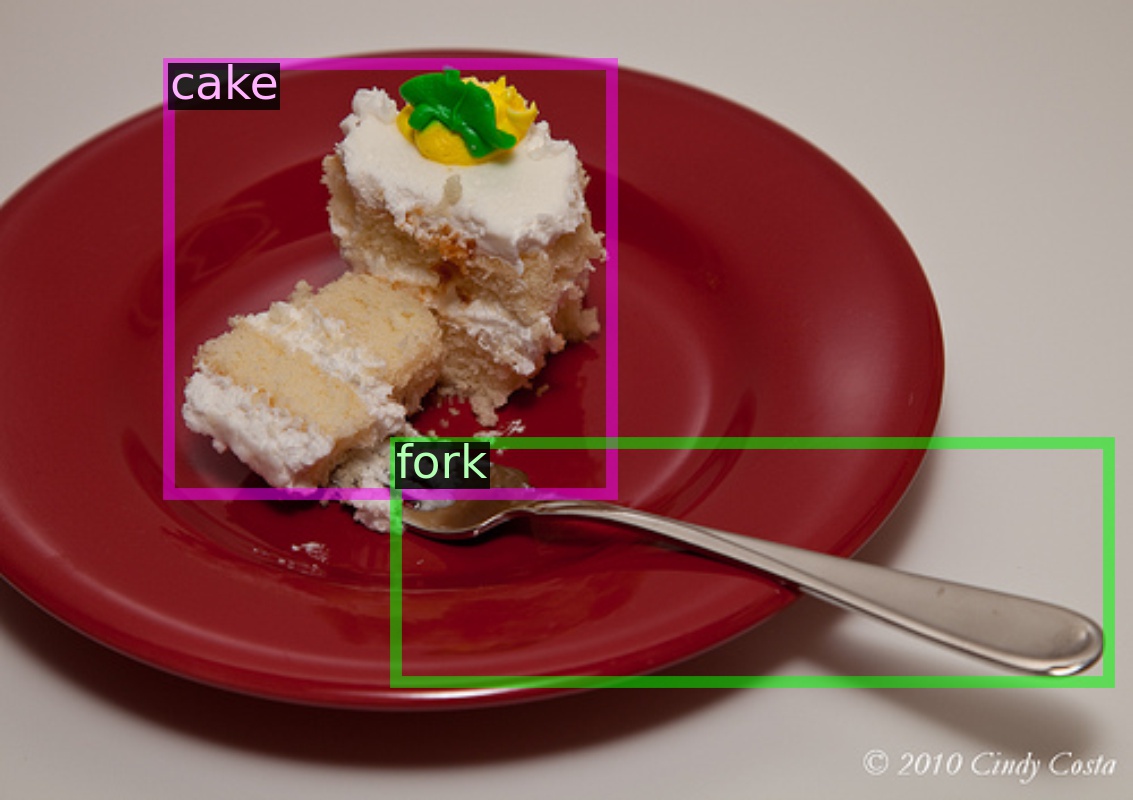} &
 \includegraphics[trim={0 0cm 0 0cm}, clip, width=30mm]{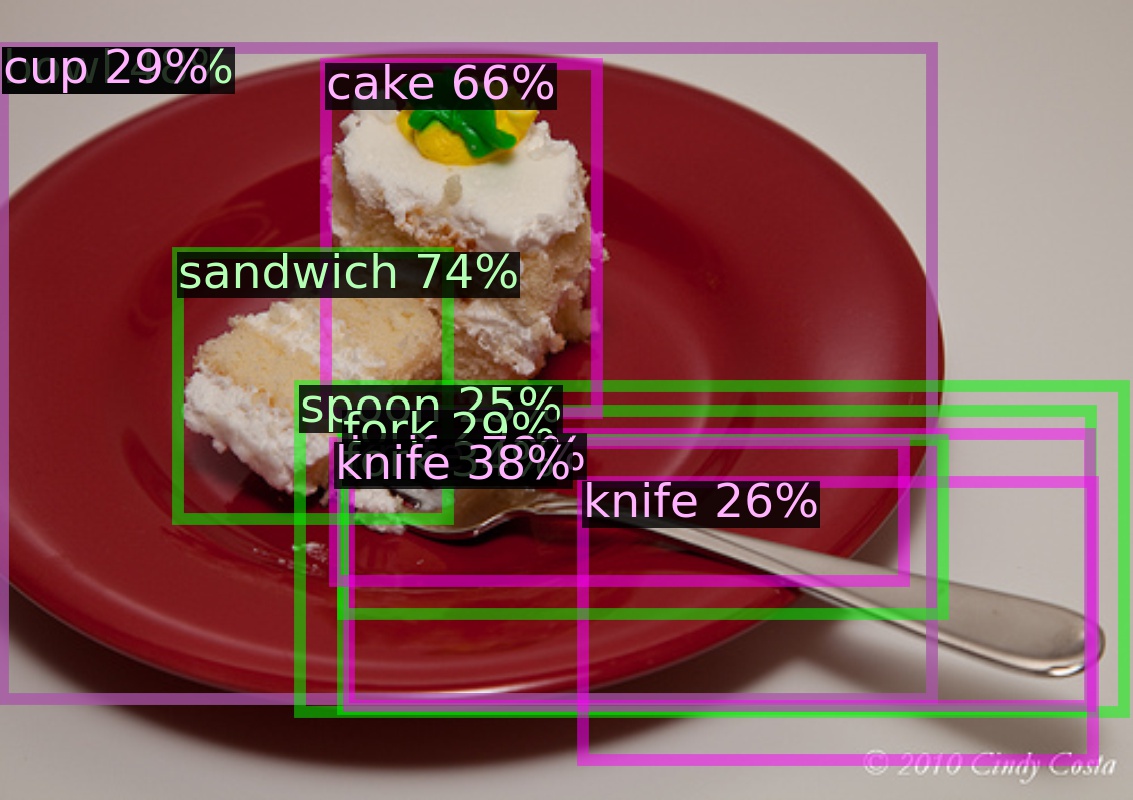}  \\
 
 \includegraphics[trim={0 2.0cm 0 1.5cm}, clip, width=30mm]{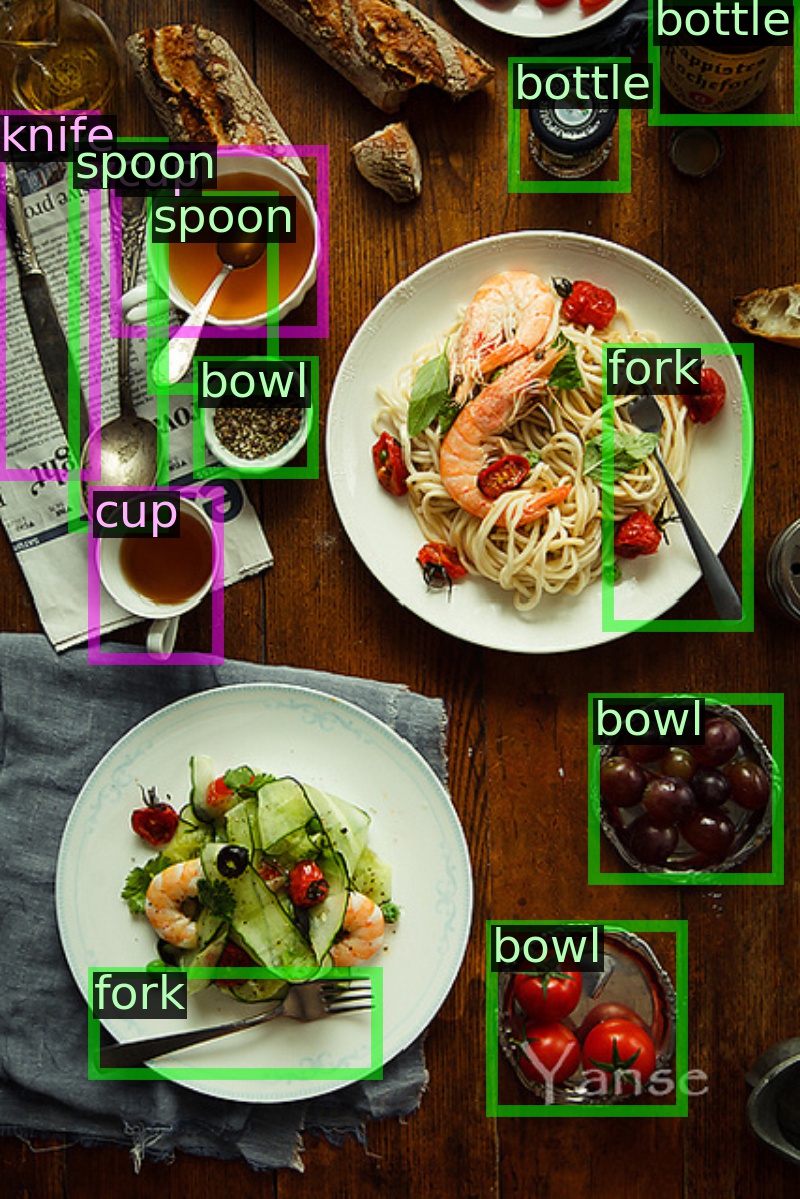} &
 \includegraphics[trim={0 2.0cm 0 1.5cm}, clip, width=30mm]{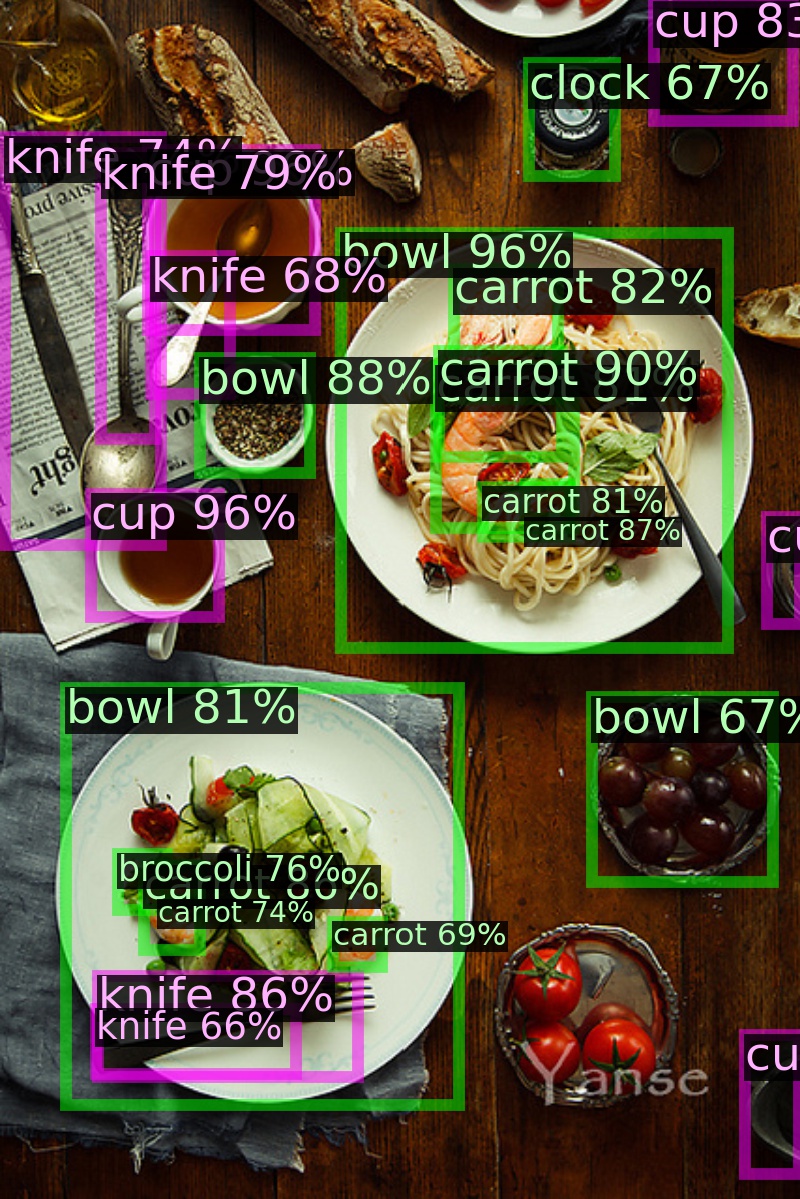} &
 \includegraphics[trim={0 0cm 0 0cm}, clip, width=30mm]{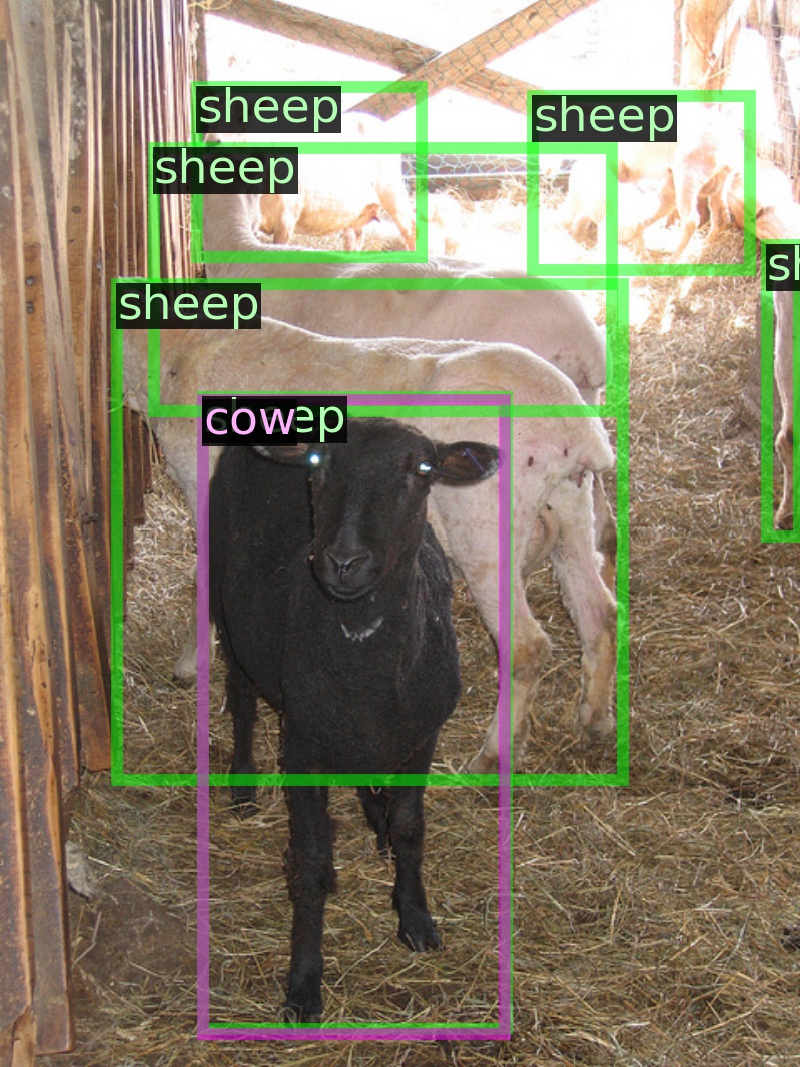} &
 \includegraphics[trim={0 0cm 0 0cm}, clip, width=30mm]{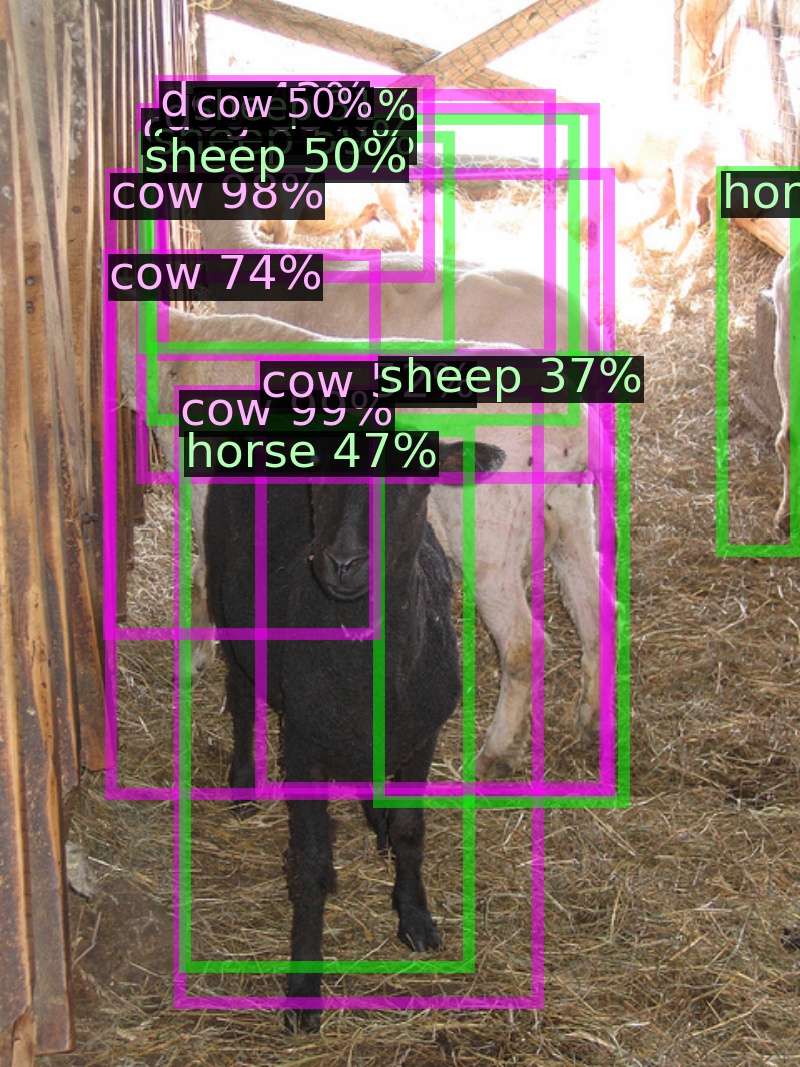}  \\
 
 \includegraphics[trim={0 0cm 0 0cm}, clip, width=30mm]{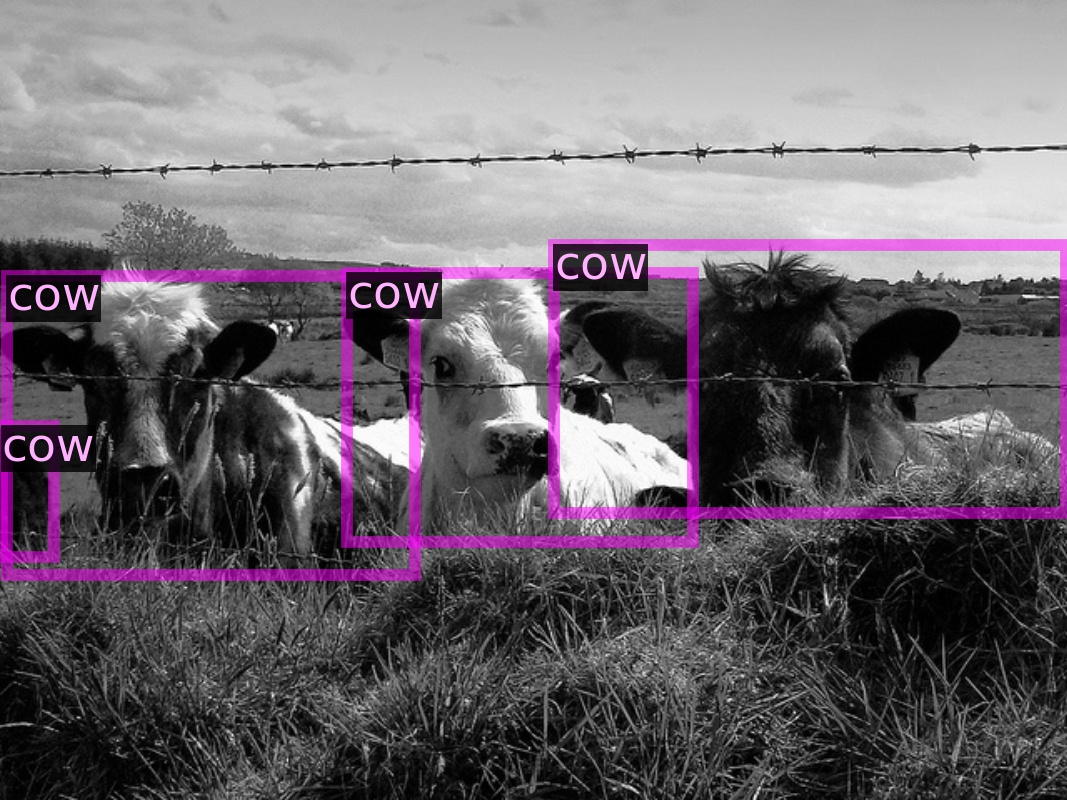} &
 \includegraphics[trim={0 0cm 0 0cm}, clip, width=30mm]{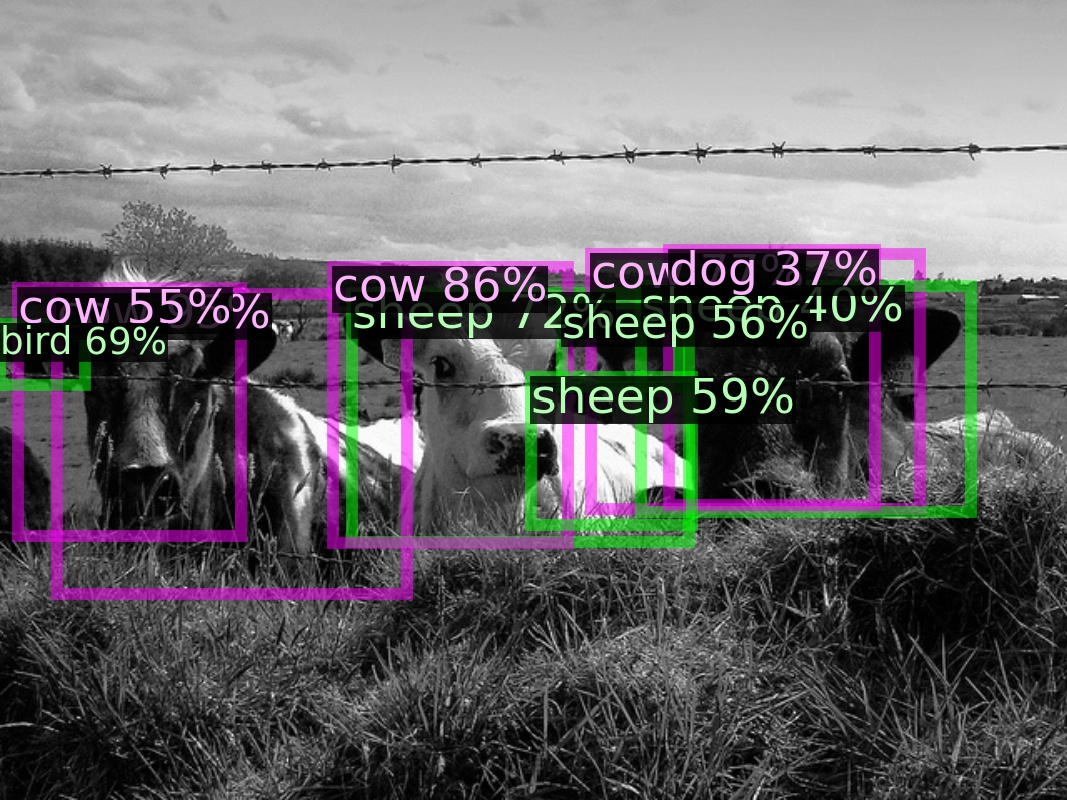} &
 \includegraphics[trim={0 0cm 0 0cm}, clip, width=30mm]{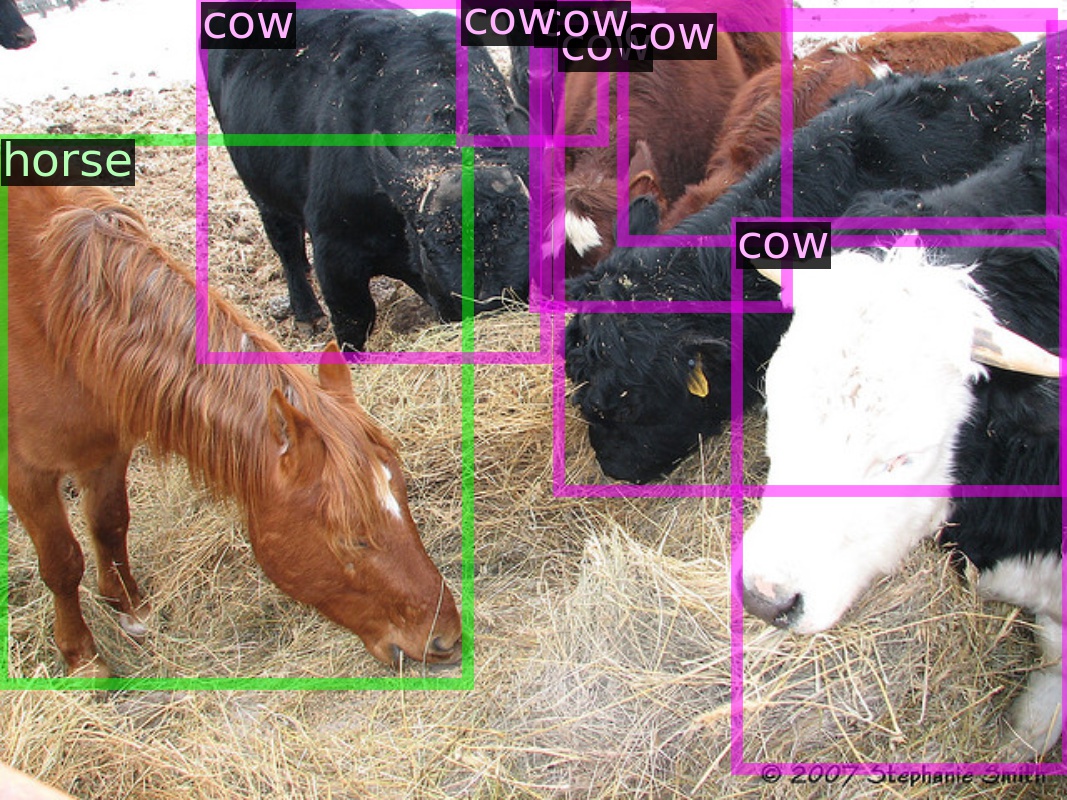} &
 \includegraphics[trim={0 0cm 0 0cm}, clip, width=30mm]{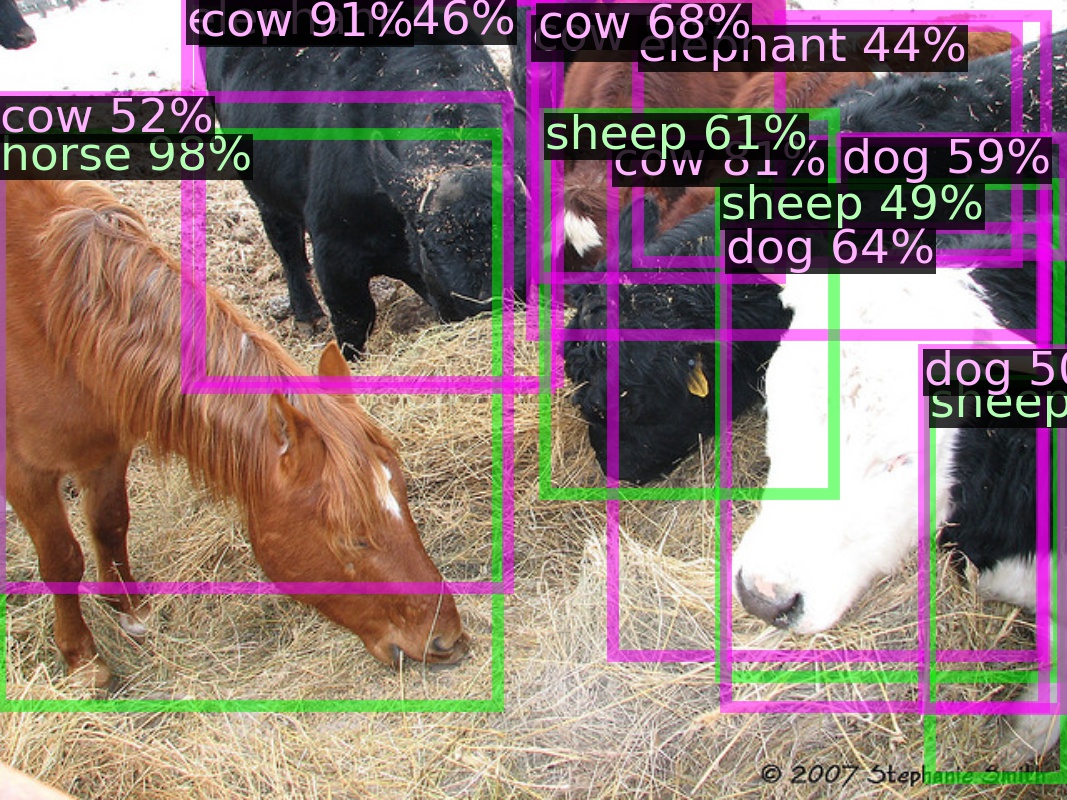} \\
 
 \includegraphics[trim={0 0cm 0 2.0cm}, clip, width=30mm]{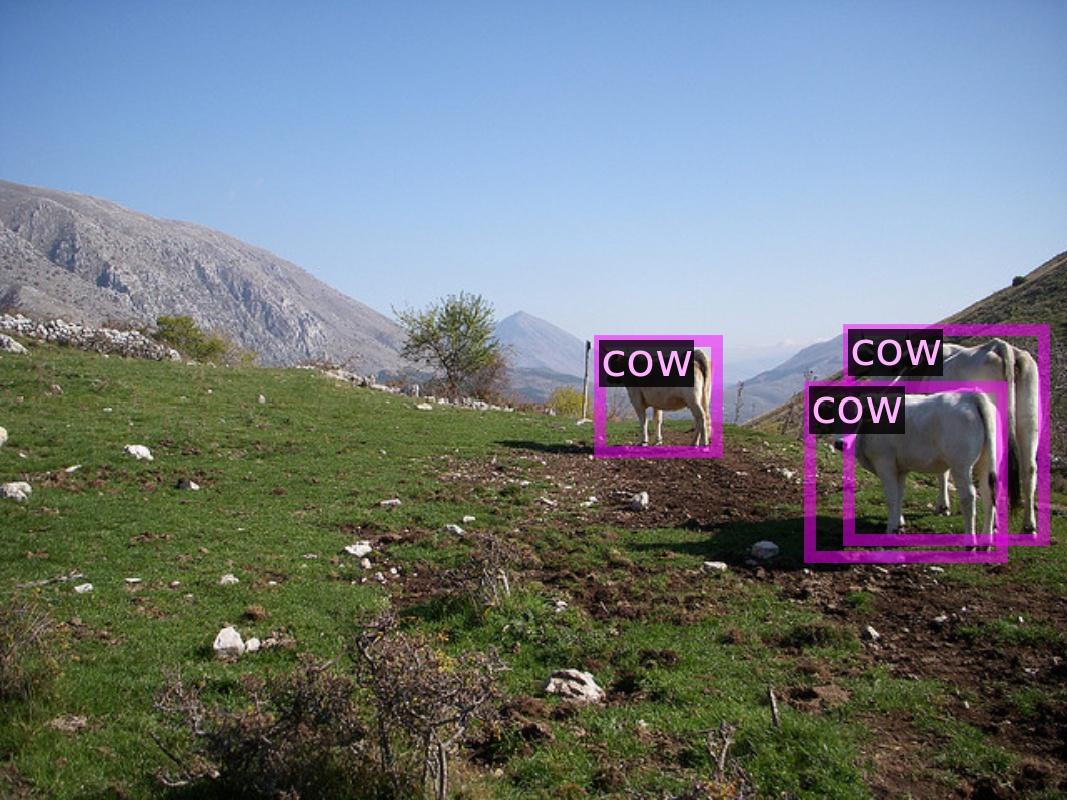} &
 \includegraphics[trim={0 0cm 0 2.0cm}, clip, width=30mm]{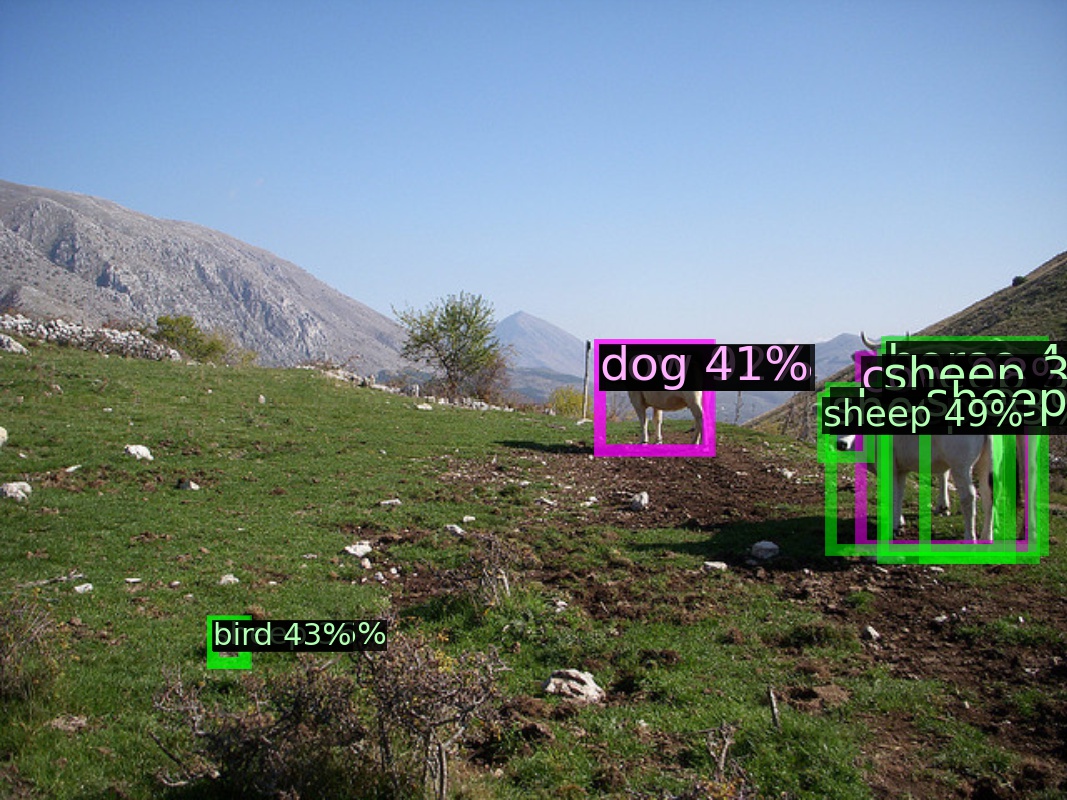} &
 \includegraphics[trim={0 0cm 0 0cm}, clip, width=30mm]{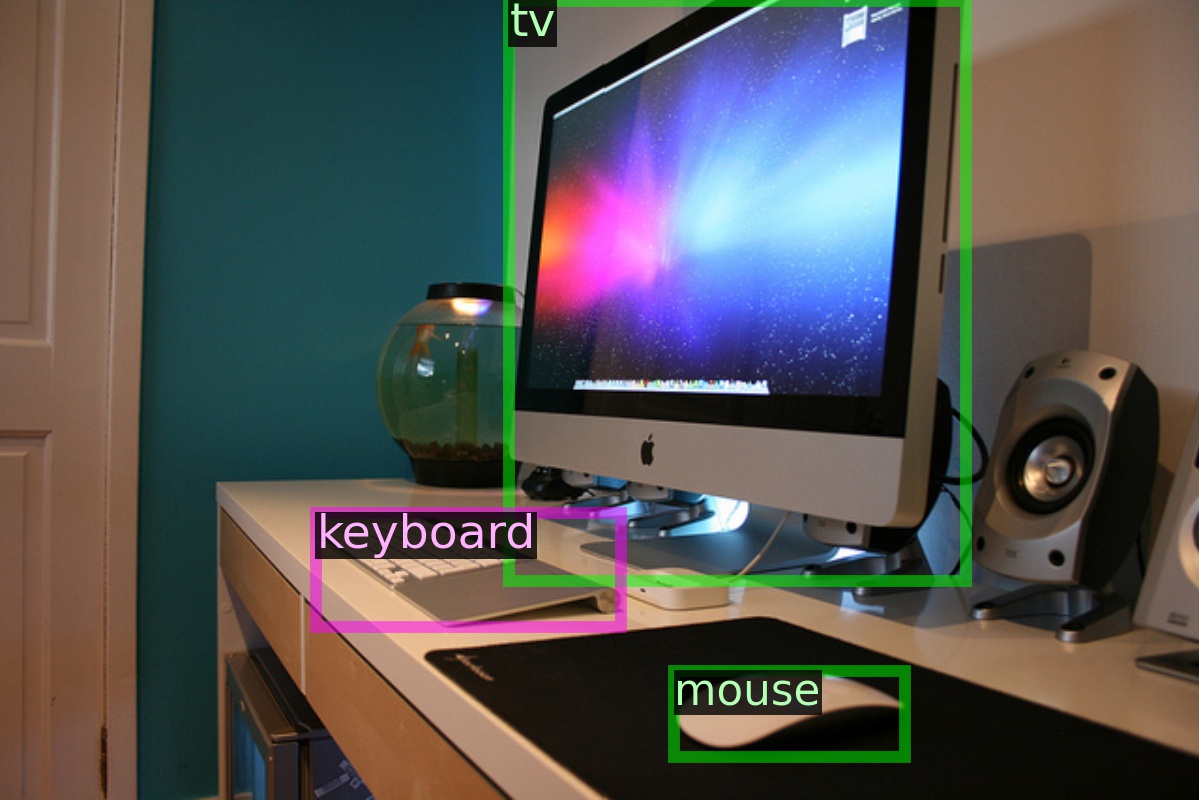} &
 \includegraphics[trim={0 0cm 0 0cm}, clip, width=30mm]{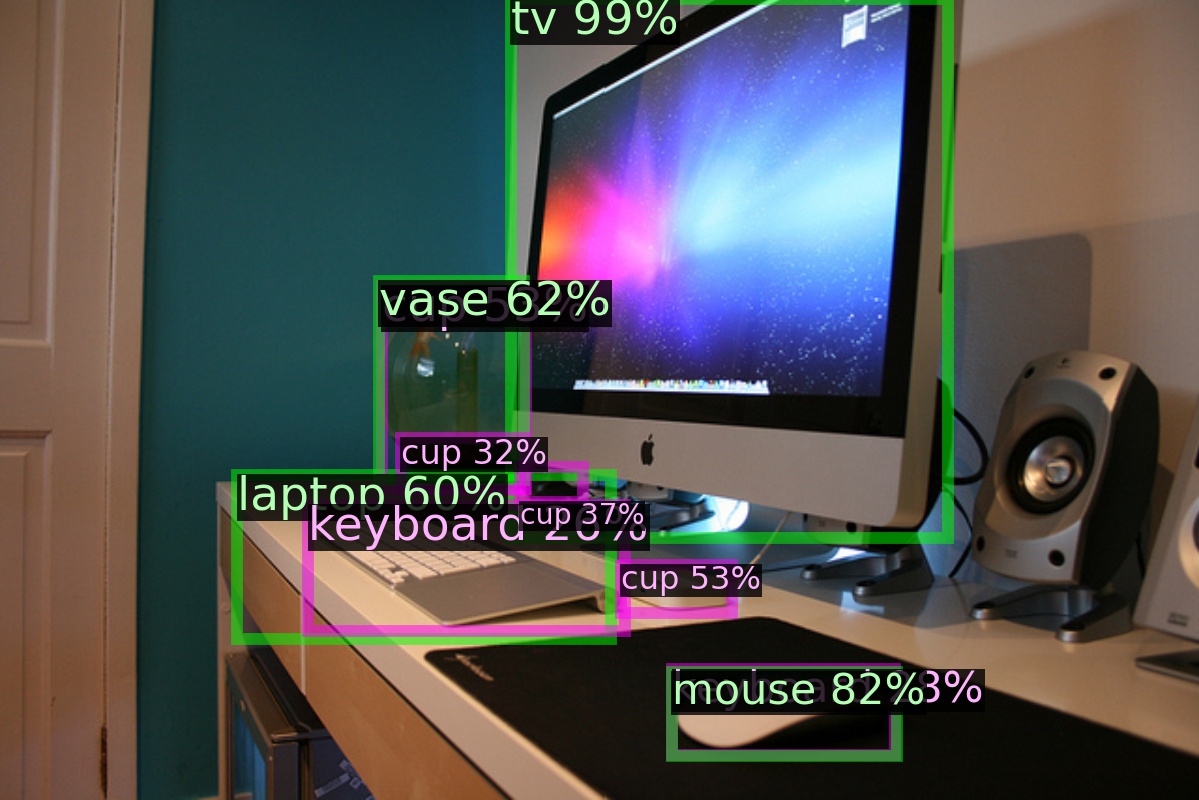}  \\

\end{tabular}
\caption{Qualitative results obtained using \modelname on the COCO dataset. Novel classes are shown in \textcolor{magenta}{magenta} while known are in \textcolor{green}{green}.  (Best viewed in color)}
\label{fig:coco_qual_results_extended1}
\end{figure*}

\begin{figure*}[h!]
\begin{tabular}{c@{\hspace{1mm}}c@{\hspace{1mm}}c@{\hspace{1mm}}c}
(a) Ground Truth & (b) Our Results & (c) Ground Truth & (d) Our Results  \\[6pt]
 \includegraphics[trim={0 0cm 0 2.0cm}, clip, width=30mm]{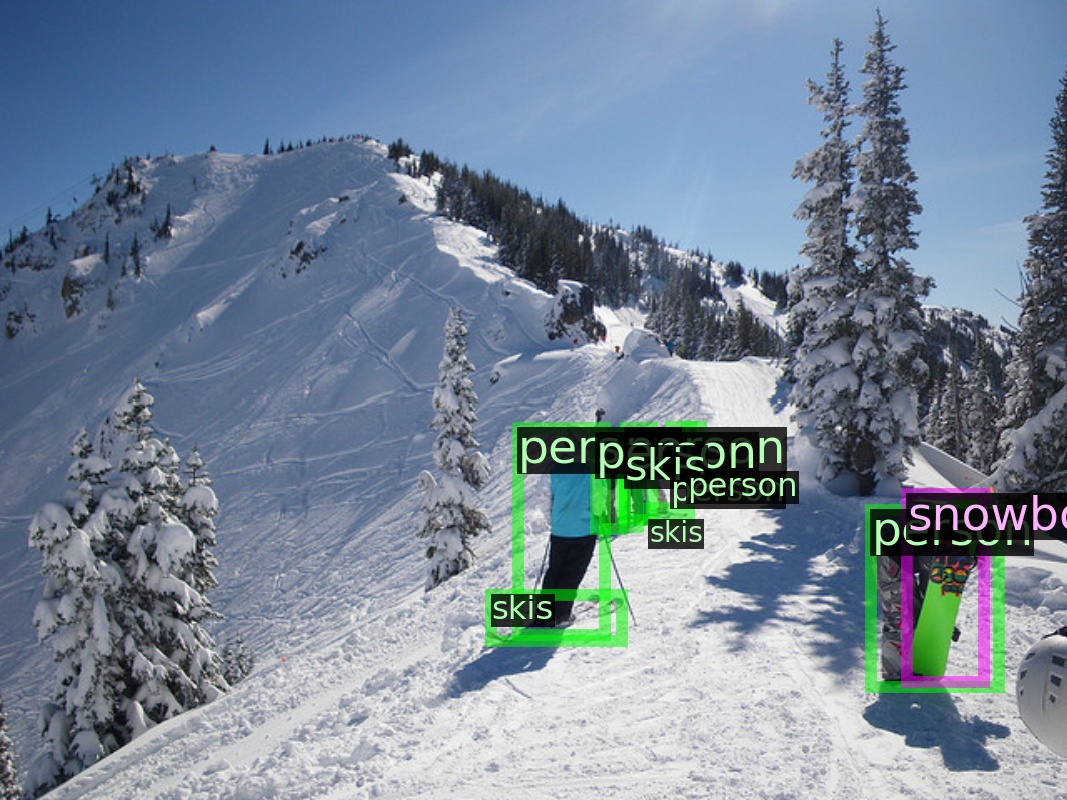} &
 \includegraphics[trim={0 0cm 0 2.0cm}, clip, width=30mm]{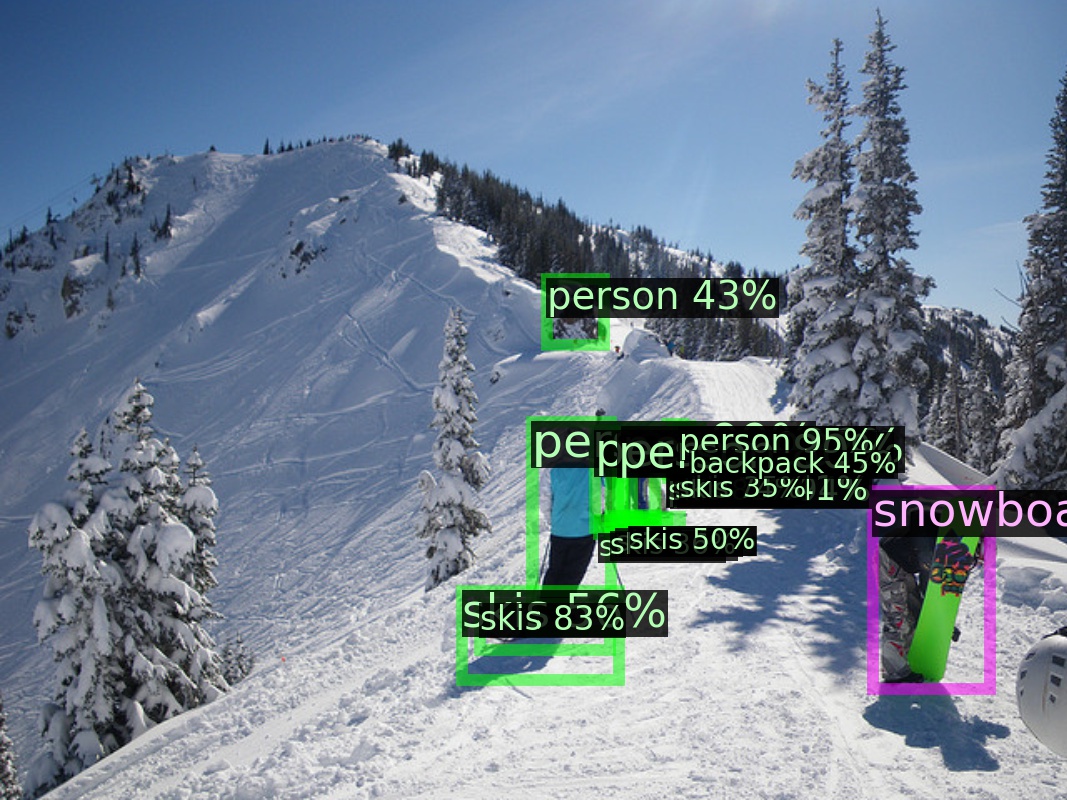} &
 \includegraphics[trim={0 0cm 0 0cm}, clip, width=30mm]{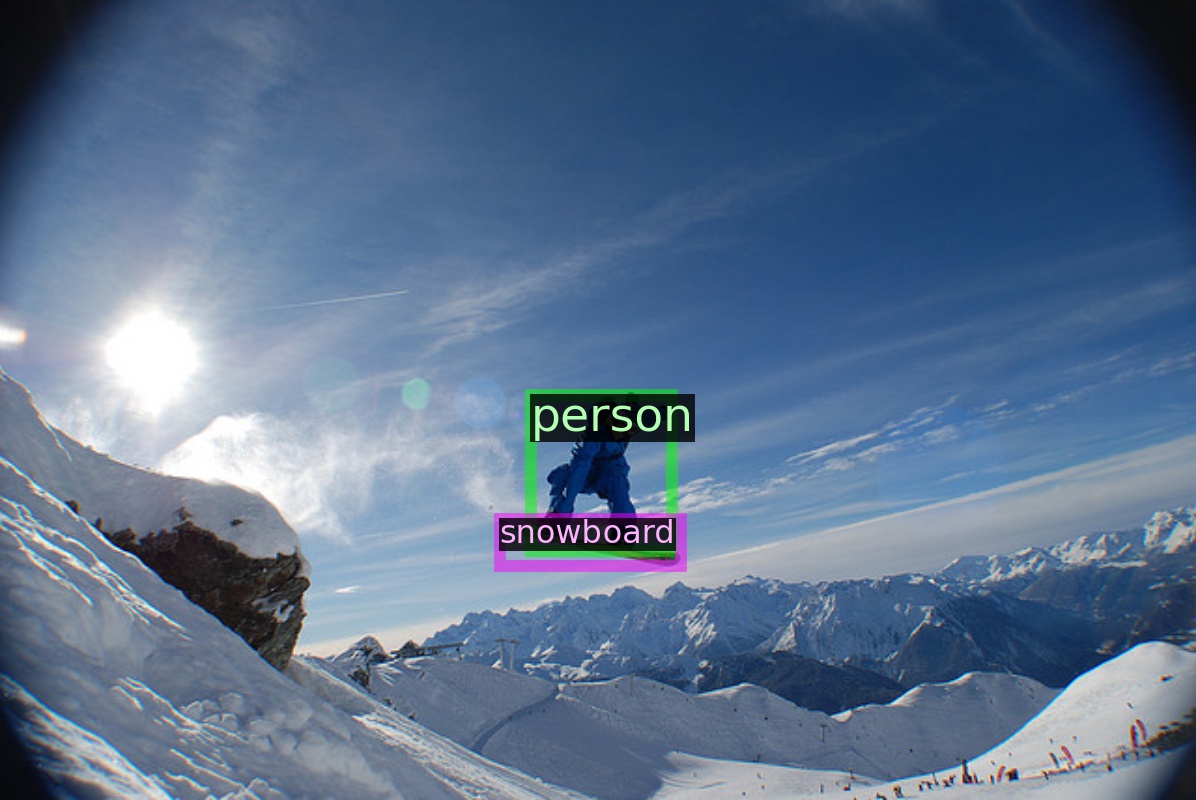} &
 \includegraphics[trim={0 0cm 0 0cm}, clip, width=30mm]{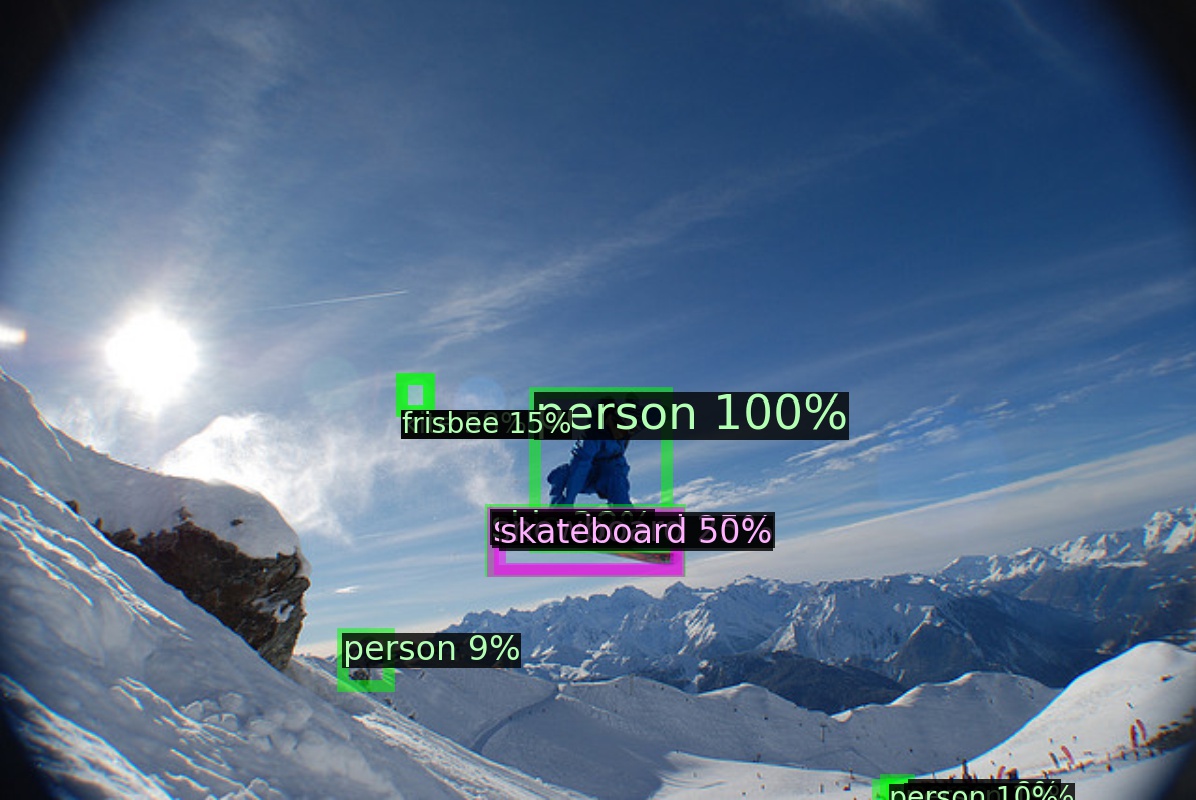} \\
 
 \includegraphics[trim={0 0cm 0 0cm}, clip, width=30mm]{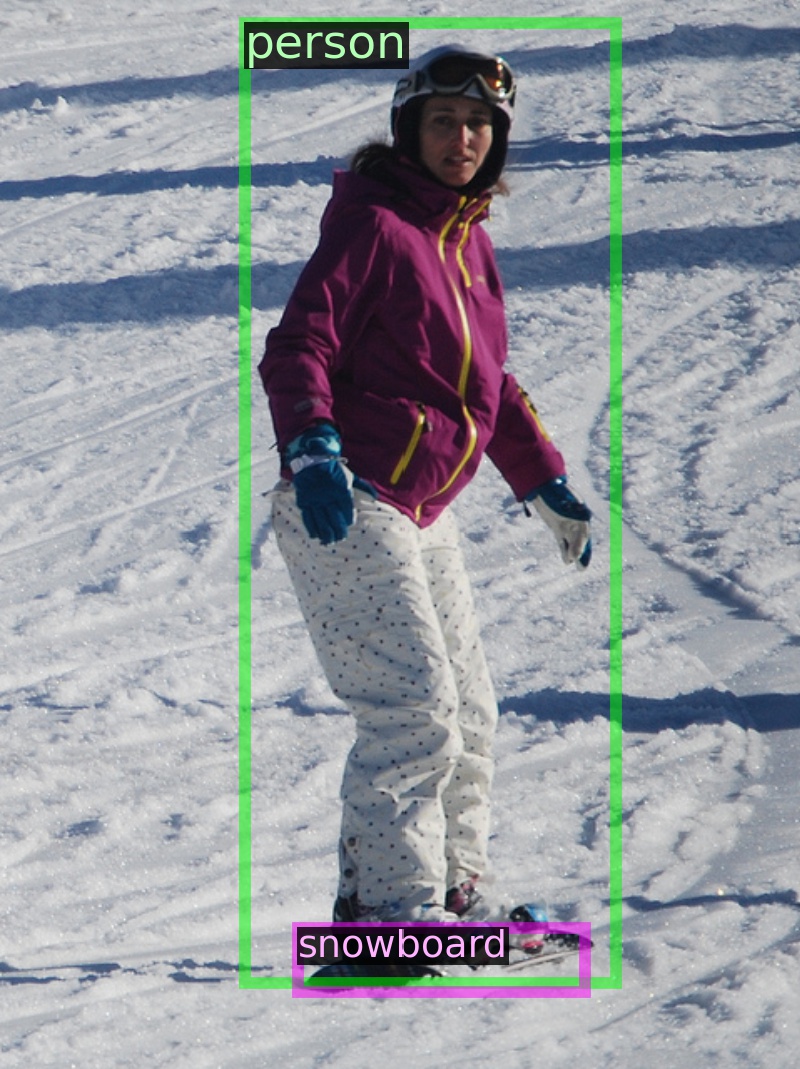} &
 \includegraphics[trim={0 0cm 0 0cm}, clip, width=30mm]{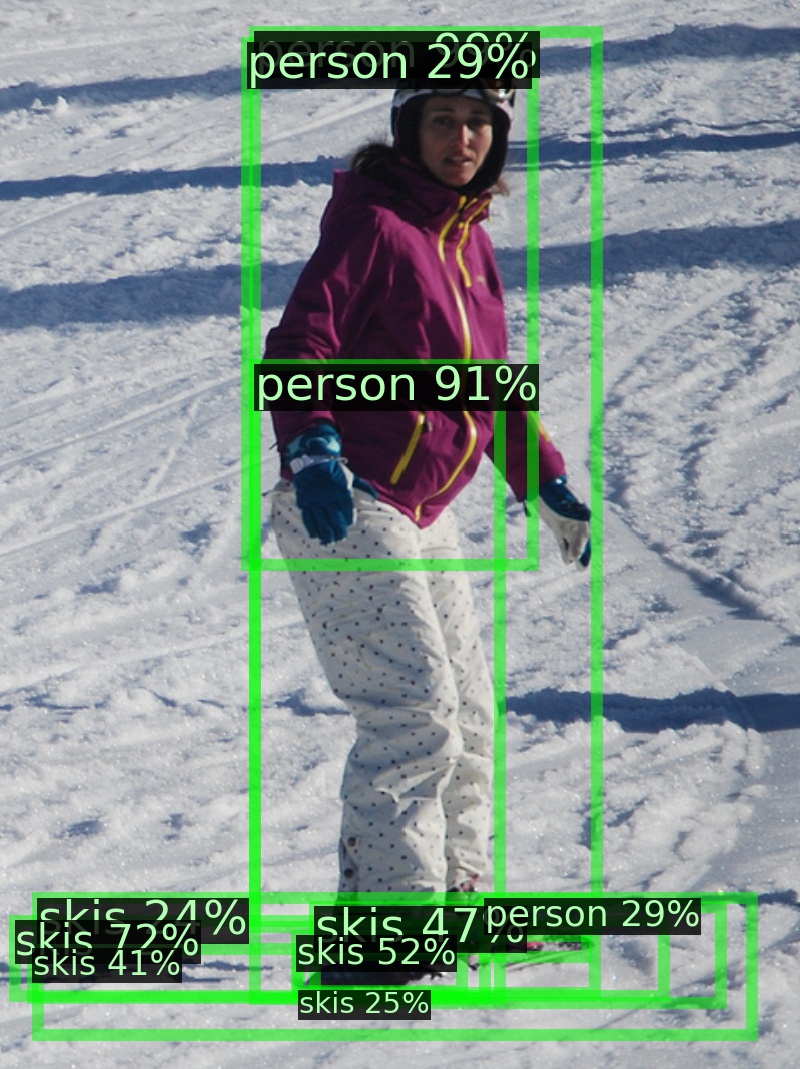} &
 \includegraphics[trim={0 0cm 0 1.0cm}, clip, width=30mm]{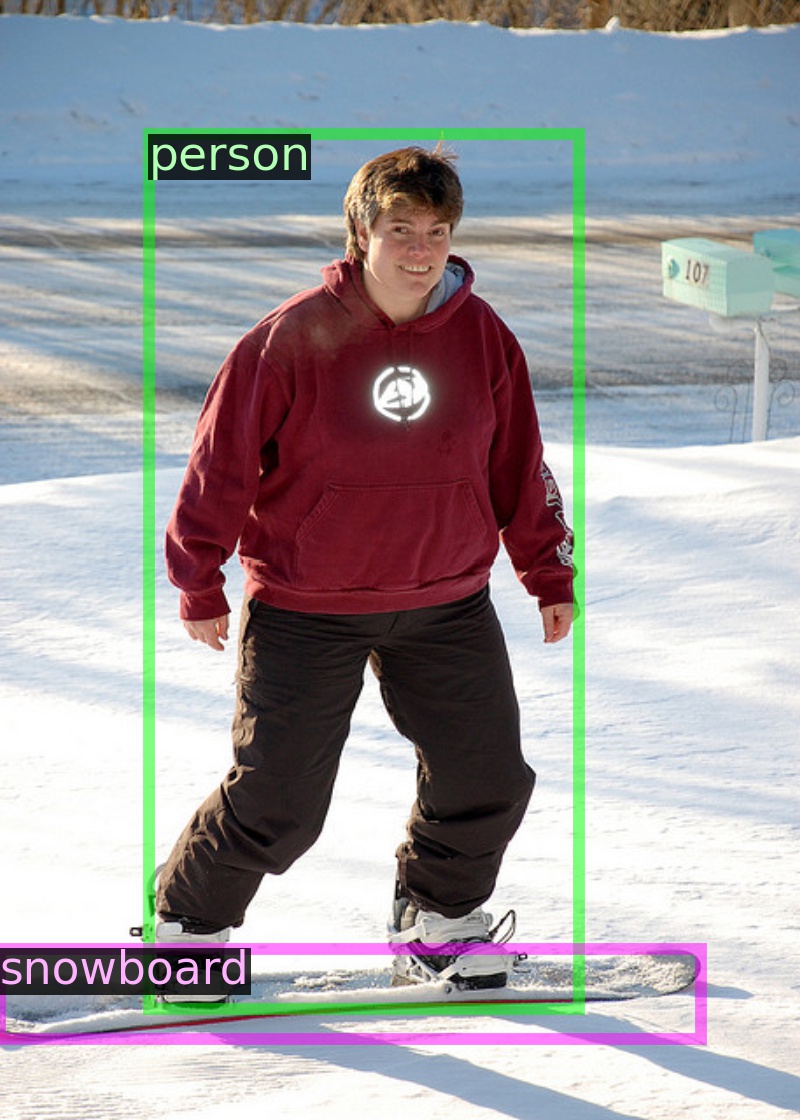} &
 \includegraphics[trim={0 0cm 0 1.0cm}, clip, width=30mm]{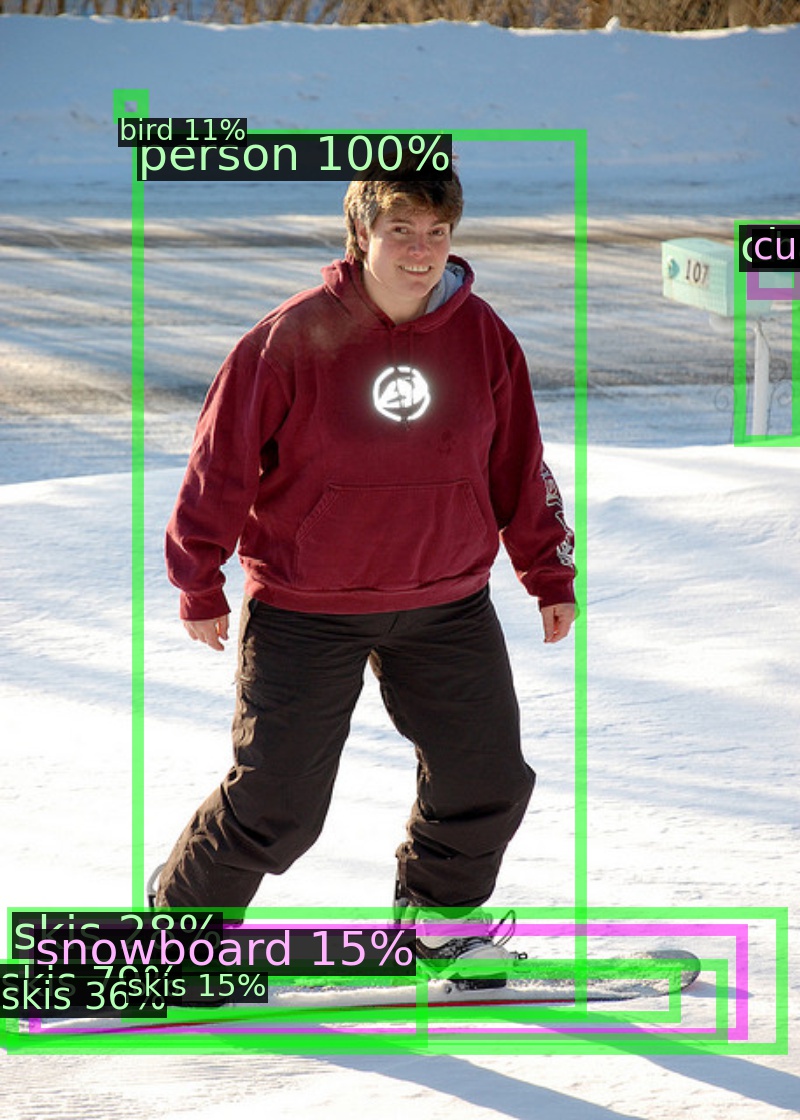}  \\
 
 \includegraphics[trim={0 0cm 0 0cm}, clip, width=30mm]{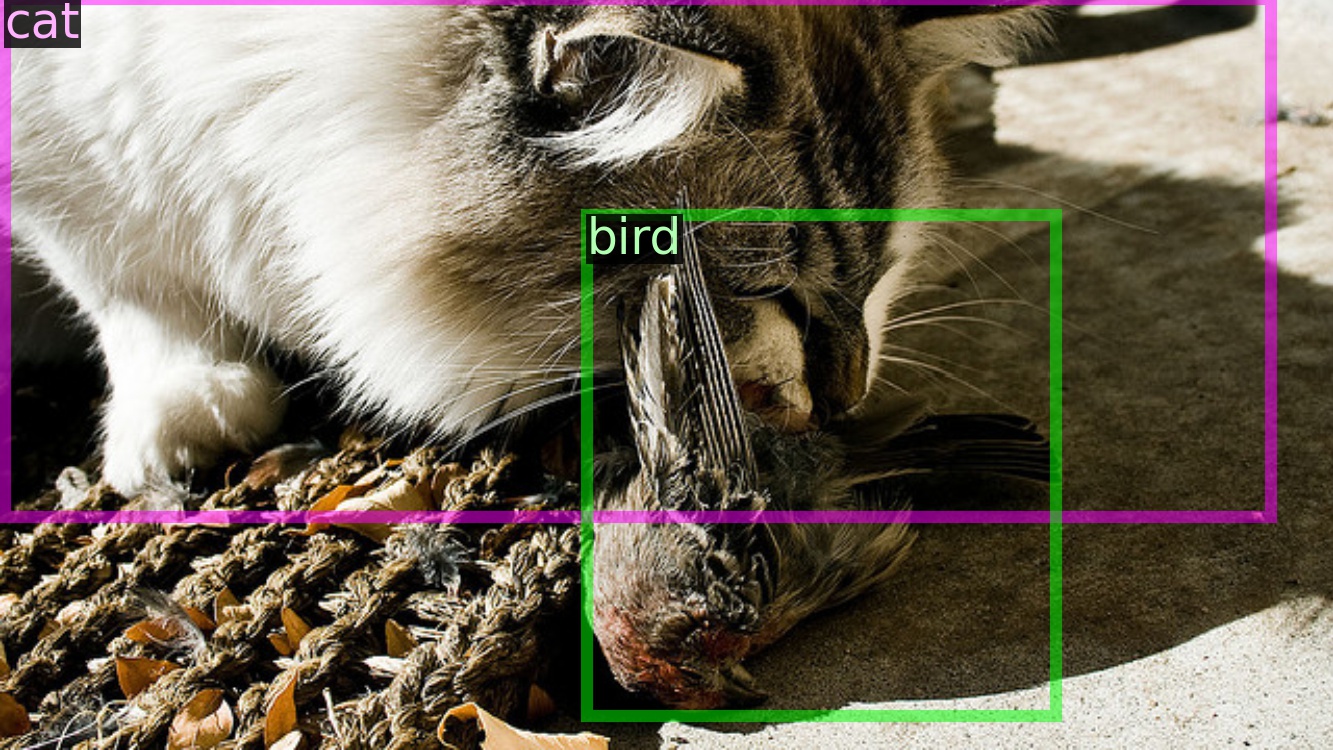} &
 \includegraphics[trim={0 0cm 0 0cm}, clip, width=30mm]{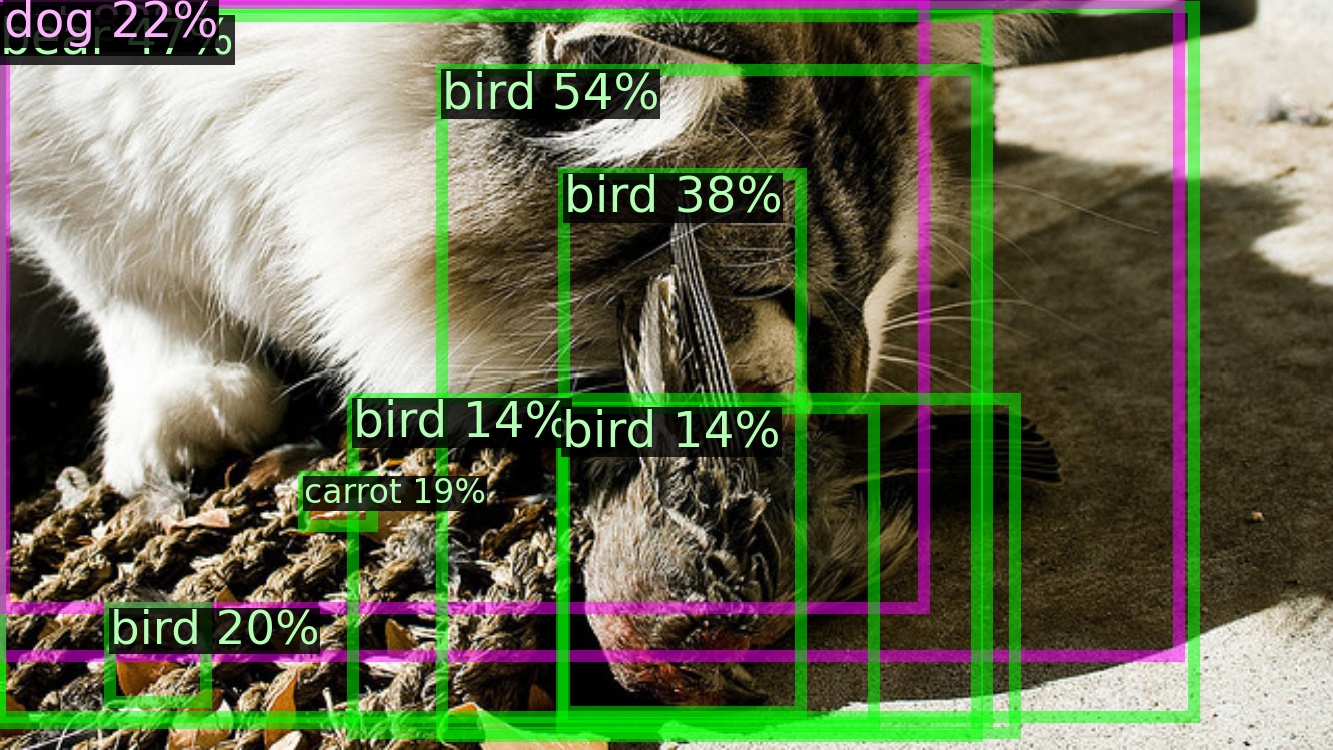} &
 \includegraphics[trim={0 1.0cm 0 1.5cm}, clip, width=30mm]{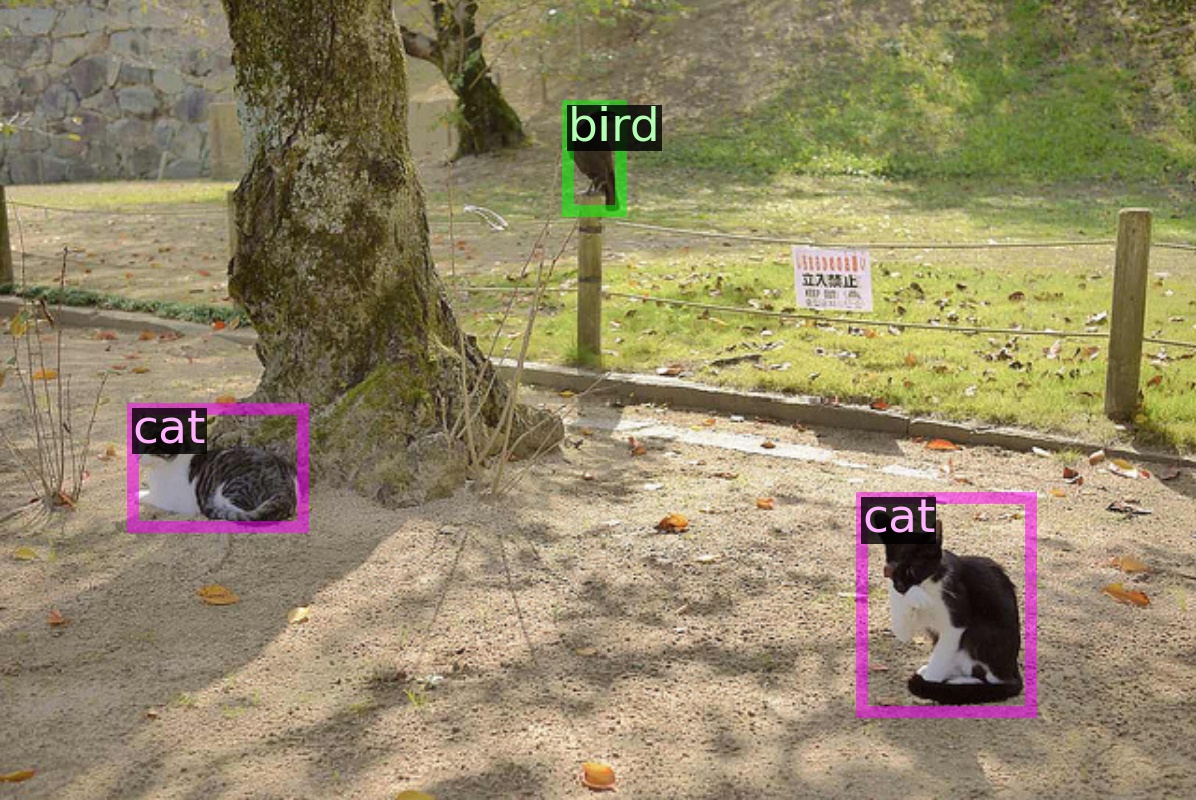} &
 \includegraphics[trim={0 1.0cm 0 1.5cm}, clip, width=30mm]{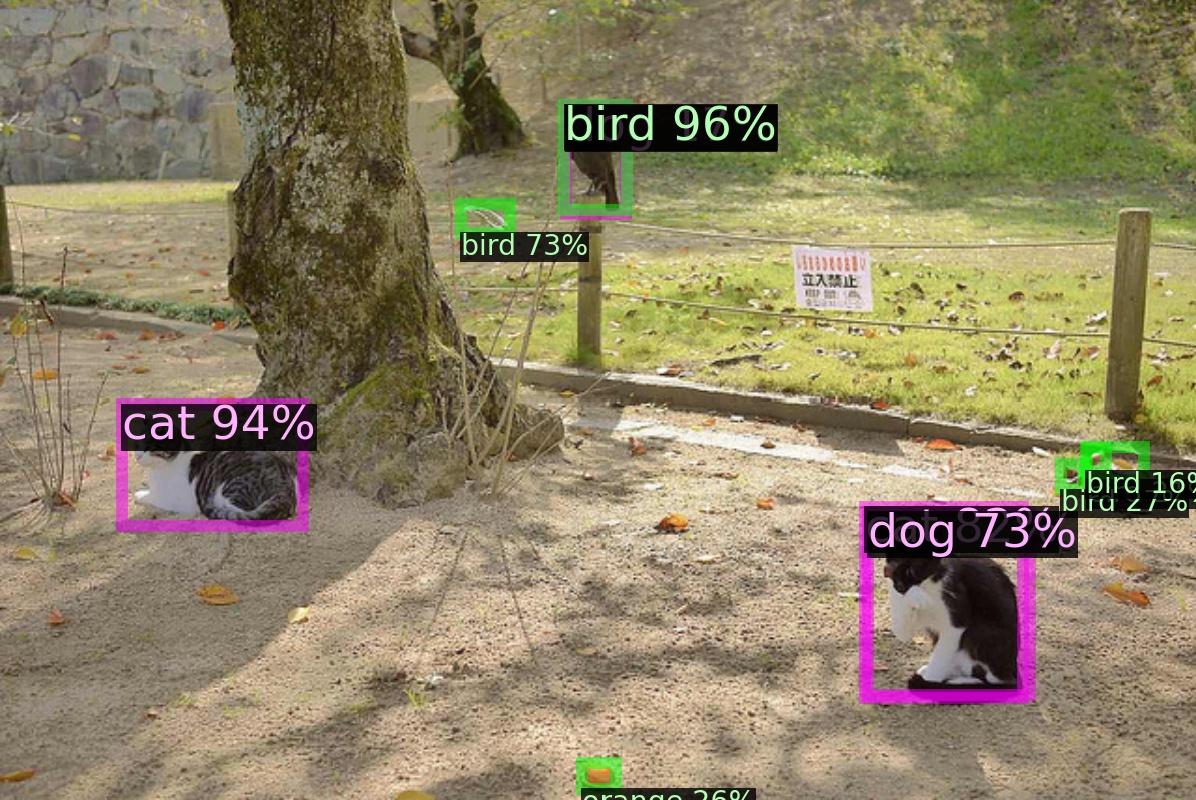}  \\
 
 \includegraphics[trim={0 0cm 0 0cm}, clip, width=30mm]{arvix_version/gt_000000271728.jpg} &
 \includegraphics[trim={0 0cm 0 0cm}, clip, width=30mm]{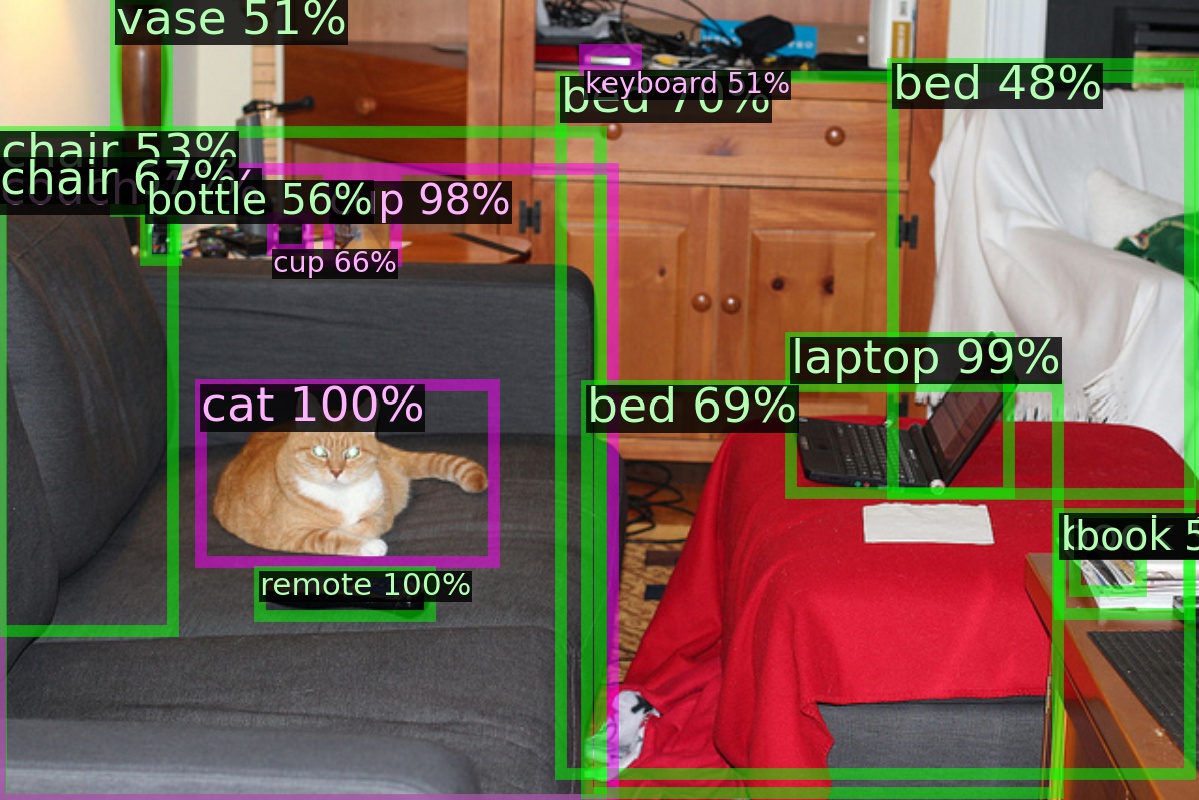} &
 \includegraphics[trim={0 1.0cm 0 0cm}, clip, width=30mm]{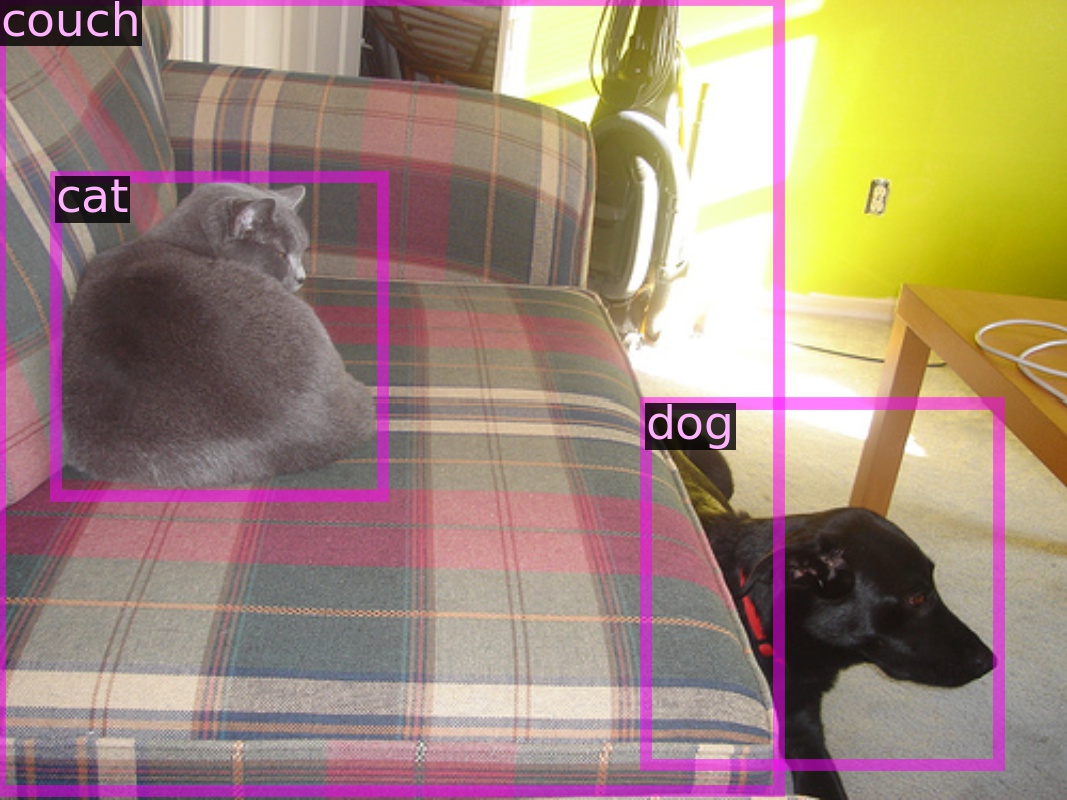} &
 \includegraphics[trim={0 1.0cm 0 0cm}, clip, width=30mm]{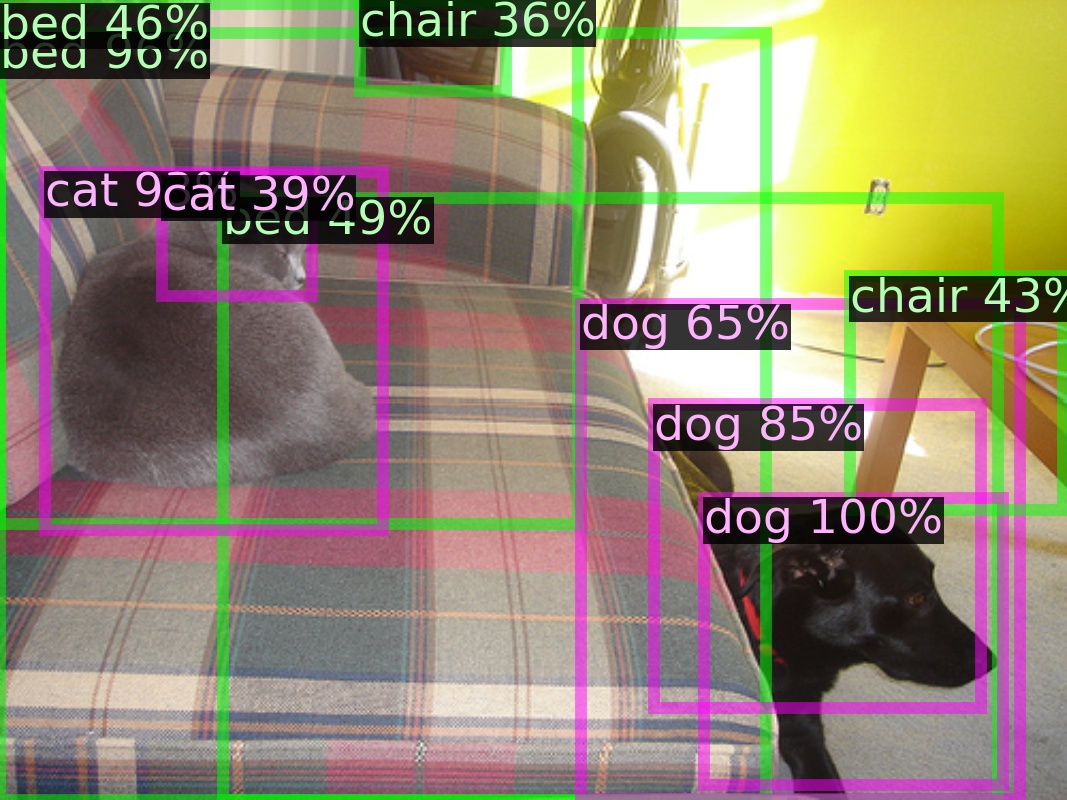}  \\
 
 \includegraphics[trim={0 0cm 0 0.5cm}, clip, width=30mm]{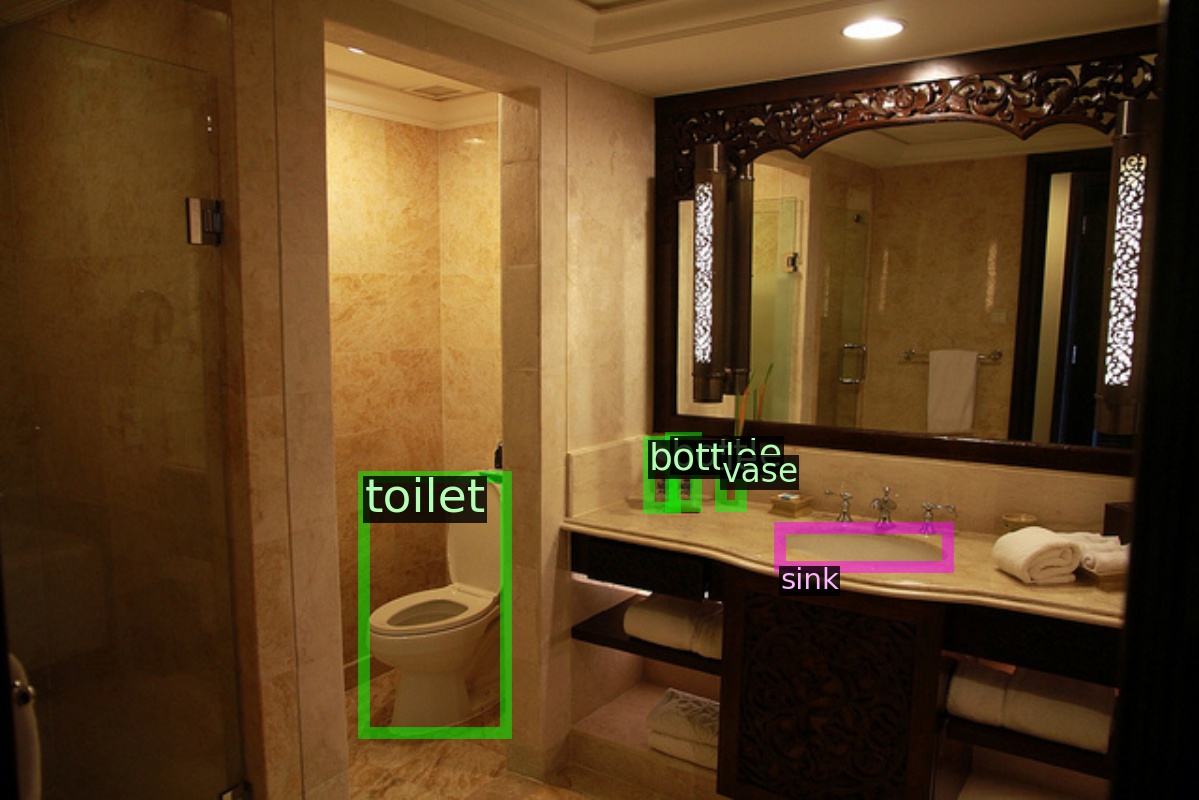} &
 \includegraphics[trim={0 0cm 0 0.5cm}, clip, width=30mm]{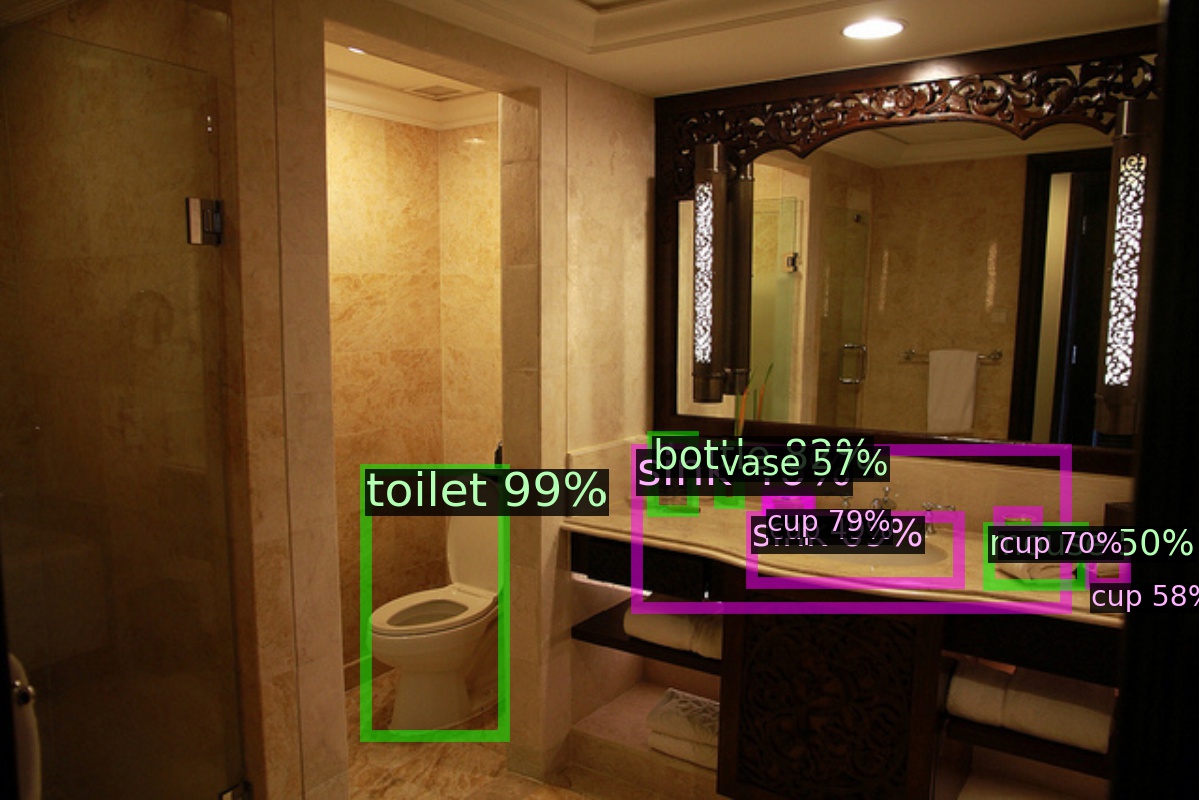} &
 \includegraphics[trim={0 0cm 0 0cm}, clip, width=30mm]{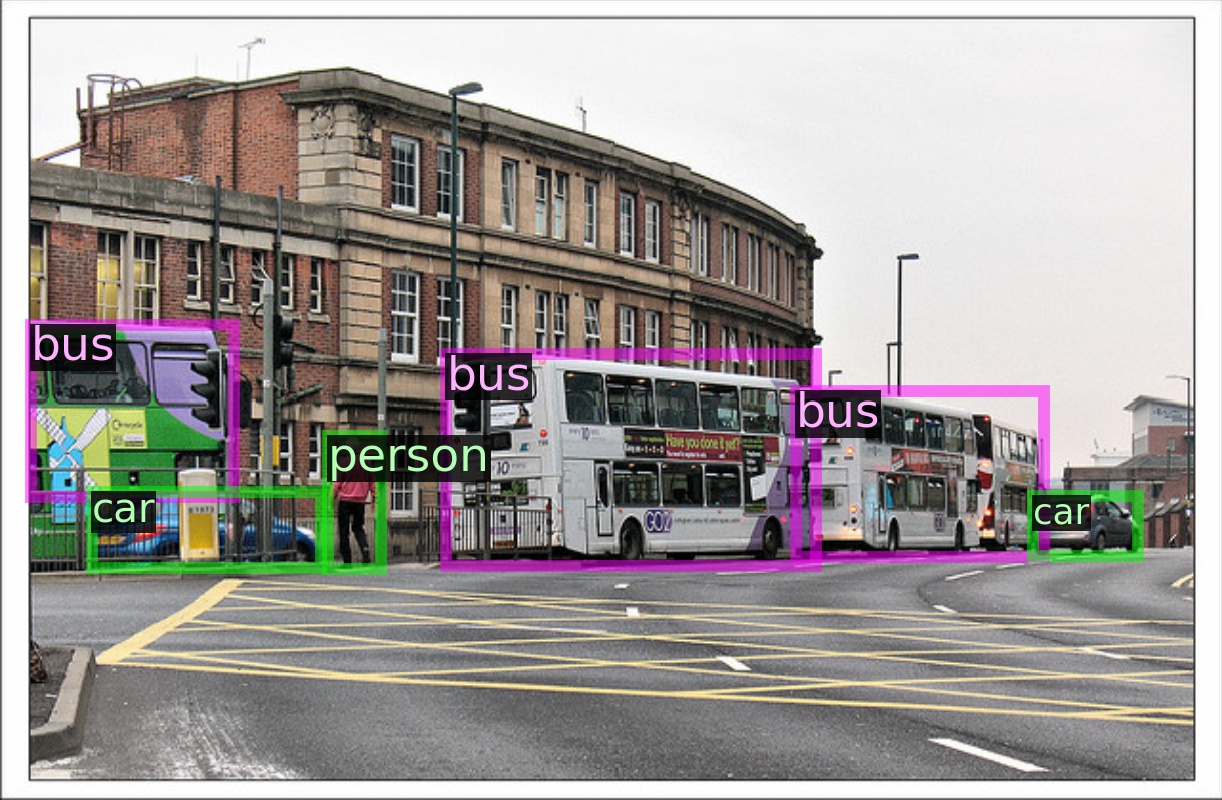} &
 \includegraphics[trim={0 0cm 0 0cm}, clip, width=30mm]{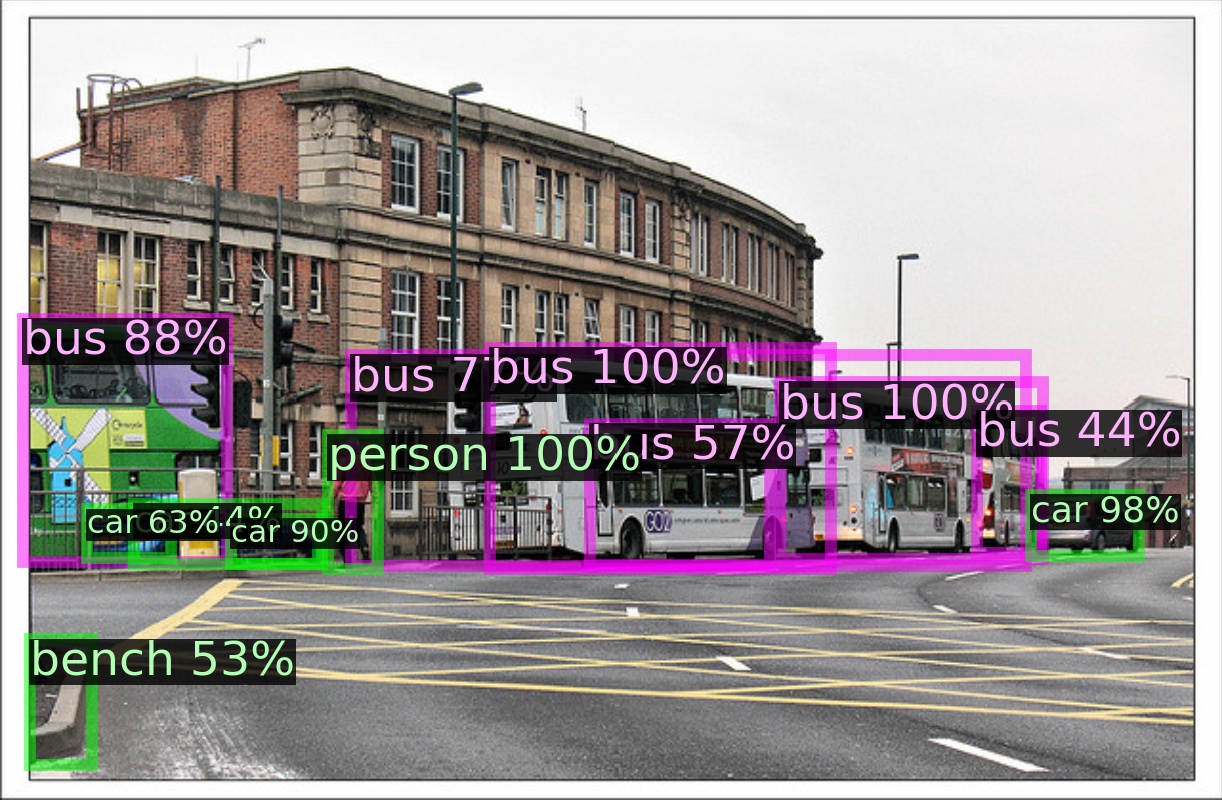}  \\

\end{tabular}
\caption{Qualitative results obtained using \modelname on the COCO dataset. Novel classes are shown in \textcolor{magenta}{magenta} while known are in \textcolor{green}{green}. (Best viewed in color)}
\label{fig:coco_qual_results_extended2}
\end{figure*}

Figures \ref{fig:coco_qual_results_extended1} and \ref{fig:coco_qual_results_extended2} show some random qualitative examples of \modelname. Our method is capable of discovering novel classes such as cat, dog, sink, bus with high confidence, specially when there is no ambiguity or similarity with other categories. Similar visual classes such as fork, knife and spoon; cow and sheep; cat and dog; couch and bed; or snowboard, skis, and skateboard or are sometimes confused by our model. 
 
\end{document}